\documentclass{article}

 \usepackage[preprint]{neurips_2026}

\usepackage[utf8]{inputenc} 
\usepackage[T1]{fontenc}    
\usepackage{hyperref}       
\usepackage{url}            
\usepackage{booktabs}       
\usepackage{amsfonts}       
\usepackage{nicefrac}       
\usepackage{microtype}      
\usepackage[dvipsnames]{xcolor}

\usepackage{xspace}
\def\onedot{.\xspace}
\def\etal{\emph{et al}\onedot}
\usepackage{graphicx}
\newcommand{\fittable}[1]{\resizebox{\ifdim\width>\linewidth\linewidth\else\width\fi}{!}{#1}}
\usepackage{rotating}
\usepackage{pifont}
\usepackage{amssymb}
\usepackage{amsmath} 
\usepackage{caption}

\usepackage{pifont}
\usepackage{pifont}
\newcommand{\cmark}{\ding{51}}
\newcommand{\xmark}{\ding{55}}
\renewcommand{\cmark}{\textcolor{green!70!black}{\ding{51}}}
\renewcommand{\xmark}{\textcolor{red!70!black}{\ding{55}}}
\newcommand{\pmark}{%
  \textcolor{orange!80!black}{%
    \raisebox{0.2ex}{\scalebox{1.4}{$\sim$}}%
  }%
}

\title{RipBench: A Unified Benchmark for Multi-Level\\Rip Current Detection, Classification and Segmentation}

\author{
Andrei Dumitriu$^{1, 2}$, 
Aakash Ralhan$^{1}$, 
Florin Tatui$^{2}$, 
Florin Miron$^{2}$,\\ 
\textbf{Radu Tudor Ionescu}$^{2}$, \textbf{Radu Timofte}$^{1}$ \\
$^{1}$Computer Vision Lab, CAIDAS \& IFI, University of Würzburg, Germany\\
$^{2}$University of Bucharest, Romania\\
\texttt{\small andrei.dumitriu@uni-wuerzburg.de}\\
}

\begin{document}

\maketitle

\begin{abstract}
Rip currents are a serious and often under-addressed threat to beach safety and the leading cause of coastal drownings worldwide. They are difficult to detect due to their amorphous structure, similarity to the background, and large variability in viewpoints and environments. Despite their dangers and growing interest in automated detection, existing research remains fragmented across datasets, metrics, models, and approaches, leading to an incomplete understanding of model performance. We introduce RipBench, a unified benchmark that enables controlled evaluation of rip current detection spanning multiple tasks, including classification, axis-aligned and oriented object detection, and instance, semantic, and panoptic segmentation, on the same data with standardized splits. This allows direct comparison of model performance on all levels of visual abstraction. Across $304$ videos ($303{,}491$ frames, $51{,}011$ of them annotated for all six tasks) collected from diverse coastlines, our results expose a clear performance gap between coarse recognition and precise spatial understanding: frame-level classification reaches $F_2 = 0.93$, while the best detector and the best instance segmenter reach $0.72$ and $0.66$, with model rankings that change from task to task. Video-level confidence intervals, a blind inter-annotator study of the spatial labels, stratified analyses by viewpoint and visual signature, and video views with temporal baselines (single-seed, short-schedule reference runs) accompany the benchmark. All tasks are supported with carefully curated annotations and evaluated using standard metrics, with a focus on the $F_2$ score to emphasize recall in this safety-critical setting. RipBench, along with multiple baselines per task, is publicly available at \url{https://RipBench.ai} to support progress in real-world, multi-task vision for beach safety.

\end{abstract}

\begin{figure*}
\centering
\setlength{\tabcolsep}{1pt}
\begin{tabular}{c c c c}
     Aerial - Bird's Eye & Aerial - Tilted & Elevated Beachfront & Water-Level Beachfront
     \tabularnewline
    
     \begin{turn}{90}{\raggedright Classif./Orig.}\end{turn}
     \includegraphics[width=0.236\textwidth]{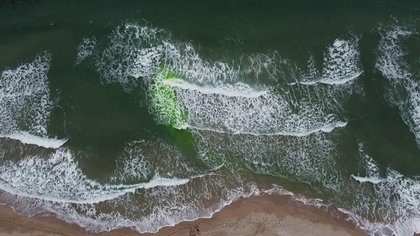} &
     \includegraphics[width=0.236\textwidth]{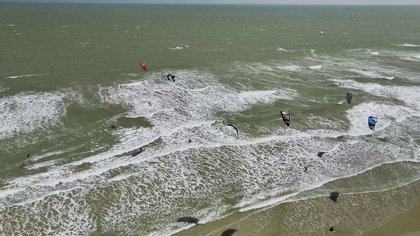} &
     \includegraphics[width=0.236\textwidth]{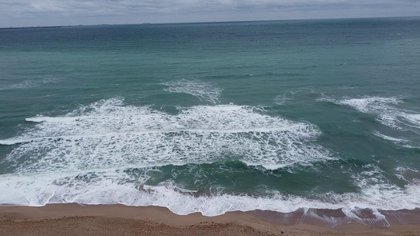} &
     \includegraphics[width=0.236\textwidth]{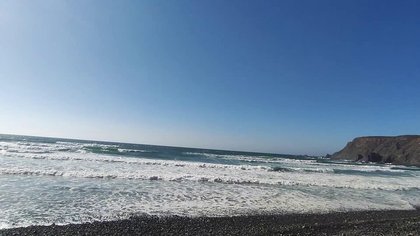}
     \tabularnewline
     
     \begin{turn}{90}{\raggedright BBox}\end{turn}
     \includegraphics[width=0.236\textwidth]{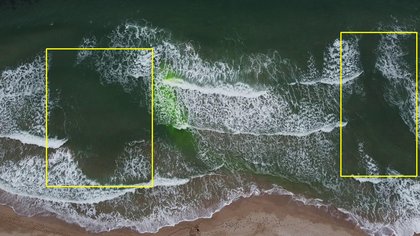} &
     \includegraphics[width=0.236\textwidth]{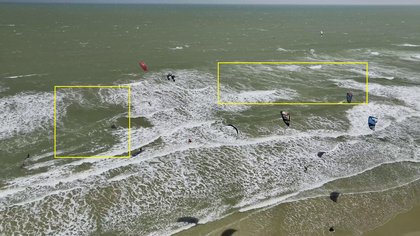} &
     \includegraphics[width=0.236\textwidth]{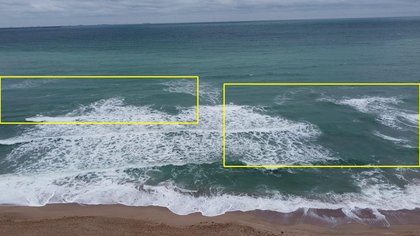} &
     \includegraphics[width=0.236\textwidth]{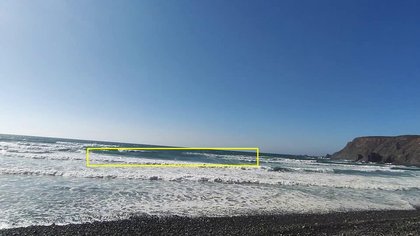}
     \tabularnewline
     
     \begin{turn}{90}{\raggedright Or. BBox}\end{turn}
     \includegraphics[width=0.236\textwidth]{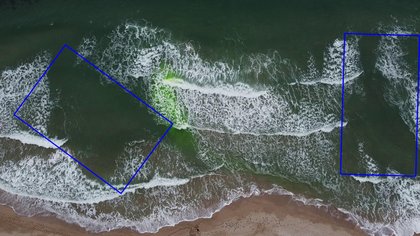} &
     \includegraphics[width=0.236\textwidth]{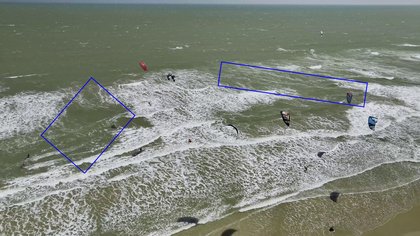} &
     \includegraphics[width=0.236\textwidth]{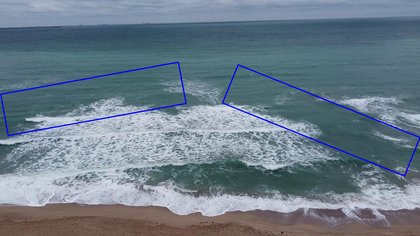} &
     \includegraphics[width=0.236\textwidth]{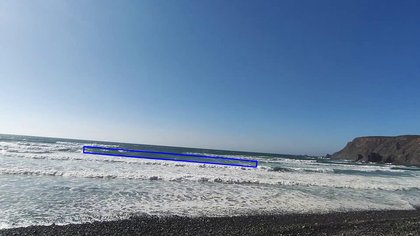}
     \tabularnewline
     
     \begin{turn}{90}{\raggedright Inst. Seg.}\end{turn}
     \includegraphics[width=0.236\textwidth]{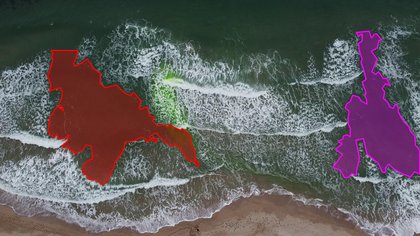} &
     \includegraphics[width=0.236\textwidth]{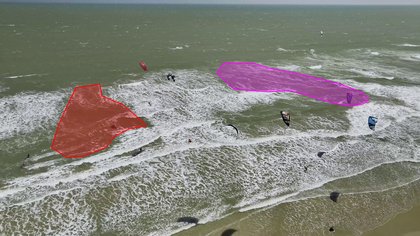} &
     \includegraphics[width=0.236\textwidth]{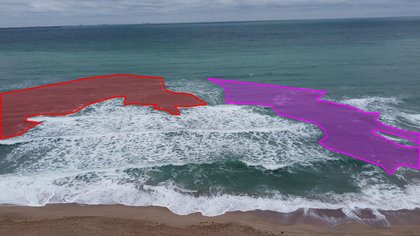} &
     \includegraphics[width=0.236\textwidth]{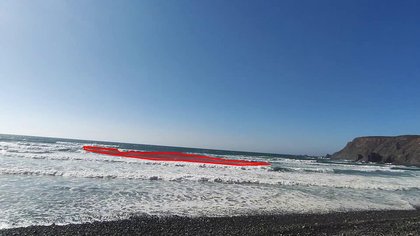}
     \tabularnewline
     
     \begin{turn}{90}{\raggedright Sem. Seg.}\end{turn}
     \includegraphics[width=0.236\textwidth]{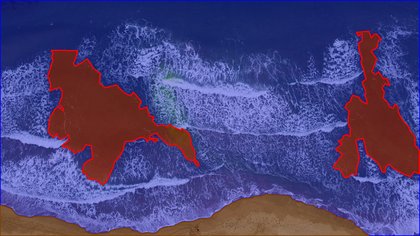} &
     \includegraphics[width=0.236\textwidth]{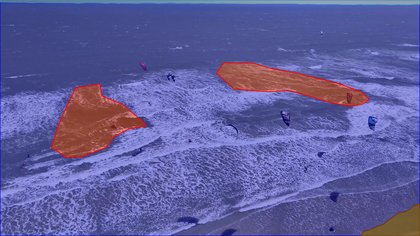} &
     \includegraphics[width=0.236\textwidth]{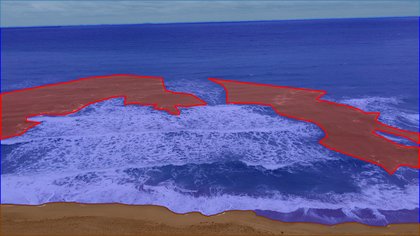} &
     \includegraphics[width=0.236\textwidth]{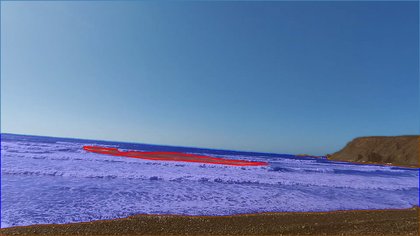}
     \tabularnewline
     
     \begin{turn}{90}{\raggedright Pan. Seg.}\end{turn}
     \includegraphics[width=0.236\textwidth]{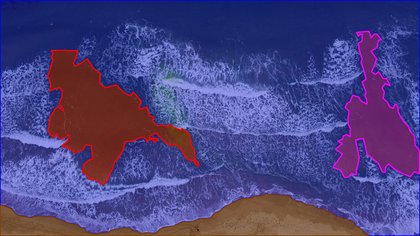} &
     \includegraphics[width=0.236\textwidth]{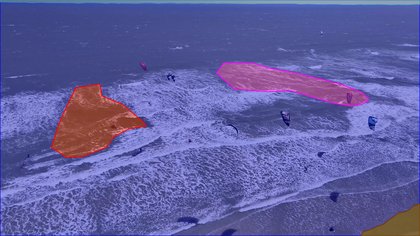} &
     \includegraphics[width=0.236\textwidth]{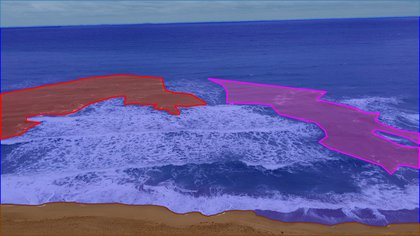} &
     \includegraphics[width=0.236\textwidth]{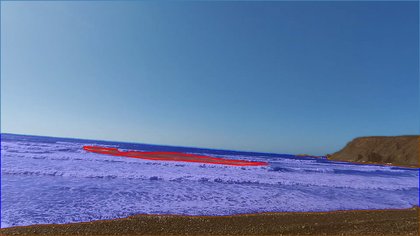}
\end{tabular}
\vspace{-0.25cm}
\caption{Examples from our dataset, showcasing diversity in locations, viewpoints, rip current appearance, and annotation types. Columns correspond to different camera viewpoints, while rows illustrate the supported task annotations and representations. Best viewed in color.}\label{fig:annotations}
\vspace{-0.35cm}
\end{figure*}

\vspace{-0.2cm}
\section{Introduction}
\vspace{-0.2cm}
Rip currents are powerful, fast-moving surface currents that flow seaward from coastlines, posing a significant threat to beachgoers worldwide \cite{brander2013brief,brewster2019estimations,da2003analysis, lushine1991study}. Found in oceans, seas, and large lakes, these hazardous flows vary in size and speed, influenced by nearshore hydrodynamics, underwater morphology, and, at times, human-made coastal structures \cite{brander2000morphodynamics, castelle2016rip}. With speeds reaching up to $8.7$ km/h, faster than an Olympic swimmer, rip currents can quickly pull individuals offshore \cite{noaa2023ripcurrents}. The danger is often exacerbated by a lack of public awareness: swimmers commonly panic and attempt to fight the current, leading to exhaustion and, tragically, drownings. Recommended responses, such as swimming parallel to the shore, are effective only if the current is correctly identified, preferably before entering the water.

One of the most common and dangerous misconceptions about rip currents is that they are easy to identify. However, studies across continents consistently show that beachgoers are unable to recognize them reliably, most often misidentifying hazardous areas as safe (see Appendix~\ref{sec:global_studies} for the underlying studies).

Rip currents pose unique vision challenges due to their amorphous, shifting appearance, often blending into the dynamic water environment through subtle cues such as wave breaks, sediment flows, or water color changes. Addressing these complexities requires a combination of tasks, each offering strengths in different scenarios (see Fig.~\ref{fig:annotations}). While segmentation provides fine-grained localization, classification can be more suitable for fast presence alerts, bounding boxes offer interpretable localization cues, oriented bounding boxes help capture flow direction and instance segmentation provides both direction and exact shape, at the cost of increased annotation difficulty. Together with semantic and panoptic segmentation to contextualize the broader scene, these tasks enable more flexible and robust detection systems adaptable to diverse beach monitoring settings.

To address this gap, we introduce \textbf{RipBench}, the first complete, unified and multi-level benchmark for rip current training and evaluation, together with multiple models for each task, including state-of-the-art, real-time and popular methods. In summary, our contributions include: 
\begin{itemize}
    \item \textbf{Unified multi-level evaluation framework.} We introduce RipBench, a benchmark that enables controlled evaluation across multiple task formulations: classification, detection (axis-aligned and oriented), and segmentation (instance, semantic, and panoptic), on the same underlying phenomenon, allowing direct comparison across levels of visual abstraction.
    \item \textbf{Systematic analysis of task-dependent performance.} We conduct a comprehensive evaluation, with video-level confidence intervals, showing that model performance varies significantly across tasks: classification is far ahead ($F_2{=}0.93$), while localization remains substantially harder (best $F_2$ is $0.72$ for detection and $0.66$ for instance segmentation), and the localization gap is a matter of architecture family rather than of task: within the strongest lineage, boxes are slightly ahead of masks, revealing a gap between coarse recognition and precise spatial understanding that survives target harmonization.
    \item \textbf{Representation-specific failure modes and non-transferability.} We demonstrate that different task formulations exhibit distinct failure modes, and that conclusions about model effectiveness depend strongly on the chosen representation, highlighting the limitations of evaluating models in isolation.
    \item \textbf{Annotation framework for ambiguous, amorphous phenomena.} We analyze the inherent ambiguity of rip current boundaries and show that even bounding box and oriented box annotations depend on expert judgment. We introduce practical annotation guidelines (e.g., tighter boxes than full masks) for signal-to-background ratio and evaluation consistency. The box-protocol ablation (Appendix~\ref{sec:box_protocol_ablation}) shows that neither box protocol consistently improves $F_2$, so the guidelines are only meant to standardize evaluation.
    \item \textbf{Balanced multi-task dataset enabling controlled comparison.} RipBench comprises 304 videos ($303,491$ frames), evenly balanced between rip and non-rip scenarios, with consistent annotations across tasks, standardized splits, and increased temporal coverage through minimum $5$ FPS sampling. Unlike standard object benchmarks where the negative class is visually heterogeneous background, our non-rip videos depict the same water under similar conditions as our rip videos, providing meaningful false-positive evaluation and forcing models to learn fine-grained discrimination between flowing and non-flowing water.
\end{itemize}

\begin{table}

  \caption{Public rip current datasets and the vision tasks they support. \#Labeled = frames with annotations for the supported tasks; \pmark\ = not provided but derivable from the released annotations (frame labels from annotated and negative frames, boxes from masks or oriented boxes). $^{*}$RipScout reported $73$ videos, but only sampled frames from $8$ are publicly available.}
  \label{tab:rip_dataset_comparison}
  \centering
  \small
  \begin{tabular}{lrrrcccccc}
    \toprule
    & & & & & \multicolumn{2}{c}{\textbf{Detection}} & \multicolumn{3}{c}{\textbf{Segmentation}}\\
    \cmidrule(r){6-7}
    \cmidrule(r){8-10}
    \textbf{Dataset}                        & \textbf{Videos}       & \textbf{\#Frames} & \textbf{\#Labeled} & \textbf{\rotatebox{90}{Classif.}} & \textbf{\rotatebox{90}{BBox}} & \textbf{\rotatebox{90}{Or.BBox}} & \textbf{\rotatebox{90}{Inst.}} & \textbf{\rotatebox{90}{Sem.}} & \textbf{\rotatebox{90}{Pan.}}\\
    \midrule
    Maryan~\etal (2019)~\cite{maryan2019machine}   & $0$                     & $1{,}314$   &   $514$          & \xmark & \cmark & \xmark & \xmark & \xmark & \xmark \\
    de Silva~\etal (2021)~\cite{desilva2021frcnn}  & $23$                    & $19{,}782$  &   $1{,}740$      & \pmark & \cmark & \xmark & \xmark & \xmark & \xmark \\
    YOLO-Rip (2022)~\cite{zhu2022yolo}             & $0$                     & $22{,}580$  &   $2{,}486$      & \pmark & \cmark & \xmark & \xmark & \xmark & \xmark \\
    Dumitriu~\etal (2023)~\cite{dumitriu2023rip}   & $17$                    & $24{,}950$  &   $7{,}325$      & \pmark & \pmark & \pmark & \cmark & \xmark & \xmark \\
    RipScout (2025)~\cite{khan2025ripscout}        & $0$\textsuperscript{*}  & $2,555$     &   $2{,}555$      & \xmark & \cmark & \xmark & \xmark & \xmark & \xmark \\
    RipVIS (2025)~\cite{dumitriu2025ripvis}        & $184$                   & $212{,}328$ &   $15{,}784$     & \pmark & \pmark & \pmark & \cmark & \xmark & \xmark \\
    RipAID (2025)~\cite{soriano2025ripaid}         & $0$                     & $6{,}789$   &   $6{,}789$      & \xmark & \pmark & \cmark & \xmark & \xmark & \xmark \\
    \textbf{RipBench (ours)}                       & $\mathbf{304}$          & $\mathbf{303{,}491}$ & $\mathbf{51{,}011}$ & \cmark & \cmark & \cmark & \cmark & \cmark & \cmark \\
    \bottomrule
  \end{tabular}
  \vspace{-0.3cm}
\end{table}

\vspace{-0.2cm}
\section{Related Work}
\vspace{-0.2cm}
\label{sec:related_work}
Rip current detection sits at the intersection of coastal science, remote sensing and computer vision, with decades of geophysical research on causes, behavior and morphology laying the foundation for recent vision-based approaches. We review recent datasets, methods and challenges for automatic detection.

\vspace{-0.2cm}
\subsection{Rip Currents Vision Datasets}
\vspace{-0.2cm}
Existing datasets typically target a single task formulation, limiting controlled comparison across representations (see Table~\ref{tab:rip_dataset_comparison}). Early work introduced classification and bounding box annotations on timex patches and image-level data~\cite{desilva2021frcnn,maryan2019machine}, with later datasets expanding to instance segmentation~\cite{dumitriu2023rip,dumitriu2025ripvis}, oriented detection~\cite{soriano2025ripaid}, and mobile or drone settings~\cite{khan2025ripscout}. Choi~\etal~\cite{choi2024explainable} reported a large bounding box dataset of over $110{,}000$ images, but access is restricted, limiting reproducibility. Across these efforts, annotations, metrics and evaluation protocols remain inconsistent. RipBench is the largest of these datasets and the only one with official annotations for all six task formulations. Classification labels, axis-aligned boxes and oriented boxes can in principle be derived from instance masks. RipBench nevertheless releases separate expert passes for boxes and oriented boxes (see details in Sec.~\ref{sec:bbox_vs_mask}) and treats them as the official targets, so that results across methods are measured against one shared ground truth.

\vspace{-0.2cm}
\subsection{Detection and Segmentation Methods}
\vspace{-0.2cm}
Early approaches relied on optical flow, handcrafted features and anomaly detection to identify offshore motion patterns~\cite{de2023ripviz,mori2022flow,philip2016flow}. While interpretable, these methods depended on camera stability and fixed viewpoints. Subsequent work adopted CNN-based classification and detection~\cite{desilva2021frcnn,maryan2019machine,rashid2023reducing,rashid2020ripnet,rashid2021ripdet}, followed by YOLO, EfficientDet and transformer architectures for real-time and mobile deployment~\cite{khan2025ripfinder,khan2025ripscout,zhu2022yolo}, and temporal forecasting for early warning~\cite{wan2026ripalert}. Segmentation approaches include instance segmentation with YOLOv8-seg~\cite{dumitriu2023rip}, large-scale video instance benchmarks emphasizing recall-oriented $F_2$ metrics~\cite{dumitriu2025ripvis}, box-supervised segmentation~\cite{box2rip2025}, GAN-based augmentation~\cite{qian2025ripgan}, and oriented detection with safety-critical metrics~\cite{soriano2025ripaid}. Community challenges such as RipSeg~\cite{ripseg2025} and RipDetSeg~\cite{ripdetseg2026} introduced standardized settings for segmentation and joint detection-segmentation.

Despite this progress, prior work remains fragmented across datasets, annotations, metrics and task formulations. RipBench addresses these limitations by providing unified annotations across classification, axis-aligned and oriented detection, and instance, semantic and panoptic segmentation on the same videos and standardized splits, under a shared evaluation framework.

\vspace{-0.2cm}
\section{RipBench: Unified Evaluation Framework}
\vspace{-0.2cm}

RipBench formulates rip current understanding as a multi-level perception problem spanning increasing levels of spatial detail: from binary classification of rip current presence, through axis-aligned and oriented detection, to instance, semantic, and panoptic segmentation.

\vspace{-0.2cm}
\subsection{Tasks and Metrics}
\label{sec:task_and_metrics}
\vspace{-0.2cm}

RipBench supports six tasks across classification, detection, and segmentation, evaluated on identical train, validation, and test splits. We report standard task-appropriate metrics ($IoU$, $AP$, $mAP$, $F_1$, $PQ$, $RQ$; full definitions and per-task choices in Appendix~\ref{sec:metrics}). Given the safety-critical nature of rip current detection, we emphasize the $F_2$ score across all tasks where it applies: $F_2$ prioritizes recall over precision and reflects the substantially higher cost of missed detections relative to false positives in real-world beach monitoring.
\vspace{-0.2cm}
\subsection{Dataset Overview}
\vspace{-0.2cm}

The RipBench dataset comprises $304$ videos ($303{,}491$ frames), evenly balanced between $152$ rip current and $152$ non-rip videos. Every video is sampled in frame order at a minimum of $5$ FPS, yielding $51{,}011$ annotated frames ($31{,}083$ train / $10{,}369$ validation / $9{,}559$ test), each carrying labels for all six tasks. The dataset includes diverse viewpoints (see Fig.~\ref{fig:annotations}), environmental conditions, and imaging modalities collected across multiple coastlines worldwide, and all tasks are defined over the same frames and the same video-level splits, enabling direct comparison between classification, detection, and segmentation models under a unified evaluation framework.

\vspace{-0.2cm}
\subsection{Collection and Diversity}
\label{sec:collection_diversity}
\vspace{-0.2cm}
\textbf{Provenance.} RipBench builds on the video pool of RipVIS \cite{dumitriu2025ripvis}, which we collected in prior work and which itself contains the videos from de Silva~\etal  \cite{desilva2021frcnn}: all $150$ RipVIS rip videos and $33$ of its $34$ non-rip videos (excluding duplicates) are inherited, and all of them were re-annotated from scratch for RipBench (Sec.~\ref{sec:protocol_details}). To this pool, we add $121$ newly collected videos: $2$ rip videos, including the only flash-rip footage in any public rip current dataset ($1{,}490$ annotated frames), and $119$ non-rip videos. Newly contributed footage thus accounts for $121$ of the $304$ videos and $15{,}051$ of the $51{,}011$ annotated frames ($29.5\%$). Yet, the annotation content is new for the entire pool. The $3{,}773$ standalone stills of the opt-in \texttt{additional\_data} scenario come from our earlier work~\cite{dumitriu2023rip}, which assembled them from the YOLO-Rip collection of Zhu~\etal~\cite{zhu2022yolo} (itself including the images from de Silva~\etal~\cite{desilva2021frcnn}) and annotated them from scratch. No upstream annotation is reused, and every RipBench annotation on them (boxes, oriented boxes and instance masks) is ours. A per-video attribution file (origin, RipVIS identifier, license, viewpoint, visual signature, camera motion, native resolution and frame rate) ships with the dataset, so every number in this paper can be recomputed per source. The resulting dataset is balanced at $152$ rip current and $152$ non-rip videos. Including non-rip videos is essential for both training and evaluation, and rip current detection makes this especially important. In standard object detection, the implicit negative class is everything that is not the object: trees, sky, varied textures. For rip currents, the negative class is the same water under the same viewing conditions, differing only in the presence of seaward flow. Models trained only on rip-positive videos can therefore learn to associate generic beach water with rip presence, rather than discriminating between rip and no rip instances. Balancing the dataset with $152$ non-rip videos collected under conditions visually similar to the rip videos (same viewpoints, lighting, and wave action) forces models to learn this finer-grained distinction, and makes false-positive evaluation meaningful by ensuring negative scenes look like the positive ones rather than being trivially distinguishable. 

All videos are sampled at a minimum of 5 FPS, substantially increasing annotated frame coverage relative to prior benchmarks (see Table~\ref{tab:rip_dataset_comparison}) and improving the statistical reliability of performance metrics. The dataset spans four viewpoint types: water-level beachfront, elevated beachfront, aerial tilted, and aerial bird's eye (top-down drone view, see Fig.~\ref{fig:annotations}). Frame rates range from $24$ to $60$ FPS and resolutions from $360 \times 640$ to $4096 \times 2160$ pixels.

\vspace{-0.2cm}
\subsection{Annotation Protocol}
\label{sec:protocol_details}
\vspace{-0.2cm}
All $51{,}011$ sampled frames are annotated consistently across the six tasks, and the same frame sets serve every task. Rip currents are annotated as instance masks, as manually drawn axis-aligned bounding boxes (BBox) and as oriented bounding boxes (OBB) on the $29{,}415$ frames in which a rip is visible ($41{,}547$ rip instances). Semantic and panoptic masks covering the three contextual classes (sky, water and land) are provided for every frame, rip and non-rip alike, so that non-rip frames carry scene labels, but no rip masks. The classification label is the per-frame presence of a rip annotation, which places the $344$ frames inside rip videos where the rip leaves the field of view in the negative class. For the spatial tasks, all remaining frames act as negatives. Annotation proceeded frame by frame in temporal order, so that the annotator always sees the motion context in which the rip current manifests. This unified scheme ensures all tasks are defined over the same data, enabling direct comparison across representations.

Annotation was performed by two co-authors with domain expertise in coastal geomorphology and rip current dynamics. To validate annotation consistency, an initial pilot subset was double-annotated independently by both experts, yielding Cohen's $\kappa{=}0.82$ on frame-level rip presence. This is a pilot figure measured \emph{before} the consensus step described next, and it is reported as such: the ambiguous cases it surfaced (rip-like patterns in non-rip footage, rips in unusually calm water) were then resolved by discussion, after which presence judgments are no longer independent and $\kappa$ cannot be honestly re-measured. Given the high pilot agreement, the remaining frames were partitioned between annotators, with each annotator's outputs subsequently reviewed by the other and any disagreements resolved through discussion.

\textbf{Blind agreement study for the spatial labels.} What the pilot did not measure is geometric agreement: given consensus that a rip is present, how consistently do two experts delineate it? To quantify this, each expert re-annotated, blind to the released labels, $200$ frames originally annotated by the other ($400$ frames, $600$ rip-instance pairs per task, spanning all rip videos, viewpoints and the three spatial identification tasks). Median expert-to-expert IoU is $0.805$ for boxes, $0.759$ for oriented boxes and $0.770$ for instance masks (Dice $0.86$--$0.89$), and $95$--$98\%$ of matched instances exceed the benchmark's IoU $0.5$ operating threshold, while only $52$--$68\%$ exceed IoU $0.75$. Hence, the released labels are consistent where the evaluation operates, and the strict-IoU regime is dominated by the boundary ambiguity of the phenomenon rather than by annotation error. Agreement is nearly viewpoint-invariant (box median IoU $0.79$--$0.81$ across the four viewpoints), so the viewpoint effects reported in Appendix~\ref{sec:stratified} correspond to model limitations, not to annotation artifacts. Full tables, including per-viewpoint statistics, are discussed in Appendix~\ref{sec:iaa_appendix}.

\vspace{-0.2cm}
\subsection{Annotation Under Ambiguity}
\label{sec:bbox_vs_mask}
\vspace{-0.2cm}
Unlike rigid objects with well-defined boundaries, rip currents are dynamic, amorphous phenomena whose spatial extent is inherently difficult to delineate. Their appearance depends on complex interactions between waves, currents, sediment, and lighting, often producing diffuse or partially observable structures. Annotation therefore requires expert judgment, and multiple valid interpretations of the same scene may exist, particularly at boundaries where transitions between flowing and surrounding water are gradual.

This ambiguity directly shapes how we annotate detection tasks. Bounding boxes derived from segmentation masks tend to enclose large background regions, reducing the signal-to-background ratio. We instead draw boxes that tightly cover high-confidence regions of the rip current rather than the full mask extent, improving the quality of detection targets in scenes where boundaries are uncertain. Oriented bounding boxes provide a more flexible representation that better aligns with the elongated structure of rip currents, but they remain constrained by image boundaries: for rip currents that extend diagonally towards the edge, the box must truncate once a corner reaches the boundary, and may fail to fully cover the spatial extent of such structures.

Because small annotation differences can produce large variations in detection and segmentation metrics, this ambiguity affects both training and evaluation. Both box views ship with the dataset: the manual tight boxes (the official detection target) and \texttt{bbox\_from\_instance}, the axis-aligned bounding rectangles of the instance masks (median IoU between the two sets is $0.82$; the derived box covers $20\%$ more area at the median and $43\%$ or more for a tenth of the instances, most of it being background water). In Appendix~\ref{sec:box_protocol_ablation}, we train the same detector family on both views: neither protocol is systematically easier, so the tight boxes give equally difficult but cleaner supervision. The agreement study in Sec.~\ref{sec:protocol_details} gives the strict-IoU regime its interpretation: two experts exceed IoU $0.75$ on only $52$--$68\%$ of instances. Therefore, the stricter end of the $50{:}95$ range increasingly measures boundary ambiguity rather than model quality, which contextualizes the consistent gap between $AP_{50}$ and $mAP_{50:95}$ in the result tables. By explicitly accounting for this ambiguity in our annotation design, RipBench supports a more realistic assessment of model performance under conditions encountered in real-world deployment.

\vspace{-0.2cm}
\subsection{Licensing and Access}
\vspace{-0.2cm}

RipBench is hosted at \url{https://RipBench.ai}, where the dataset, annotations, evaluation tooling, and baseline code are available. Due to licensing constraints on a subset of the videos, the dataset is released under a \textbf{CC-BY-NC 4.0} license. The inherited RipVIS pool is reused under the same license, and the per-video attribution file records the source of every video.

\vspace{-0.2cm}
\section{Experimental Setup}
\label{sec:experimental_setup}
\vspace{-0.2cm}
\subsection{Models}
\vspace{-0.2cm}
To provide a representative evaluation across tasks, we benchmark a diverse set of modern computer vision models spanning different architectural paradigms and performance regimes. These include convolutional neural networks (CNNs), transformer-based models, and hybrid architectures, as well as both real-time and accuracy-oriented systems. Details on the training strategy and implementation are provided in Appendix \ref{sec:hyperparameter_tuning}.

Where applicable, we include both task-specific models (e.g., dedicated classification or detection networks) and unified frameworks capable of addressing multiple tasks. The released pipeline covers: ResNet-18/34/50/101/152~\cite{he2016deep} and YOLO26~\cite{yolo26_ultralytics} (n--x) for classification (see Table \ref{tab:classification-results}); YOLO26, RT-DETR~\cite{rt-detr} (l, x) and RF-DETR~\cite{rf-detr2025} (n--l) for detection (see Table \ref{tab:box-detection}); YOLO26-OBB for oriented detection (see Table \ref{tab:obb-detection}); YOLO26-seg, RF-DETR-Seg (n--2XL), SparseInst~\cite{sparseinst} (R50, R101, PVTv2-b1), MaskDINO~\cite{mask-dino} and Mask2Former~\cite{mask2former} (R50, Swin-L) for instance segmentation (see Table \ref{tab:inst-segmentation}); YOLO26 semantic heads, MaskDINO and Mask2Former for semantic segmentation (see Table \ref{tab:semantic-segmentation-full}); and MaskDINO and Mask2Former for panoptic segmentation (see Table \ref{tab:pan-segmentation-full}), with an opt-in additional-still scenario for selected classification, axis-aligned detection and instance-segmentation variants. MMDetection~\cite{mmdetection} runs (RTMDet~\cite{rtmdet}, TOOD~\cite{feng2021tood}, Sparse R-CNN~\cite{sparse_rcnn}, DINO~\cite{dino} and Co-DETR~\cite{co-detr} for detection, RTMDet-Ins for instance segmentation) complement these tables. This selection allows us to evaluate not only performance within individual tasks, but also the extent to which model design influences cross-task behavior.

\begin{table}[!t]
  \caption{Frame-level classification on the test split. $F_2$ with its 95\% video-level CI, best in \textcolor{blue}{blue}.}
  \label{tab:classification-results}
  \centering
  \small
  \fittable{\begin{tabular}{lccccccc}
    \toprule
    Model & Acc. & P & R & $F_1$ & $F_2$ [95\% CI] & AUROC & FPS \\
    \midrule
    ResNet-34 @640 & 0.875 & 0.847 & 0.890 & 0.868 & 0.881 [0.77, 0.94] & 0.960 & 108 \\
    ResNet-50 @640 & 0.908 & 0.905 & 0.894 & 0.900 & 0.896 [0.78, 0.96] & 0.966 & 88 \\
    ResNet-101 @640 & 0.900 & 0.908 & 0.871 & 0.889 & 0.878 [0.75, 0.95] & 0.959 & 80 \\
    ResNet-152 @224 & 0.842 & 0.769 & 0.941 & 0.846 & 0.901 [0.78, 0.95] & 0.928 & 108 \\
    YOLO26-n & 0.860 & 0.790 & 0.948 & 0.862 & 0.911 [0.81, 0.96] & 0.960 & 110 \\
    YOLO26-x & 0.935 & 0.925 & 0.935 & 0.930 & \textcolor{blue}{0.933} [0.85, 0.97] & 0.971 & 84 \\
    \bottomrule
  \end{tabular}}
   \vspace{-0.3cm}
\end{table}




\vspace{-0.2cm}
\subsection{Evaluation Protocol}
\label{sec:eval_protocol}
\vspace{-0.2cm}
Evaluation is performed independently for each task on the held-out test split, using the metrics introduced in Sec.~\ref{sec:task_and_metrics}. All metrics are computed on a per-frame basis to ensure consistency. Checkpoint selection follows the framework-specific validation criteria in Appendix~\ref{sec:hyperparameter_tuning}. $F_2$ operating thresholds are selected on validation and frozen before test scoring. Headline operating-point $F_2$ scores carry their $95\%$ video-level bootstrap CIs (Appendix~\ref{sec:uncertainty}), and the number of frames per second (FPS) is measured during inference with one-sample batches on an RTX 4090, at a frozen threshold. Main tables show selected sizes of the leading family per task and representative configurations of other families, 
with the best primary metric in \textcolor{blue}{blue}; the main tables report the canonical scenario (sampled frames only), every configuration of both scenarios is in Appendix~\ref{sec:full_tables}.

For classification, we report standard frame-level metrics (accuracy, precision, recall, $F_1$, $F_2$, AUROC). For detection, oriented detection and instance segmentation, we report AP and $F_{\{1,2\}}$ scores at IoU $0.50$ and averaged over $0.50{:}0.95$ (rotated-polygon IoU for oriented boxes). For semantic segmentation, we report per-class IoU and pixel-level $F_{\{1,2\}}$ scores. For panoptic segmentation, we follow the COCO panoptic protocol with PQ, SQ and RQ at the canonical IoU${>}0.5$ threshold. Classification and instance-level $F_{\{1,2\}}$ scores use an operating point selected \emph{per model on the validation split} (best validation $F_2$ over a dense confidence grid, with IoU${=}0.50$ for instance-level tasks), and applied unchanged to the test split. AP and AUROC are threshold-free. Every number in this setup is computed on split s1.3 with the released evaluators, and every headline table carries a $95\%$ confidence interval from a video-level bootstrap of the $71$ test videos (Appendix~\ref{sec:uncertainty}). Comparisons between models or tasks use paired intervals on the same resampled videos. To assess practical usability, we also report inference speed (FPS), measured with batch size $1$ on a single NVIDIA RTX 4090 with each model running at its frozen validation-selected operating threshold (the threshold of the reported $F_2$), for the models whose deployment path we release.

\begin{table}[!t]
  \caption{Object detection (manual tight boxes) on the test split: AP (threshold-free) and $F_{\{1,2\}}$ scores at IoU $0.50$ and averaged over $0.50{:}0.95$. $F_2$ with its 95\% video-level CI, best in \textcolor{blue}{blue}.}
  \label{tab:box-detection}
  \centering
  \small
  \fittable{\begin{tabular}{lcccccccc}
    \toprule
    Model & $AP_{50}$ & $AP_{75}$ & $mAP_{50:95}$ & ${F_1}_{[50]}$ & ${F_1}_{[50:95]}$ & ${F_2}_{[50]}$ [95\% CI] & ${F_2}_{[50:95]}$ & FPS \\
    \midrule
    Co-DETR-Swin-L$^{\ddagger}$ & 0.683 & 0.150 & 0.268 & 0.601 & 0.303 & 0.692 [0.54, 0.80] & 0.348 & 14 \\
    DINO-Swin-L$^{\ddagger}$ & 0.609 & 0.071 & 0.211 & 0.655 & 0.293 & 0.679 [0.50, 0.80] & 0.303 & 7 \\
    MaskDINO-Swin-L$^{\dagger}$ & 0.588 & 0.174 & 0.257 & 0.638 & 0.352 & 0.612 [0.35, 0.78] & 0.337 & 17 \\
    RF-DETR-n & 0.699 & 0.106 & 0.244 & 0.659 & 0.298 & 0.695 [0.51, 0.81] & 0.314 & 73 \\
    RF-DETR-s & 0.718 & 0.132 & 0.280 & 0.681 & 0.323 & \textcolor{blue}{0.715} [0.56, 0.83] & 0.340 & 70 \\
    RF-DETR-m & 0.678 & 0.169 & 0.283 & 0.683 & 0.343 & 0.669 [0.48, 0.81] & 0.337 & 63 \\
    RF-DETR-l & 0.695 & 0.171 & 0.279 & 0.682 & 0.327 & 0.708 [0.52, 0.84] & 0.339 & 60 \\
    RT-DETR-l & 0.448 & 0.055 & 0.153 & 0.478 & 0.202 & 0.535 [0.35, 0.65] & 0.226 & 74 \\
    RTMDet-l $^{\ddagger}$ & 0.586 & 0.095 & 0.227 & 0.598 & 0.307 & 0.623 [0.41, 0.76] & 0.320 & 103 \\
    Sparse R-CNN-R50$^{\ddagger}$ & 0.568 & 0.070 & 0.205 & 0.559 & 0.262 & 0.599 [0.41, 0.75] & 0.281 & 34 \\
    TOOD-R50$^{\ddagger}$ & 0.528 & 0.101 & 0.216 & 0.494 & 0.254 & 0.543 [0.27, 0.69] & 0.278 & 85 \\
    YOLO26-x & 0.479 & 0.125 & 0.207 & 0.492 & 0.249 & 0.507 [0.27, 0.67] & 0.257 & 164 \\
    \bottomrule
  \end{tabular}}
  \\[2pt] {\footnotesize $^{\dagger}$Instance-segmentation model scored as a detector: its boxes are the bounding rectangles of the predicted masks, evaluated against the manual-box ground truth; FPS is that of the instance model. $^{\ddagger}$Run trained on the predecessor split s1.2 with the additional stills (one flash-rip video fewer in training; validation identical), evaluated here on the s1.3 test split with the released evaluator.}
   \vspace{-0.3cm}
\end{table}

\begin{table}[!t]
  \caption{Oriented detection on the test split (rotated-polygon IoU $0.50$). $F_2$ with its 95\% video-level CI, best in \textcolor{blue}{blue}.}
  \label{tab:obb-detection}
  \centering
  \small
  \fittable{\begin{tabular}{lcccccc}
    \toprule
    Model & $AP_{50}$ & P & R & $F_1$ & $F_2$ [95\% CI] & FPS \\
    \midrule
    YOLO26-n & 0.379 & 0.491 & 0.432 & 0.460 & 0.443 [0.16, 0.64] & 251 \\
    YOLO26-s & 0.329 & 0.439 & 0.401 & 0.419 & 0.408 [0.15, 0.63] & 244 \\
    YOLO26-m & 0.367 & 0.430 & 0.442 & 0.436 & 0.439 [0.18, 0.63] & 229 \\
    YOLO26-l & 0.388 & 0.508 & 0.451 & 0.478 & \textcolor{blue}{0.462} [0.23, 0.66] & 178 \\
    YOLO26-x & 0.376 & 0.455 & 0.441 & 0.448 & 0.444 [0.18, 0.65] & 154 \\
    \bottomrule
  \end{tabular}}
   \vspace{-0.3cm}
\end{table}

\begin{table}[!t]
  \caption{Instance segmentation on the test split: AP (threshold-free) and $F_{\{1,2\}}$ scores at IoU $0.50$ and averaged over $0.50{:}0.95$. $F_2$ with its 95\% video-level CI, best in \textcolor{blue}{blue}.}
  \label{tab:inst-segmentation}
  \centering
  \small
  \fittable{\begin{tabular}{lcccccccc}
    \toprule
    Model & $AP_{50}$ & $AP_{75}$ & $mAP_{50:95}$ & ${F_1}_{[50]}$ & ${F_1}_{[50:95]}$ & ${F_2}_{[50]}$ [95\% CI] & ${F_2}_{[50:95]}$ & FPS \\
    \midrule
    Mask2Former-Swin-L & 0.563 & 0.072 & 0.206 & 0.630 & 0.283 & 0.630 [0.37, 0.79] & 0.283 & 23 \\
    MaskDINO-Swin-L & 0.585 & 0.113 & 0.223 & 0.639 & 0.299 & 0.619 [0.36, 0.79] & 0.290 & 17 \\
    RF-DETR-Seg-n & 0.651 & 0.128 & 0.258 & 0.671 & 0.324 & 0.656 [0.45, 0.81] & 0.317 & 54 \\
    RF-DETR-Seg-s & 0.626 & 0.124 & 0.255 & 0.647 & 0.308 & 0.644 [0.42, 0.79] & 0.306 & 60 \\
    RF-DETR-Seg-m & 0.612 & 0.112 & 0.249 & 0.606 & 0.296 & 0.632 [0.42, 0.77] & 0.308 & 50 \\
    RF-DETR-Seg-l & 0.630 & 0.151 & 0.267 & 0.602 & 0.305 & 0.629 [0.40, 0.78] & 0.319 & 50 \\
    RF-DETR-Seg-x & 0.660 & 0.099 & 0.247 & 0.615 & 0.283 & \textcolor{blue}{0.663} [0.46, 0.79] & 0.305 & -- \\
    RF-DETR-Seg-2xl & 0.654 & 0.087 & 0.254 & 0.643 & 0.304 & 0.661 [0.45, 0.79] & 0.312 & -- \\
    RTMDet-Ins-x$^{\ddagger}$ & 0.666 & 0.040 & 0.224 & 0.622 & 0.256 & 0.662 [0.45, 0.79] & 0.273 & 60 \\
    SparseInst-R101-GIAM & 0.585 & 0.019 & 0.179 & 0.608 & 0.228 & 0.600 [0.35, 0.76] & 0.225 & 44 \\
    YOLO26-x & 0.553 & 0.164 & 0.249 & 0.534 & 0.267 & 0.582 [0.27, 0.75] & 0.291 & 129 \\
    \bottomrule
  \end{tabular}}
  \\[2pt] {\footnotesize $^{\ddagger}$Run trained on the predecessor split s1.2 with the additional stills (one flash-rip video fewer in training; validation identical), evaluated here on the s1.3 test split with the released evaluator.}
   \vspace{-0.3cm}
\end{table}

\vspace{-0.2cm}
\section{Results and Analysis}
\vspace{-0.2cm}
\subsection{Performance Across Tasks}
\label{sec:results_tasks}
\vspace{-0.2cm}

We compare performance across all supported tasks, with emphasis on $F_2$ and with its $95\%$ video-level confidence interval in brackets. Frame-level presence detection is the easiest task (Table~\ref{tab:classification-results}), as the best classifier (YOLO26-x) reaches $F_2{=}0.933$ and AUROC${=}0.97$, while most configurations exceed $0.88$ and $0.92$ in terms of $F_2$ and AUROC, respectively. Localization is substantially harder. Detection (Table~\ref{tab:box-detection}) tops out at $F_2{=}0.715$ (RF-DETR-s), with the four RF-DETR sizes between $0.67$ and $0.72$. The wide intervals reflect the structure of the data, such that models may even succeed or fail on whole videos. Oriented detection (Table~\ref{tab:obb-detection}) lands between $0.41$ and $0.46$ under rotated IoU, which we attribute to the fact that rip currents do not necessarily have a canonical orientation. Instance segmentation (Table~\ref{tab:inst-segmentation}) is led by RF-DETR-Seg ($F_2{=}0.663$ for the XL variant and $0.661$ for 2XL, with the nano at $0.656$ and small at $0.644$). Semantic segmentation (Table~\ref{tab:semantic-segmentation-full}) reaches a rip-class IoU of only $0.34$--$0.41$ (best $0.41$). As pixel metrics can be dominated by a few videos with large rips, there is a wide interval, while the contextual classes are near-perfect (background mIoU $0.96$--$0.97$). Panoptic segmentation (Table~\ref{tab:pan-segmentation-full}) shows the same picture: the rip class reaches a panoptic quality of $0.40$--$0.47$, dominated by Mask2Former-Swin-L $0.465$ and MaskDINO-Swin-L $0.459$. The Swin-L backbone improves on Mask2Former-R50 with a paired interval that excludes zero, while the two Swin-L models are not separable and swap order between validation and test. The three context classes exceed $0.94$, so the rip class is the dominant source of error.

RipBench also supports video instance segmentation (Table~\ref{tab:video_main}), and the temporal baselines already come close to the best per-frame model: RF-DETR-Seg-x leads with $F_2{=}0.663$, with CAVIS-Swin-L + refiner ($0.630$) and MinVIS-R50 ($0.593$) inside its confidence interval, followed by Video-Mask2Former-R50 ($0.568$) and the DVIS++ segmenter stage ($0.517$). Temporal context pays off within a matched backbone, as CAVIS-Swin-L raises $AP_{50}$ from $0.563$ to $0.612$ over its Mask2Former image counterpart (Appendix~\ref{sec:video}); since the video models start from Mask2Former initializations that score $0.616$ (R50) and $0.630$ (Swin-L) per frame, temporal modeling on top of stronger per-frame models is a direct next step that the released track annotations and dense views make straightforward.

The paired analysis makes the cross-task ordering precise (Table~\ref{tab:paired_gaps}): the validation-selected classifier exceeds the validation-selected detector by $+0.18$ $F_2$ and the validation-selected instance segmenter by $+0.24$. Pairing the best test configurations instead, optimistic by construction, gives $+0.22$ and $+0.27$. These gaps never approach zero under resampling of the test videos and exceed four times the largest seed standard deviation of the baselines ($0.040$), as detailed in Appendix~\ref{sec:multiseed}. Within the RF-DETR lineage, the validation-selected detector in turn edges out the validation-selected instance segmenter by $0.07$ (best test pairing: $0.05$). In summary, the DINOv2-based RF-DETR leads both localization tasks, with detection variants slightly ahead of segmentation variants.

We caution that cross-task $F_2$ comparisons reflect both task difficulty and differences in annotation targets: detection boxes are tighter than the underlying segmentation masks (see Sec.~\ref{sec:bbox_vs_mask}), and OBB targets are constrained by image boundaries. In Appendix~\ref{sec:box_protocol_ablation}, we bound the target effect: training and evaluating the same detector family on the tight or the mask-derived boxes shifts $F_2$ by at most $0.07$ with no consistent winner, and all ten variants land in one band on a shared target. Thus, the gap of roughly $0.2$--$0.3$ $F_2$ between presence detection and spatial localization reflects task difficulty rather than target design. In summary, the numbers characterize what each task asks the model to predict, and localization is a substantively harder problem than presence detection (a.k.a.~classification).

\begin{table}[!t]
  \caption{Semantic segmentation on the test split: per-class IoU and pixel-level $F$-scores (rip class and mean over water/land/sky); rip IoU with its 95\% video-level CI, best in \textcolor{blue}{blue}.}
  \label{tab:semantic-segmentation-full}
  \centering
  \small
  \fittable{\begin{tabular}{lcccccccc}
    \toprule
    Model & $IoU_{rip}$ [95\% CI] & $mIoU_{bg}$ & ${F_1}_{rip}$ & ${F_2}_{rip}$ & ${F_1}_{bg}$ & ${F_2}_{bg}$ & Pixel acc. & FPS \\
    \midrule
    Mask2Former-R50 & 0.376 [0.18, 0.50] & 0.967 & 0.547 & 0.506 & 0.983 & 0.984 & 0.969 & 31 \\
    MaskDINO-R50 & 0.397 [0.19, 0.52] & 0.969 & 0.568 & 0.519 & 0.984 & 0.986 & 0.972 & 31 \\
    YOLO26-n & \textcolor{blue}{0.406} [0.16, 0.55] & 0.955 & 0.577 & 0.508 & 0.977 & 0.978 & 0.964 & 423 \\
    YOLO26-s & 0.342 [0.13, 0.48] & 0.958 & 0.510 & 0.436 & 0.978 & 0.980 & 0.965 & 426 \\
    YOLO26-m & 0.379 [0.15, 0.51] & 0.965 & 0.550 & 0.468 & 0.982 & 0.984 & 0.970 & 362 \\
    YOLO26-l & 0.395 [0.18, 0.53] & 0.964 & 0.567 & 0.501 & 0.982 & 0.983 & 0.969 & 265 \\
    YOLO26-x & 0.387 [0.17, 0.51] & 0.960 & 0.558 & 0.481 & 0.980 & 0.981 & 0.967 & 216 \\
    \bottomrule
  \end{tabular}}
  \vspace{-0.3cm}
\end{table}

\begin{table}[!t]
  \caption{Panoptic segmentation on the test split (COCO panoptic protocol; rip current is the only thing class, so $PQ_{rip}=PQ_{th}$; background = water/land/sky); $PQ_{rip}$ with its 95\% video-level CI, best in \textcolor{blue}{blue}.}
  \label{tab:pan-segmentation-full}
  \centering
  \small
  \fittable{\begin{tabular}{lccccccc}
    \toprule
    Model & $PQ_{rip}$ [95\% CI] & $SQ_{rip}$ & $RQ_{rip}$ & $PQ_{bg}$ & $RQ_{bg}$ & $PQ$ & FPS \\
    \midrule
    Mask2Former-R50 & 0.396 [0.21, 0.52] & 0.696 & 0.568 & 0.943 & 0.990 & 0.806 & 41 \\
    Mask2Former-Swin-L & \textcolor{blue}{0.465} [0.27, 0.58] & 0.695 & 0.670 & 0.946 & 0.985 & 0.826 & 23 \\
    MaskDINO-R50 & 0.433 [0.18, 0.58] & 0.709 & 0.611 & 0.954 & 0.996 & 0.824 & 30 \\
    MaskDINO-Swin-L & 0.459 [0.25, 0.58] & 0.695 & 0.661 & 0.947 & 0.985 & 0.825 & 17 \\
    \bottomrule
  \end{tabular}}
  \vspace{-0.3cm}
\end{table}

\begin{table}[!t]
  \caption{Video instance segmentation on the test split (per-frame protocol): best RipBench-trained configuration per family by reported $F_2$, best in \textcolor{blue}{blue}; every configuration in Appendix~\ref{sec:video}.}
  \label{tab:video_main}
  \centering
  \small
  \fittable{\begin{tabular}{llccc}
    \toprule
    Family & Configuration & ${F_2}_{[50]}$ [95\% CI] & $AP_{50}$ & $mAP_{50:95}$ \\
    \midrule
    Per-frame image model & RF-DETR-Seg-x & \textcolor{blue}{0.663} [0.46, 0.79] & 0.660 & 0.247 \\
    MinVIS & MinVIS-R50 (dense-interp view) & 0.593 [0.29, 0.78] & 0.598 & 0.240 \\
    CAVIS & CAVIS-Swin-L + offline refiner & 0.630 [0.36, 0.81] & 0.610 & 0.229 \\
    Video-Mask2Former & Video-Mask2Former-R50 & 0.568 [0.31, 0.72] & 0.580 & 0.241 \\
    DVIS++ (segmenter stage) & DVIS++-R50 & 0.517 [0.31, 0.65] & 0.464 & 0.180 \\
    \bottomrule
  \end{tabular}}
  \vspace{-0.3cm}
\end{table}

\vspace{-0.2cm}
\subsection{Task-Dependent Behavior}
\vspace{-0.2cm}

Model behavior varies across tasks, and architecture choices that win on one task do not necessarily transfer. YOLO26-x is the strongest classifier but sits in the middle of the detection and instance segmentation tables. RF-DETR leads detection at its small size ($0.715$), ahead of RT-DETR-l ($0.535$) and every YOLO26 size ($0.42$--$0.51$), while RT-DETR-x trails the three larger YOLO26 sizes ($0.470$ vs.\ $0.50$--$0.51$; Appendix~\ref{sec:full_tables}); and the DINOv2-based RF-DETR-Seg leads instance segmentation, its six sizes spanning $0.629$--$0.663$. Different tasks therefore reward different inductive biases, and benchmarking models on a single task provides an incomplete view of how an architecture performs across the full pipeline of rip current understanding. Two further regularities are robust to resampling. First, threshold-free metrics and operating-point metrics disagree on rankings: below RF-DETR-s, which leads detection on both metrics, RT-DETR-l has the higher $F_2$ but a lower $AP_{50}$ than YOLO26-x ($0.448$ vs.\ $0.479$), i.e., recall-oriented deployment favours a different model than the ranking metric would. Second, the opt-in \texttt{additional\_data} stills (Appendix~\ref{sec:hyperparameter_tuning}) help the transformer families most (RT-DETR-l $0.535 \rightarrow 0.596$, RF-DETR-Seg-s $0.644 \rightarrow 0.696$, and RF-DETR-Seg-l reaches $0.718$ with the stills, the highest instance-segmentation $F_2$ in the result tables), so the canonical tables understate what these architectures reach with more supervision. From a deployment perspective, this argues for matching the architecture to the task: lighter convolutional models for fast presence detection at scale, and DETR-style detection-segmentation models when spatial extent or per-instance localization is required.

\vspace{-0.2cm}
\subsection{Failure Cases}
\label{sec:failure_modes}
\vspace{-0.2cm}
Common failure modes include breaking waves, foam patterns, and dark water channels that visually resemble rip currents, particularly at low resolution or oblique viewpoints. Detection models tend to include large background regions such as shoreline or surrounding water, while oriented boxes are constrained by image boundaries when rip currents extend diagonally toward the edge. Segmentation models produce fragmented predictions, incomplete spatial coverage, or unstable boundaries, reflecting the amorphous nature of the phenomenon. The per-video metadata makes these failures quantifiable (Appendix~\ref{sec:stratified}). Presence detection holds across all four viewpoints (best classifier $F_2$ $0.92$--$1.00$, the upper end on the five-video water-level stratum), while localization depends on viewpoint and model family: YOLO26 detection spans $0.58$--$0.68$ on overhead footage and $0.11$--$0.37$ at water level, whereas RF-DETR-l peaks at water level ($0.80$). Dark-gap rips score $F_2$ $0.83$ for detection and $0.79$ for instance segmentation; sediment plumes are harder for both ($0.54$ / $0.52$), and the single foam-plume video separates the tasks ($0.88$ vs.\ $0.30$). Expert agreement is viewpoint-invariant (Section~\ref{sec:protocol_details}), so these are model limitations, not annotation artifacts. The starkest case is RipBench-125, a mushroom-shaped sediment plume filmed from a bird's-eye view at $608\times1080$: $14$ of the $20$ canonical classifiers score all $150$ frames correctly, yet six of the sixteen YOLO26/RT-DETR detection rows reach $F_2 = 0$ on the video (RT-DETR-l: $0$ true positives, $100$ false positives) and none of them exceeds $0.30$, whereas the RF-DETR detectors localize it ($F_2$ $0.63$--$0.82$); among instance segmenters only RF-DETR-Seg-n localizes it acceptably ($F_2$ $0.55$; next best MaskDINO-Swin-L at $0.40$): the models know a rip is there without being able to say where.

\vspace{-0.2cm}
\section{Discussion, Impact, and Limitations}
\label{sec:discussion_and_limitations}
\vspace{-0.2cm}

\textbf{Safety-critical use and human oversight.}
RipBench emphasizes $F_2$ in evaluation and model selection because, in beach safety applications, missed rip currents carry far higher cost than false alarms. Even so, high-performing models may fail in visually ambiguous or rare scenarios. We envision RipBench as a foundation for expert-assisted monitoring systems, where automated predictions support lifeguards and coastal safety personnel rather than replace them.

\textbf{Dataset coverage and generalization.}
Although RipBench contains substantial geographic and viewpoint diversity, several scenarios remain underrepresented. Nighttime footage, fog, rain, and low-visibility conditions are rare due to the difficulty of collecting and annotating such data. The dataset also reflects biases in footage availability and beach accessibility, which may limit generalization to underrepresented coastlines. Rip currents visible through strong wave breaking patterns are also generally easier to detect than those expressed primarily through subtle sediment transport or directional flow, and our benchmark numbers should be read with this distribution in mind. Site identifiers are not yet part of the per-video attribution file, so the released splits should be read as within-pool generalization. 

\textbf{Annotation and evaluation under ambiguity.}
As discussed in Section~\ref{sec:bbox_vs_mask}, rip current boundaries are inherently diffuse, and no single task formulation fully captures their structure. Cross-task comparisons therefore reflect both task difficulty and differences in annotation targets. Annotations represent expert video interpretations. Several videos do contain a dye release, but dye was not used as an annotation criterion and stops following the current after a short time. No drift or current-meter measurements accompany the footage. The labels therefore record what trained annotators can identify from imagery, which is also the setting a deployed detector operates in. The blind cross-annotation study in Appendix~\ref{sec:iaa_appendix} bounds their consistency. While we have been explicit about these design choices, they remain a fundamental limitation of evaluating amorphous phenomena with standard vision benchmarks.

\textbf{Deployment beyond benchmark performance.}
Field deployment requires substantially more than the metrics reported here. Robustness under changing weather, unseen coastlines, camera motion, and compression artifacts remains an open challenge, and operational testing is essential before deployment in public safety settings.

\textbf{Future directions.}
The video views and temporal baselines (Table~\ref{tab:video_main}; every configuration and view in Appendix~\ref{sec:video}) are a first step toward video-based rip current understanding. The stratified analysis (Appendix~\ref{sec:stratified}) points at the deployment perspectives and visual signatures where localization fails. The annotation-propagation route outlined in Appendix~\ref{sec:video} is how we intend to densify supervision with expert verification. We plan to extend RipBench towards lightweight edge deployment and broader environmental coverage, with continued expansion through public challenges hosted at \url{https://RipBench.ai}.

{
\small
\vspace{-0.2cm}
\section*{Acknowledgments and Disclosures}
\vspace{-0.2cm}
\textbf{Funding.} This work was partially supported by the Alexander von Humboldt Foundation and represents the culmination of a five-year collaboration between the University of W\"urzburg and the University of Bucharest. This research is supported by the project ``Romanian Hub for Artificial Intelligence - HRIA'', Smart Growth, Digitization and Financial Instruments Program, 2021-2027, MySMIS no.~351416. 

\textbf{Competing interests.} The authors declare no competing interests.

\textbf{Data and code.} The dataset, annotations, evaluation tooling and baseline recipes are released
under CC BY-NC 4.0 with a Croissant metadata record; see the ``Licensing and Access'' subsection and the
dataset landing page.

\bibliographystyle{plain} 

\bibliography{bibliography}

}


\clearpage

\appendix

\section{Technical appendices and supplementary material}
The supplementary material provides expanded details on the RipBench dataset, benchmark and evaluation as well as motivation and context. While the main paper presents the core contribution, this document includes additional insights on data diversity, annotation protocols, evaluation metrics, rip currents research and risk and training configuration across all tasks.

We begin in Section~\ref{sec:motivation} by describing the motivation for RipBench. Section~\ref{sec:global_studies} expands on a common misconception observed across cultures: people often believe they can accurately identify rip currents, whereas prior research shows that most cannot. Section~\ref{sec:metrics} then defines the evaluation metrics, followed by Section~\ref{sec:hyperparameter_tuning}, which describes the evaluated models, training protocols, and hyperparameter settings.  We conclude with Section~\ref{sec:dataset_summary}, showing several tables showing examples from the dataset across all tasks, including a final table that overlays axis-aligned detection, oriented detection, and segmentation annotations to directly compare the different rip-current boundary representations.

\section{Motivation and Need for RipBench}
\label{sec:motivation}

Rip current research has historically been fragmented across tasks, datasets, and evaluation protocols. Prior work has approached the problem from divergent angles, including bounding box detection, optical flow, and instance segmentation, typically on different datasets with minimal overlap. Evaluation protocols are often underspecified or inconsistent: accuracy is computed differently across papers, IoU and confidence thresholds are reported inconsistently, and frame-level versus video-level aggregation is rarely stated explicitly. This makes it difficult to compare methods or to assess what level of vision-based rip current understanding has actually been achieved.

Existing public datasets compound this fragmentation. They are typically restricted to a single task formulation (Table~\ref{tab:rip_dataset_comparison}), and class balance is rarely treated as a design parameter. Most datasets are dominated by rip-positive frames, which makes false-positive evaluation unreliable and biases classification experiments toward trivially high accuracy. RipVIS~\cite{dumitriu2025ripvis} established the first large-scale video benchmark for rip currents but focused solely on instance segmentation and inherited an imbalanced rip / non-rip distribution.

RipBench addresses these limitations directly. We integrate videos from most existing public datasets, add newly collected footage, and balance the dataset to $152$ rip-positive and $152$ non-rip videos. This is a deliberate choice motivated in Section~\ref{sec:collection_diversity}, since the negative class for rip currents is not heterogeneous background but the same water under the same conditions. Annotations are provided for all six task formulations (classification, axis-aligned and oriented detection, instance, semantic, and panoptic segmentation) over the same frames, with semantic and panoptic masks covering all videos, rip and non-rip alike, so models trained on RipBench learn full-scene structure rather than rip-only foreground. We provide transparent metric definitions in Section~\ref{sec:metrics} and standardized splits across all tasks, enabling reproducible and directly comparable results.

\section{Global Studies on Rip Current Identification}
\label{sec:global_studies}
A common misconception is that rip currents are easy to spot. Experiments replicated worldwide show that most beachgoers think they can spot a rip current but are unable to do so. In a study of domestic and international students’ knowledge on beaches in Queensland, Australia, Ballantyne~\etal(2005) \cite{ballantyne2005between} found that two-thirds who claimed to know what a rip current was were unable to recognize one. Furthermore, when shown a photograph of a beach with two beach rip currents, $61\%$ of all students selected the rip currents (dark gaps between white water) as places they would swim. Similarly, Sherker~\etal(2010)\cite{sherker2010beachgoers} found that $93\%$ of surveyed Australian beachgoers thought they could identify a rip current, but only $33\%$ could when shown a picture of one. Half the survey respondents selected the rip current as the safest place to swim.

Other studies in the United States have shown beachgoer rip identification is lower than in Australia. Caldwell~\etal(2013) \cite{caldwell2013ability} conducted surveys of beachgoers at Pensacola Beach, Florida, using pictures. While the majority of respondents said they could identify rip currents, less than $20\%$ were able to do so. On three popular beaches in Texas, Brannstrom~\etal(2014) \cite{brannstrom2014perception} showed beachgoers images of a groyne under different wave conditions: only $13\%$ picked the image containing the boundary rip current as the most hazardous to swim in, the majority choosing images of breaking waves instead, and $31\%$ succeeded when asked to identify the boundary-controlled rip current running along the groyne~\cite{pitman2021beachgoers}.

Woodward~\etal(2015) \cite{woodward2015public} found that in the United Kingdom, out of $407$ surveyed beachgoers, only $10\%$ gave a good or excellent answer when surveyed about rip currents, $25\%$ had partial knowledge, and over $65\%$ of them had poor knowledge of rip currents. Sotes~\etal(2018) \cite{sotes2018biscayne} also highlighted the low level of rip current understanding among the respondents. Out of $364$ beachgoers, only $25$ of them could recognize the rips’ existence, while $18$ of them could point out the correct direction to escape from the rips.

In their study comparing beachgoers' ability to identify rip currents in photographs and in situ at a beach in New Zealand (Muriwai Beach), where a channel rip current was present, Pitman~\etal(2021) \cite{pitman2021beachgoers} showed that only $22\%$ of respondents were able to identify the in situ rip current.

A preliminary study done in Malaysia by Mohammed Isa~\etal(2021) \cite{mohammed2021preliminary} found that, while $83\%$ of respondents were aware of currents in general, only $26\%$ were aware of the existence of rip currents, and only $20\%$ could accurately define one. $104$ questionnaires filled in by tourists in July and August 2024 along two of the most crowded beaches on the Romanian Black Sea coast (Eforie Nord and Costinesti) revealed that only $27\%$ of the respondents were able to correctly identify rip current occurrence.

 The dangers of rip currents are increased by this lack of knowledge and understanding from beachgoers worldwide. In the United States, a study done between 1994 and 2007 concluded that, on average, $35$ people die from rip currents each year \cite{gensini2010examination}. A later 2019 study concluded that more than $100$ rip current fatalities occur each year \cite{brewster2019estimations}, a steep increase from the previous estimation. In Australia, rip currents cause more deaths each year than bushfires, floods and cyclones combined \cite{brander2013brief, brighton2013rip, australia2019national}. In Japan, approximately $300$ cases of rip currents have been reported between 2003 and 2011 \cite{ishikawa2014characteristics}. In India, approximately $39$ people die of rip currents each year \cite{arun2014rip}. At Dadonghai Beach in China, over 350 drowning incidents were recorded whereby most are linked to rip currents \cite{zhang2021rip}. In Costa Rica, out of $1,391$ drownings between 2001 and 2012, approximately $590$ of them have been the result of rip currents \cite{arozarena2015rip}. On the Black Sea, it is estimated that $33$ people die from rip currents each year on the Istanbul beaches alone \cite{barlas2016rip}. On the Mediterranean Sea, in Liguria, a video monitoring system concluded that $13$ fatalities occurred from rip currents in 2014 \cite{mucerino2021rip}.

 While subsequent educational efforts to increase awareness have been shown to increase the knowledge of beachgoers on how to escape the grip of a rip current, they are still ineffective in helping them identify the rip currents correctly \cite{houser2017public}.
 
\section{Evaluation Metrics}
\label{sec:metrics}

We evaluate RipBench using task-specific metrics that reflect both localization quality and the safety-critical need to detect rip currents reliably. For object-level tasks, including object detection, oriented object detection, and instance segmentation, we report $AP_{[50]}$, $mAP_{[50:95]}$, ${F_1}_{[50]}$, ${F_1}_{[50:95]}$, ${F_2}_{[50]}$, and ${F_2}_{[50:95]}$. These metrics are computed for the rip-current class only, ensuring consistent comparison across tasks.

For semantic segmentation, we report pixel-level metrics for both the rip-current class and the background classes. Specifically, we report IoU and mIoU for the rip-current and background categories, together with class-wise $F_1$ and $F_2$ scores to capture both overlap quality and recall-oriented segmentation performance.

For panoptic segmentation, we report panoptic and pixel-level metrics separately for rip currents and background regions. These include $PQ_{\text{rip}}$ and $RQ_{\text{rip}}$ for rip-current panoptic quality and recognition quality, $PQ_{\text{bg}}$ and $RQ_{\text{bg}}$ for background regions, $IoU_{\text{rip}}$ for rip-current mask overlap, $mIoU_{\text{bg}}$ for the background classes, and ${F_2}_{\text{rip}}$ to emphasize recall for rip-current regions.

For frame-level classification, we report classification metrics such as accuracy, precision, recall, $F_1$, and $F_2$. 

For all detection-style metrics, a prediction is considered a true positive when it matches a ground-truth instance with sufficient spatial overlap according to the relevant Intersection-over-Union (IoU) threshold. For object detection, IoU is computed between predicted and ground-truth bounding boxes. For oriented object detection, IoU is computed between predicted and ground-truth oriented bounding boxes. For instance segmentation, IoU is computed between predicted and ground-truth masks.

\subsection{Precision and Recall}

Given the number of true positives ($\text{TP}$), false positives ($\text{FP}$), and false negatives ($\text{FN}$), we compute:

\begin{equation}
\text{Precision} = \frac{\text{TP}}{\text{TP} + \text{FP}} 
\hspace{2cm}
\text{Recall} = \frac{\text{TP}}{\text{TP} + \text{FN}}.
\label{eq:precision_recall}
\end{equation}

Precision measures how many predicted rip-current detections are correct, while recall measures how many ground-truth rip currents are detected. In rip-current monitoring, recall is particularly important because false negatives correspond to missed hazardous regions.

\subsection[F-beta, F1, and F2 Scores]{$\mathbf{F_\beta}$, $\mathbf{F_1}$, and $\mathbf{F_2}$ Scores}

The $F_\beta$ score is a weighted harmonic mean of precision and recall, where $\beta$ controls the relative emphasis on recall:

\begin{equation}
F_\beta = (1 + \beta^2) \cdot 
\frac{\text{Precision} \cdot \text{Recall}}
{\beta^2 \cdot \text{Precision} + \text{Recall}}.
\label{eq:fbeta}
\end{equation}

We report two special cases:

\begin{equation}
F_1 = 2 \cdot 
\frac{\text{Precision} \cdot \text{Recall}}
{\text{Precision} + \text{Recall}}
\hspace{1.5cm}
F_2 = 5 \cdot 
\frac{\text{Precision} \cdot \text{Recall}}
{4 \cdot \text{Precision} + \text{Recall}}.
\label{eq:f1_f2}
\end{equation}

The $F_1$ score balances precision and recall equally, while $F_2$ places greater emphasis on recall. We include $F_2$ because missing a rip current is more critical than producing an occasional false positive in safety-oriented monitoring scenarios.

For object detection, oriented object detection, and instance segmentation, we report $F_1$ and $F_2$ under two IoU settings. ${F_1}_{[50]}$ and ${F_2}_{[50]}$ are computed using matches at IoU $\geq 0.5$. We also report ${F_1}_{[50:95]}$ and ${F_2}_{[50:95]}$, where the scores are averaged over IoU thresholds from 0.50 to 0.95 in steps of 0.05. This provides both a permissive evaluation at IoU 0.5 and a stricter measure of localization quality across multiple overlap thresholds.

\subsection{Average Precision}

For object-level tasks, we report Average Precision at IoU 0.5 and mean Average Precision over multiple IoU thresholds:

\begin{equation}
\text{IoU} = \frac{\text{Area of Overlap}}{\text{Area of Union}}.
\label{eq:iou}
\end{equation}

$AP_{[50]}$ evaluates average precision when predictions are matched to ground truth at IoU $\geq 0.5$:

\begin{equation}
AP_{[50]} = \int_0^1 \text{Precision}(r) \, dr
\quad \text{at IoU} = 0.5.
\label{eq:ap50}
\end{equation}

We also report $mAP_{[50:95]}$, which averages AP over IoU thresholds from 0.50 to 0.95 in steps of 0.05:

\begin{equation}
mAP_{[50:95]} =
\frac{1}{10}
\sum_{t \in \{0.50, 0.55, \ldots, 0.95\}} AP_{[t]}.
\label{eq:map5095}
\end{equation}

While $AP_{[50]}$ captures whether a model can detect rip currents with reasonable localization, $mAP_{[50:95]}$ provides a stricter assessment of localization precision. Reporting both is useful because rip currents are amorphous, low-contrast phenomena whose boundaries may be less sharply defined than conventional foreground objects.

\subsection{Confidence-Threshold Selection}

For reporting test-set $F_1$ and $F_2$ scores in object detection, oriented object detection, and instance segmentation, a final confidence threshold is required to decide which model predictions are retained. Each model outputs a confidence score for every predicted box or mask, and predictions below the selected threshold are discarded before matching predictions to ground truth. We select this threshold separately for each trained model variant rather than using a single global value. Specifically, multiple confidence thresholds are evaluated on the validation set, and the threshold that maximizes validation ${F_2}_{[50]}$ is selected. This selected threshold is then fixed and used to compute the reported test-set ${F_1}_{[50]}$, ${F_1}_{[50:95]}$, ${F_2}_{[50]}$, and ${F_2}_{[50:95]}$ scores.

Average Precision metrics do not require selecting a single confidence threshold, since AP is computed from the precision--recall curve obtained by ranking predictions according to confidence.

\subsection{Frames Per Second (FPS)}

To assess runtime efficiency, we report inference-only FPS:

\begin{equation}
\text{FPS} = \frac{\text{Total Number of Frames}}{\text{Total Inference Time}}
\label{eq:fps}
\end{equation}

All FPS measurements in this version are taken on an NVIDIA RTX 4090 GPU with a batch size of 1, with each model running at its frozen validation-selected operating threshold so that post-processing (NMS, mask decoding) reflects deployment rather than the $0.001$ confidence floor used to dump predictions for the metrics (the submitted version used an RTX 3060 and timed models at that floor), simulating real-time deployment on streaming video. This can greatly impact the design of real-time detection systems, where models must balance detection accuracy with strict latency requirements. In safety-critical applications like rip current monitoring, high-speed inference enables faster alerts and broader camera coverage, but must not come at the cost of degraded recall or localization precision.

\section{Training Protocol}
\label{sec:hyperparameter_tuning}

\paragraph{Scope of the reported tables.} All models in the main tables were trained with the released cluster pipeline (\texttt{train\_cluster/}: one script per variant, $300$ epochs with early-stopping patience $50$ for the Ultralytics families, the documented shorter schedules for the detectron2 and RF-DETR families) on NVIDIA L40 GPUs, and evaluated with the released evaluators; the canonical training scenario uses the sampled frames only, and the \texttt{additional\_data} scenario adds the $3{,}773$ standalone stills (opt-in, reported in the appendix tables). 


Hyperparameter settings and data augmentation pipelines were selected separately for each model family, following the corresponding official recipes whenever applicable. For the ultralytics families we avoided random cropping beyond the framework's default pipeline; the detectron2 families keep their official large-scale-jittering recipes, which include crops, because their fine-tuning proved brittle without them. Since rip currents are visually defined by subtle background-like water patterns rather than by compact foreground objects, random crops can remove the surrounding wave, shoreline, and flow context needed to identify the rip current. In some cases, a crop may retain only an ambiguous internal region of the rip current, making the visual evidence less representative of the full instance. Therefore, we relied primarily on resizing, flipping, padding, color augmentation, and model-specific augmentation strategies.


\subsection{Released training recipes}
\label{sec:released_recipes}
All rows in the result tables come from one script per variant in \texttt{train\_cluster/} (constants at the top, no command-line arguments; one NVIDIA L40 per run on a SLURM cluster, 24-hour segments resumed from the last checkpoint), followed by prediction dumps and the released evaluators. Checkpoint selection uses each framework's validation fitness (ultralytics: mAP$_{50:95}$ for detection, oriented detection and instance segmentation, top-1 accuracy for classification, mIoU for semantic segmentation; detectron2 families: validation segm AP, PQ or mIoU, evaluated every one or two epochs with a best-checkpoint hook; RF-DETR: EMA segm mAP); the $F_2$ operating point is selected afterwards on the validation split by the evaluator and frozen. Every variant has a canonical run; selected classification, axis-aligned detection and instance-segmentation variants also use the $3{,}773$ additional stills (appendix tables).

\paragraph{Ultralytics families (YOLO26, RT-DETR, ResNet).} $300$ epochs with early-stopping patience $50$, input size $640$ (classification additionally at $224$), the framework's default augmentation suite (mosaic, translation and scale jittering, flipping, HSV perturbation; no random cropping beyond it) and default hyper-parameters. YOLO26 detection and instance-segmentation models start from the COCO-pretrained checkpoints~\cite{coco}; YOLO26 classification and oriented-detection models are trained from scratch because no pretrained YOLO26 checkpoints exist for those heads; the YOLO26 semantic heads start from the released Cityscapes \cite{cordts2016cityscapes} checkpoints. Batch size is chosen automatically per model (fixed at $16$ for RT-DETR, whose trainer has no automatic batch sizing, and $128$/$256$ for the classifiers at $640$/$224$). ResNet-18 to ResNet-152 are ImageNet-pretrained torchvision backbones wrapped in the same trainer, with a shared trainer but different architectures and initialization.

\paragraph{RF-DETR-Seg.} COCO-pretrained RF-DETR-Seg (Nano to 2XLarge), fine-tuned with the official recipe: AdamW, learning rate $10^{-4}$, effective batch size $64$ on every size (per-size batch $\times$ gradient accumulation), $60$ epochs with early stopping (patience $10$) on the EMA validation segm mAP, at most $50$ detections per image at evaluation. The $60$-epoch budget is a documented deviation from the $300/50$ schedule of the ultralytics families: one epoch costs $5$--$12$ hours on an L40 while validation AP plateaus after roughly ten epochs.

\paragraph{SparseInst.} COCO-pretrained SparseInst (R50-base, R50-GIAM with augmentation, R101-GIAM with and without augmentation, PVTv2-b1-GIAM with and without augmentation), AdamW with learning rate $5\times10^{-5}$, batch $16$, $400$ warm-up iterations, $12$ epochs with full validation every epoch, at most $20$ detections per image; negatives are kept (the crop augmentation's box recomputation is disabled so that images without instances remain valid).

\paragraph{MaskDINO and Mask2Former.} COCO-pretrained (instance, panoptic) and ADE20K-pretrained (semantic) checkpoints with R50 and Swin-L backbones, fine-tuned with the official detectron2 recipes: AdamW, learning rate $5\times10^{-5}$, batch $4$ (instance, panoptic; large-scale jittering at $640$) or $8$ (semantic; $512$ crops), $12$ epochs with full validation every two epochs. Panoptic and semantic runs track two selection criteria (overall PQ or mIoU, and the rip-class PQ or IoU); the rip-class selection is reported, because the overall scores are dominated by the three near-saturated context classes. Mixed precision is disabled for the semantic runs (numerical instability in the matcher), and inference filters queries before the full-resolution upsampling so that 4K frames fit a 24\,GB GPU (bit-identical outputs).

\section{Uncertainty of the Benchmark Numbers}
\label{sec:uncertainty}
All headline tables in the main paper report $95\%$ confidence intervals from a non-parametric bootstrap over the $71$ test videos ($B=1{,}000$ resamples with replacement, fixed seed): videos, not frames, are resampled because frames within a video are strongly correlated and models tend to succeed or fail on whole videos, so the $71$ videos are the effective sample size. Intervals are computed at each model's frozen validation-selected operating point. Table~\ref{tab:paired_gaps} reports paired differences, in which both configurations are scored on the same resampled videos so that shared video difficulty cancels; a paired interval that excludes zero is the criterion we use for every claim that compares two configurations.

\begin{table}[!htbp]
  \caption{Paired video-level bootstrap differences in test $F_2$ between representatives of two tasks, both scored on the same resampled videos ($B=1{,}000$). Rows marked val-selected pair the configuration with the highest validation metric of each task (the primary comparison); the unmarked rows pair the configurations with the highest test $F_2$ and are optimistic by construction. The classification-to-localization gap never approaches zero; detection is slightly ahead of instance segmentation in both pairings (both intervals exclude zero). Representatives were frozen on the configuration set of 5 September 2026; RF-DETR-Seg-XL, which finished later and reaches $0.663$, is listed in the tables but does not re-select a representative.}
  \label{tab:paired_gaps}
  \centering
  \small
  \fittable{\begin{tabular}{llcc}
    \toprule
    Comparison & Models (A $-$ B) & $\Delta F_2$ & 95\% CI \\
    \midrule
    cls-det (val-selected) & YOLO26-x @224 $-$ RF-DETR-n & +0.176 & [+0.048, +0.336]$^*$ \\
    cls-seg (val-selected) & YOLO26-x @224 $-$ Mask2Former-Swin-L & +0.241 & [+0.068, +0.454]$^*$ \\
    cls-obb (val-selected) & YOLO26-x @224 $-$ YOLO26-x & +0.427 & [+0.211, +0.677]$^*$ \\
    seg-det (val-selected) & Mask2Former-Swin-L $-$ RF-DETR-n & -0.065 & [-0.153, -0.006]$^*$ \\
    det-bfi (val-selected) & RF-DETR-n $-$ YOLO26-m & +0.178 & [+0.064, +0.328]$^*$ \\
    seg-det within RF-DETR (val-selected) & RF-DETR-Seg-l $-$ RF-DETR-n & -0.065 & [-0.142, -0.015]$^*$ \\
    cls-det & YOLO26-x $-$ RF-DETR-s & +0.218 & [+0.108, +0.351]$^*$ \\
    cls-seg & YOLO26-x $-$ RF-DETR-Seg-2xl & +0.272 & [+0.152, +0.452]$^*$ \\
    cls-obb & YOLO26-x $-$ YOLO26-l & +0.471 & [+0.282, +0.683]$^*$ \\
    seg-det & RF-DETR-Seg-2xl $-$ RF-DETR-s & -0.054 & [-0.140, -0.012]$^*$ \\
    det-bfi & RF-DETR-s $-$ YOLO26-m & +0.198 & [+0.096, +0.352]$^*$ \\
    \bottomrule
  \end{tabular}}
  \\[2pt]{\footnotesize $^*$ interval excludes zero.}
\end{table}

\subsection{Multi-seed spread}
\label{sec:multiseed}
The bootstrap quantifies sensitivity to the composition of the test set; training randomness is quantified separately by retraining the medium detection and instance-segmentation baselines with three seeds under otherwise identical recipes (Table~\ref{tab:multiseed}). Threshold-free metrics are the most stable ($AP_{50}$ within $\pm0.02$--$0.03$, $mAP_{50:95}$ within $\pm0.003$--$0.03$), while $F_2$ at the validation-selected operating point varies more (detection $0.467 \pm 0.040$, instance segmentation $0.519 \pm 0.035$), because operating-point selection adds its own variance on top of training noise; this is why the protocol fixes the threshold on validation identically for every model. The seed range sits inside the corresponding bootstrap intervals, and both analyses point to the same reading: differences of a few $F_2$ points between models within a task are noise from either source, while the cross-task gaps of Table~\ref{tab:paired_gaps} exceed four times the largest seed standard deviation in Table~\ref{tab:multiseed} ($0.040$). Additional seeds of the headline configuration of every task, trained on the cluster under identical conditions, extend this table as they complete.

\begin{table}[!htbp]
  \caption{Seed spread of the headline configurations: mean $\pm$ one sample standard deviation over the available seeds, individual values in parentheses (seed 0 = the run in the main tables; further seeds are cluster re-runs of the identical script with only the seed changed). Threshold-free metrics are the most stable; $F_2$ at the validation-selected operating point adds selection variance. The last block repeats the two medium baselines retrained on an RTX 4090 at batch 24 (upper bound on pure seed variance).}
  \label{tab:multiseed}
  \centering
  \scriptsize
  \fittable{\begin{tabular}{llp{0.62\linewidth}}
    \toprule
    Configuration & Seeds & Metrics: mean $\pm$ sd (values) \\
    \midrule
    RT-DETR-l (bbox) & 0, 1, 2 & $F_2$: 0.533 $\pm$ 0.011 (0.535, 0.542, 0.520); $AP_{50}$: 0.467 $\pm$ 0.028 (0.448, 0.499, 0.454); $mAP_{50:95}$: 0.172 $\pm$ 0.025 (0.153, 0.163, 0.200) \\
    YOLO26-x (bbox) & 0, 1, 2 & $F_2$: 0.515 $\pm$ 0.015 (0.507, 0.505, 0.532); $AP_{50}$: 0.476 $\pm$ 0.016 (0.479, 0.490, 0.459); $mAP_{50:95}$: 0.197 $\pm$ 0.013 (0.207, 0.200, 0.183) \\
    ResNet-152 @224 (classification) & 0, 1, 2 & $F_2$: 0.858 $\pm$ 0.037 (0.901, 0.840, 0.834); AUROC: 0.907 $\pm$ 0.029 (0.928, 0.920, 0.874); Acc.: 0.790 $\pm$ 0.097 (0.842, 0.849, 0.678) \\
    YOLO26-x (classification) & 0, 1, 2 & $F_2$: 0.931 $\pm$ 0.002 (0.933, 0.932, 0.930); AUROC: 0.964 $\pm$ 0.011 (0.971, 0.952, 0.970); Acc.: 0.914 $\pm$ 0.027 (0.935, 0.924, 0.885) \\
    RF-DETR-Seg-n (instance segmentation) & 0, 1, 2 & $F_2$: 0.647 $\pm$ 0.008 (0.656, 0.642, 0.643); $AP_{50}$: 0.642 $\pm$ 0.009 (0.651, 0.640, 0.634); $mAP_{50:95}$: 0.258 $\pm$ 0.008 (0.258, 0.267, 0.251) \\
    Mask2Former-Swin-L (instance segmentation) & 0, 1, 2 & $F_2$: 0.617 $\pm$ 0.012 (0.630, 0.606, 0.616); $AP_{50}$: 0.553 $\pm$ 0.012 (0.563, 0.540, 0.554); $mAP_{50:95}$: 0.210 $\pm$ 0.006 (0.206, 0.216, 0.207) \\
    MaskDINO-Swin-L (instance segmentation) & 0, 1, 2 & $F_2$: 0.638 $\pm$ 0.017 (0.619, 0.645, 0.651); $AP_{50}$: 0.610 $\pm$ 0.024 (0.585, 0.613, 0.632); $mAP_{50:95}$: 0.231 $\pm$ 0.013 (0.223, 0.246, 0.224) \\
    RF-DETR-Seg-s (instance segmentation) & 0, 1, 2 & $F_2$: 0.647 $\pm$ 0.008 (0.644, 0.656, 0.640); $AP_{50}$: 0.630 $\pm$ 0.007 (0.626, 0.638, 0.627); $mAP_{50:95}$: 0.262 $\pm$ 0.007 (0.255, 0.268, 0.262) \\
    YOLO26-x (instance segmentation) & 0, 1, 2 & $F_2$: 0.549 $\pm$ 0.037 (0.582, 0.555, 0.509); $AP_{50}$: 0.517 $\pm$ 0.033 (0.553, 0.509, 0.489); $mAP_{50:95}$: 0.197 $\pm$ 0.045 (0.249, 0.164, 0.180) \\
    YOLO26-l (obb) & 0, 1, 2 & $F_2$: 0.446 $\pm$ 0.024 (0.462, 0.457, 0.418); $AP_{50}$: 0.373 $\pm$ 0.039 (0.388, 0.402, 0.329) \\
    MaskDINO-R50 (panoptic segmentation) & 0, 1, 2 & $PQ_{rip}$: 0.442 $\pm$ 0.019 (0.433, 0.464, 0.428); $PQ$: 0.826 $\pm$ 0.005 (0.824, 0.831, 0.822) \\
    Mask2Former-Swin-L (panoptic segmentation) & 0, 1, 2 & $PQ_{rip}$: 0.470 $\pm$ 0.018 (0.465, 0.490, 0.454); $PQ$: 0.826 $\pm$ 0.004 (0.826, 0.830, 0.823) \\
    MaskDINO-Swin-L (panoptic segmentation) & 0, 1, 2 & $PQ_{rip}$: 0.460 $\pm$ 0.013 (0.459, 0.474, 0.448); $PQ$: 0.825 $\pm$ 0.005 (0.825, 0.830, 0.821) \\
    MaskDINO-R50 (semantic segmentation) & 0, 1, 2 & $IoU_{rip}$: 0.389 $\pm$ 0.029 (0.397, 0.357, 0.414); mIoU: 0.822 $\pm$ 0.009 (0.826, 0.812, 0.828) \\

    YOLO26-n (semantic segmentation) & 0, 1, 2 & $IoU_{rip}$: 0.384 $\pm$ 0.019 (0.406, 0.370, 0.376); mIoU: 0.813 $\pm$ 0.005 (0.818, 0.808, 0.814) \\
    \midrule
    YOLO26-m detection (RTX 4090 seeds) & 0, 1, 2 & $F_2$: 0.467 $\pm$ 0.040 (0.504, 0.473, 0.423); $AP_{50}$: 0.377 $\pm$ 0.020 (0.374, 0.398, 0.358); $mAP_{50:95}$: 0.154 $\pm$ 0.003 (0.151, 0.157, 0.154) \\
    YOLO26-m instance seg. (RTX 4090 seeds) & 0, 1, 2 & $F_2$: 0.519 $\pm$ 0.035 (0.538, 0.478, 0.539); $AP_{50}$: 0.489 $\pm$ 0.029 (0.521, 0.463, 0.485); $mAP_{50:95}$: 0.202 $\pm$ 0.029 (0.234, 0.193, 0.178) \\
    \bottomrule
  \end{tabular}}
\end{table}

\section{Inter-Annotator Agreement}
\label{sec:iaa_appendix}
Tables~\ref{tab:iaa} and \ref{tab:iaa_viewpoint} give the full results of the blind cross-annotation study summarized in Section~\ref{sec:protocol_details}: each expert re-annotated $200$ frames originally annotated by the other, blind to the released labels, for boxes, oriented boxes and instance masks; instances are paired per frame by maximum IoU (Hungarian assignment) and both directions are pooled ($600$ pairs per task; the two directions agree, e.g., box medians $0.785$ and $0.814$). The sample is consensus-positive by design: presence had been adjudicated before, so the study isolates geometric agreement. Two readings matter for the benchmark. At the operating threshold (IoU $0.5$) the experts agree on $95$--$98\%$ of instances, so the released spatial labels are consistent where the evaluation operates; at IoU $0.75$ agreement drops to $52$--$68\%$, which quantifies the boundary ambiguity of the phenomenon and sets a ceiling on what the strict end of the $50{:}95$ range can measure. Agreement is nearly viewpoint-invariant (box medians $0.79$--$0.81$; the water-level cell has $18$ pairs), so the viewpoint effects in Appendix~\ref{sec:stratified} are model limitations. The scripts that recompute these tables from the raw annotation files ship with the tooling.

\begin{table}[!htbp]
  \caption{Blind inter-annotator agreement on the spatial labels. Each expert re-annotated, blind to the released labels, $200$ frames originally annotated by the other ($400$ frames); instances are paired per frame by maximum IoU (Hungarian assignment) and both directions are pooled. Medians over matched pairs; the last two columns give the share of pairs above the benchmark operating threshold (IoU $0.5$) and above IoU $0.75$.}
  \label{tab:iaa}
  \centering
  \fittable{\begin{tabular}{lrccccc}
    \toprule
    Task & Pairs & Median IoU & Mean IoU & Median Dice & $\geq 0.5$ & $\geq 0.75$ \\
    \midrule
    Bounding boxes & 600 & 0.805 & 0.785 & 0.892 & 98.3\% & 67.8\% \\
    Oriented boxes & 600 & 0.759 & 0.736 & 0.863 & 94.8\% & 52.0\% \\
    Instance masks & 600 & 0.770 & 0.747 & 0.870 & 97.5\% & 57.3\% \\
    \bottomrule
  \end{tabular}}
\end{table}

\begin{table}[!htbp]
  \caption{Inter-annotator agreement by viewpoint: median IoU of matched pairs (number of pairs in parentheses). Expert agreement is nearly viewpoint-invariant, while model performance is not (Appendix~\ref{sec:stratified}); the water-level cell is small.}
  \label{tab:iaa_viewpoint}
  \centering
  \fittable{\begin{tabular}{lcccc}
    \toprule
    Task & Bird's eye & Aerial tilted & Elevated & Water-level \\
    \midrule
    Bounding boxes & 0.794 (157) & 0.803 (273) & 0.813 (152) & 0.811 (18) \\
    Oriented boxes & 0.750 (157) & 0.770 (273) & 0.745 (152) & 0.706 (18) \\
    Instance masks & 0.751 (157) & 0.772 (273) & 0.777 (152) & 0.711 (18) \\
    \bottomrule
  \end{tabular}}
\end{table}

\section{Box Protocol Ablation: Tight vs.\ Mask-Derived Boxes}
\label{sec:box_protocol_ablation}
RipBench releases two box views (Section~\ref{sec:bbox_vs_mask}): the manual tight boxes, which are the official detection target, and \texttt{bbox\_from\_instance}, the axis-aligned bounding rectangles of the instance masks (median inter-pass IoU $0.82$). To measure how much the protocol choice affects the detection numbers, the same YOLO26 family (five sizes, canonical scenario, identical recipe) was trained on each view and evaluated against both ground truths, with the confidence threshold selected on the matching validation ground truth and frozen (Table~\ref{tab:box_protocol}). Evaluated against its own target, neither protocol is systematically easier: $F_2$ differs by less than $0.07$ at every size and in no consistent direction (tight $0.42$--$0.51$, derived $0.44$--$0.52$), and the same holds under the stricter $50{:}95$ criterion. On a shared target all ten variants land in one band (on the tight target: $AP_{50}$ $0.35$--$0.49$, $F_2$ $0.41$--$0.53$), so the cross-task gap of Section~\ref{sec:results_tasks} survives full target harmonization. The one asymmetry in the table, derived-trained models scoring slightly higher on the tight target than tight-trained ones ($0.47$--$0.53$ vs.\ $0.42$--$0.51$), is within the seed spread of the medium baseline (Appendix~\ref{sec:multiseed}) and is not claimed. The tight boxes thus give equally difficult but cleaner supervision: the derived box covers $20\%$ more area at the median, and $43\%$ or more for a tenth of the instances, all of it background water.

\begin{table}[!htbp]
  \caption{Box protocol ablation: YOLO26 detectors trained on the manual tight boxes (T) or on the mask-derived boxes (D), each evaluated against both ground truths on the test split (confidence selected on the matching validation ground truth, frozen). Cells: $F_2@0.50$ / $AP_{50}$ / $mAP_{50:95}$. Diagonal cells are the official rows of the detection tables; off-diagonal cells score the same predictions against the other target.}
  \label{tab:box_protocol}
  \centering
  \small
  \fittable{\begin{tabular}{llccccc}
    \toprule
    Train & Eval GT & nano & small & medium & large & xlarge \\
    \midrule
    Tight (T) & tight & 0.444 / 0.409 / 0.191 & 0.425 / 0.351 / 0.136 & 0.504 / 0.374 / 0.151 & 0.505 / 0.430 / 0.184 & 0.507 / 0.479 / 0.207 \\
    Tight (T) & derived & 0.409 / 0.373 / 0.137 & 0.407 / 0.327 / 0.106 & 0.482 / 0.345 / 0.111 & 0.446 / 0.354 / 0.124 & 0.467 / 0.400 / 0.116 \\
    \midrule
    Derived (D) & tight & 0.495 / 0.448 / 0.179 & 0.506 / 0.441 / 0.181 & 0.529 / 0.450 / 0.167 & 0.471 / 0.452 / 0.166 & 0.521 / 0.489 / 0.179 \\
    Derived (D) & derived & 0.437 / 0.352 / 0.162 & 0.489 / 0.401 / 0.154 & 0.517 / 0.437 / 0.153 & 0.467 / 0.458 / 0.228 & 0.502 / 0.473 / 0.233 \\
    \bottomrule
  \end{tabular}}
\end{table}

\section{Stratified Analysis}
\label{sec:stratified}
All strata are defined by the per-video attribution file and scored with the released tooling (\texttt{eval/stratified\_eval.py}): predictions are grouped by video, the operating point stays the one selected on the full validation split (nothing is re-selected per stratum), and $F_2$ is computed from the stratum's summed counts; the tables give the number of test videos per stratum because the strata are small and the statistical limits differ.

\paragraph{Viewpoint} (Table~\ref{tab:stratified_viewpoint}). Classification is robust across the four viewpoints; localization varies by model family, with water-level performance especially weak for YOLO26 detection. Water level is not uniformly the weakest viewpoint for detection or instance segmentation. The water-level stratum holds five test videos (three rip-positive, $299$ instances) and its cells are indicative only.

\paragraph{Visual signature} (Table~\ref{tab:stratified_signature}). Signatures exist for the $31$ rip-positive test videos only, so these rows omit the false positives of the $40$ negative videos and are not comparable with the overall numbers. Dark-gap rips score above breaking-wave gaps and sediment plumes in both tasks; the single foam-plume video is harder only for instance segmentation. The breakdown cannot separate the signature from correlated properties (viewpoint, resolution, video identity) but identifies the difficult cases that deserve controlled study: can models become less dependent on the most visually explicit manifestation of a rip?

\paragraph{Resolution} (Table~\ref{tab:stratified_resolution}). Treating each video as the unit of analysis, the association with native resolution is non-monotonic (detection peaks below 720p, instance segmentation at 1080p), buckets hold $2$--$11$ videos, and resolution is confounded with viewpoint (the highest resolutions are mostly drone footage). The honest conclusion is descriptive: on this test set, resolution effects cannot be separated from video identity and content. Per-video results for every model ship with the tooling (\texttt{eval/results/stratified/}).

\begin{table}[!htbp]
  \caption{Test $F_2@0.50$ by camera viewpoint (frozen validation-selected thresholds; number of test videos per stratum in the header, rip-positive videos in parentheses). Classification is robust across viewpoints; localization varies by viewpoint and model family (small water-level stratum). Rows marked (seed 1) and (seed 2) are the seed re-runs of Table~\ref{tab:multiseed}; unmarked rows are the runs of the main tables.}
  \label{tab:stratified_viewpoint}
  \centering
  \small
  \textbf{Classification}\\[2pt]
  \fittable{\begin{tabular}{lccccc}
    \toprule
    Model & Bird's eye [17 (7)] & Aerial tilted [27 (13)] & Elevated [22 (8)] & Water-level [5 (3)] & Overall \\
    \midrule
    YOLO26-x & 0.917 & 0.947 & 0.917 & 0.997 & 0.933 \\
    YOLO26-x (seed 1) & 0.917 & 0.980 & 0.867 & 0.997 & 0.932 \\
    YOLO26-x (seed 2) & 0.934 & 0.960 & 0.885 & 0.911 & 0.930 \\
    YOLO26-n & 0.940 & 0.918 & 0.911 & 0.723 & 0.911 \\
    ResNet-152 @224 & 0.896 & 0.938 & 0.846 & 0.932 & 0.901 \\
    \bottomrule
  \end{tabular}}\\[6pt]
  \textbf{Detection}\\[2pt]
  \fittable{\begin{tabular}{lccccc}
    \toprule
    Model & Bird's eye [17 (7)] & Aerial tilted [27 (13)] & Elevated [22 (8)] & Water-level [5 (3)] & Overall \\
    \midrule
    RF-DETR-s & 0.780 & 0.566 & 0.846 & 0.593 & 0.715 \\
    RF-DETR-l & 0.765 & 0.570 & 0.770 & 0.799 & 0.708 \\
    RF-DETR-n & 0.755 & 0.522 & 0.883 & 0.428 & 0.695 \\
    RF-DETR-m & 0.728 & 0.540 & 0.807 & 0.334 & 0.669 \\
    RT-DETR-l (seed 1) & 0.634 & 0.319 & 0.738 & 0.433 & 0.542 \\
    RT-DETR-l & 0.634 & 0.394 & 0.641 & 0.174 & 0.535 \\
    YOLO26-x (seed 2) & 0.614 & 0.377 & 0.625 & 0.340 & 0.532 \\
    RT-DETR-l (seed 2) & 0.696 & 0.400 & 0.393 & 0.236 & 0.520 \\
    YOLO26-x & 0.653 & 0.263 & 0.593 & 0.177 & 0.507 \\
    YOLO26-l & 0.631 & 0.269 & 0.610 & 0.367 & 0.505 \\
    YOLO26-x (seed 1) & 0.626 & 0.250 & 0.618 & 0.370 & 0.505 \\
    YOLO26-m & 0.577 & 0.428 & 0.537 & 0.162 & 0.504 \\
    RT-DETR-x & 0.630 & 0.421 & 0.325 & 0.118 & 0.470 \\
    YOLO26-n & 0.681 & 0.098 & 0.468 & 0.170 & 0.444 \\
    YOLO26-s & 0.656 & 0.150 & 0.382 & 0.112 & 0.425 \\
    \bottomrule
  \end{tabular}}\\[6pt]
  \textbf{Instance segmentation}\\[2pt]
  \fittable{\begin{tabular}{lccccc}
    \toprule
    Model & Bird's eye [17 (7)] & Aerial tilted [27 (13)] & Elevated [22 (8)] & Water-level [5 (3)] & Overall \\
    \midrule
    RF-DETR-Seg-x & 0.715 & 0.498 & 0.846 & 0.461 & 0.663 \\
    RF-DETR-Seg-2xl & 0.726 & 0.482 & 0.850 & 0.385 & 0.661 \\
    RF-DETR-Seg-n & 0.715 & 0.485 & 0.817 & 0.538 & 0.656 \\
    RF-DETR-Seg-s (seed 1) & 0.675 & 0.456 & 0.913 & 0.573 & 0.655 \\
    MaskDINO-Swin-L (seed 2) & 0.711 & 0.465 & 0.840 & 0.500 & 0.651 \\
    MaskDINO-Swin-L (seed 1) & 0.668 & 0.515 & 0.855 & 0.417 & 0.645 \\
    RF-DETR-Seg-s & 0.705 & 0.445 & 0.842 & 0.509 & 0.644 \\
    RF-DETR-Seg-n (seed 2) & 0.693 & 0.469 & 0.833 & 0.471 & 0.643 \\
    CAVIS-Swin-L + refiner (video) & 0.778 & 0.362 & 0.760 & 0.321 & 0.630 \\
    CAVIS-Swin-L online (video) & 0.775 & 0.384 & 0.633 & 0.350 & 0.603 \\
    MinVIS-Swin-L (video) & 0.693 & 0.424 & 0.678 & 0.186 & 0.588 \\
    \bottomrule
  \end{tabular}}\\[6pt]
\end{table}

\begin{table}[!htbp]
  \caption{Test $F_2@0.50$ / $AP_{50}$ by primary visual signature of the rip current, for the strongest canonical model per task. Signatures exist only for the 31 rip-positive test videos: these rows omit the false positives of the 40 negative test videos and are not comparable with the overall benchmark numbers; several strata are tiny (videos / instances in parentheses) and the values are descriptive.}
  \label{tab:stratified_signature}
  \centering
  \small
  \fittable{\begin{tabular}{lcc}
    \toprule
    Signature (videos / instances) & Detection: RF-DETR-s & Inst. seg.: RF-DETR-Seg-x \\
    \midrule
    dark gap (20 / 3608) & 0.829 / 0.856 & 0.788 / 0.821 \\
    breaking wave gap (4 / 340) & 0.599 / 0.617 & 0.485 / 0.454 \\
    sediment plume (6 / 2026) & 0.544 / 0.463 & 0.519 / 0.433 \\
    foam plume (1 / 203) & 0.878 / 0.906 & 0.296 / 0.142 \\
    \bottomrule
  \end{tabular}}
\end{table}

\begin{table}[!htbp]
  \caption{Per-video test $F_2@0.50$ of the strongest canonical detector and instance segmenter, grouped by the native resolution of the source video (rip-positive test videos; median and inter-quartile range over videos). Buckets hold 2--11 videos and resolution is confounded with viewpoint (the highest resolutions are mostly drone footage), so the association is descriptive, not causal.}
  \label{tab:stratified_resolution}
  \centering
  \small
  \fittable{\begin{tabular}{lccc}
    \toprule
    Bucket (rip videos) & Detection: RF-DETR-s & Inst. seg.: RF-DETR-Seg-x \\
    \midrule
    <720p (2) & 0.910 [0.86, 0.95] & 0.623 [0.53, 0.72] \\
    720p (11) & 0.569 [0.47, 0.79] & 0.487 [0.23, 0.71] \\
    1080p (9) & 0.824 [0.48, 0.88] & 0.852 [0.30, 0.91] \\
    >=1440p (9) & 0.363 [0.06, 0.69] & 0.066 [0.00, 0.62] \\
    \bottomrule
  \end{tabular}}
\end{table}

\section{Label Compatibility with RipVIS}
\label{sec:transfer}
Because RipBench inherits the RipVIS video pool (Sec.~\ref{sec:collection_diversity}), a cross-dataset experiment between the two cannot measure beach-to-beach generalization; it measures whether the two label pools are compatible. We trained the same architecture (YOLO26-m) on the RipVIS v1.8.4 labels as released (instance masks; standalone stills excluded, unlabeled frames kept as negatives) and on the RipBench labels, and evaluated both on the RipBench test split under the standard protocol (threshold selected on RipBench validation, frozen). A split-history audit, confirmed by content-based matching of every video, found that four RipBench test videos (RipBench-059, -072, -079, -147) had belonged to the RipVIS validation split on which the RipVIS checkpoints were selected; they are therefore excluded from the evaluation of \emph{both} models, leaving a clean subset of 67 videos and 9,046 frames. No RipVIS training video occurs in the RipBench test split. Table~\ref{tab:transfer} shows that both paired differences include zero and change sign between tasks: existing RipVIS supervision remains usable against the RipBench targets, and neither label pool is resolved as better at the current test-set size. The value of the re-annotation lies elsewhere: denser and uniform frame-order sampling, balanced difficult negatives, verified spatial agreement (Sec.~\ref{sec:protocol_details}), official targets for all six tasks, and one shared split, which are what the video views of Appendix~\ref{sec:video} rely on.

\begin{table}[!htbp]
  \caption{Same architecture (YOLO26-m) trained on the RipVIS labels vs.\ the RipBench labels, both evaluated on the clean RipBench test subset (67 videos, 9,046 frames; the four former RipVIS-validation videos are excluded because the RipVIS checkpoints were selected on them). Thresholds selected on RipBench validation and frozen. Paired video-level bootstrap difference (RipVIS-trained minus RipBench-trained, $B=1{,}000$).}
  \label{tab:transfer}
  \centering
  \small
  \fittable{\begin{tabular}{lccc}
    \toprule
    Task (evaluation GT) & RipVIS-trained $F_2$ & RipBench-trained $F_2$ & Difference [95\% CI] \\
    \midrule
    Instance segmentation (instance masks) & 0.559 & 0.522 & +0.037 [-0.016, +0.136] \\
    Derived boxes (mask-derived boxes) & 0.484 & 0.494 & -0.010 [-0.078, +0.052] \\
    \bottomrule
  \end{tabular}}
\end{table}

\section{Full Result Tables}
\label{sec:full_tables}
Tables~\ref{tab:classification-results-full}--\ref{tab:inst-segmentation-full} list every trained variant of the classification, detection and instance-segmentation families in both training scenarios (rows marked +add were trained with the $3{,}773$ opt-in standalone stills in addition to the sampled frames; the stills and their tight-box, oriented-box and instance annotations come from our earlier work \cite{dumitriu2023rip} and were annotated by the same two experts; validation and test contain no stills). Oriented detection was trained in the canonical scenario only, so its complete table is Table~\ref{tab:obb-detection} in the main text. The semantic and panoptic families have no stills scenario, so their complete tables are Tables~\ref{tab:semantic-segmentation-full} and \ref{tab:pan-segmentation-full} in the main text.

\begin{table}[!htbp]
  \caption{Frame-level classification on the test split. $F_2$ with its 95\% video-level CI, best in \textcolor{blue}{blue}; both training scenarios (+add = trained with the additional stills), protocol in Sec.~\ref{sec:eval_protocol}.}
  \label{tab:classification-results-full}
  \centering
  \small
  \fittable{\begin{tabular}{lccccccc}
    \toprule
    Model & Acc. & P & R & $F_1$ & $F_2$ [95\% CI] & AUROC & FPS \\
    \midrule
    ResNet-18 @224 & 0.840 & 0.840 & 0.808 & 0.824 & 0.814 [0.64, 0.90] & 0.903 & 226 \\
    ResNet-18 @224 +add & 0.819 & 0.732 & 0.959 & 0.830 & 0.903 [0.81, 0.95] & 0.941 & 230 \\
    ResNet-18 @640 & 0.766 & 0.681 & 0.924 & 0.784 & 0.863 [0.73, 0.93] & 0.930 & 120 \\
    ResNet-18 @640 +add & 0.855 & 0.806 & 0.903 & 0.852 & 0.882 [0.78, 0.96] & 0.936 & 120 \\
    ResNet-34 @224 & 0.881 & 0.902 & 0.832 & 0.866 & 0.845 [0.70, 0.93] & 0.933 & 204 \\
    ResNet-34 @224 +add & 0.837 & 0.761 & 0.942 & 0.842 & 0.899 [0.79, 0.96] & 0.941 & 204 \\
    ResNet-34 @640 & 0.875 & 0.847 & 0.890 & 0.868 & 0.881 [0.77, 0.94] & 0.960 & 108 \\
    ResNet-34 @640 +add & 0.781 & 0.680 & 0.988 & 0.806 & 0.906 [0.81, 0.95] & 0.945 & 108 \\
    ResNet-50 @224 & 0.815 & 0.796 & 0.805 & 0.800 & 0.803 [0.65, 0.92] & 0.923 & 106 \\
    ResNet-50 @224 +add & 0.772 & 0.725 & 0.812 & 0.766 & 0.793 [0.61, 0.88] & 0.835 & 184 \\
    ResNet-50 @640 & 0.908 & 0.905 & 0.894 & 0.900 & 0.896 [0.78, 0.96] & 0.966 & 88 \\
    ResNet-50 @640 +add & 0.839 & 0.757 & 0.958 & 0.846 & 0.910 [0.79, 0.97] & 0.972 & 94 \\
    ResNet-101 @224 & 0.823 & 0.763 & 0.895 & 0.824 & 0.865 [0.72, 0.94] & 0.928 & 139 \\
    ResNet-101 @224 +add & 0.753 & 0.651 & 0.999 & 0.788 & 0.903 [0.79, 0.95] & 0.929 & 112 \\
    ResNet-101 @640 & 0.900 & 0.908 & 0.871 & 0.889 & 0.878 [0.75, 0.95] & 0.959 & 80 \\
    ResNet-101 @640 +add & 0.793 & 0.700 & 0.963 & 0.811 & 0.895 [0.77, 0.96] & 0.953 & 82 \\
    ResNet-152 @224 & 0.842 & 0.769 & 0.941 & 0.846 & 0.901 [0.78, 0.95] & 0.928 & 108 \\
    ResNet-152 @224 +add & 0.798 & 0.719 & 0.922 & 0.808 & 0.873 [0.75, 0.94] & 0.921 & 118 \\
    ResNet-152 @640 & 0.850 & 0.795 & 0.909 & 0.848 & 0.884 [0.75, 0.95] & 0.932 & 72 \\
    ResNet-152 @640 +add & 0.790 & 0.692 & 0.980 & 0.811 & 0.905 [0.79, 0.96] & 0.957 & 74 \\
    YOLO26-n & 0.860 & 0.790 & 0.948 & 0.862 & 0.911 [0.81, 0.96] & 0.960 & 110 \\
    YOLO26-n +add & 0.716 & 0.623 & 0.974 & 0.760 & 0.875 [0.74, 0.94] & 0.958 & 115 \\
    YOLO26-n @224 & 0.853 & 0.812 & 0.885 & 0.847 & 0.870 [0.73, 0.94] & 0.939 & 210 \\
    YOLO26-n @224 +add & 0.852 & 0.774 & 0.961 & 0.857 & 0.917 [0.83, 0.96] & 0.959 & 210 \\
    YOLO26-s & 0.845 & 0.779 & 0.928 & 0.847 & 0.893 [0.76, 0.96] & 0.946 & 114 \\
    YOLO26-s +add & 0.755 & 0.662 & 0.958 & 0.783 & 0.879 [0.74, 0.95] & 0.942 & 112 \\
    YOLO26-s @224 & 0.810 & 0.735 & 0.921 & 0.817 & 0.876 [0.75, 0.94] & 0.920 & 205 \\
    YOLO26-s @224 +add & 0.782 & 0.721 & 0.859 & 0.784 & 0.827 [0.68, 0.94] & 0.903 & 193 \\
    YOLO26-m & 0.864 & 0.847 & 0.860 & 0.853 & 0.857 [0.73, 0.94] & 0.947 & 108 \\
    YOLO26-m +add & 0.840 & 0.758 & 0.962 & 0.848 & 0.913 [0.81, 0.96] & 0.963 & 102 \\
    YOLO26-m @224 & 0.797 & 0.706 & 0.961 & 0.814 & 0.896 [0.78, 0.95] & 0.937 & 162 \\
    YOLO26-m @224 +add & 0.853 & 0.778 & 0.953 & 0.856 & 0.912 [0.82, 0.96] & 0.948 & 191 \\
    YOLO26-l & 0.626 & 0.553 & 0.990 & 0.709 & 0.854 [0.72, 0.93] & 0.921 & 91 \\
    YOLO26-l +add & 0.855 & 0.774 & 0.968 & 0.860 & 0.922 [0.82, 0.97] & 0.971 & 88 \\
    YOLO26-l @224 & 0.817 & 0.738 & 0.935 & 0.825 & 0.888 [0.75, 0.95] & 0.936 & 164 \\
    YOLO26-l @224 +add & 0.891 & 0.821 & 0.977 & 0.892 & \textcolor{blue}{0.941} [0.87, 0.98] & 0.981 & 165 \\
    YOLO26-x & 0.935 & 0.925 & 0.935 & 0.930 & 0.933 [0.85, 0.97] & 0.971 & 84 \\
    YOLO26-x +add & 0.767 & 0.670 & 0.975 & 0.794 & 0.894 [0.77, 0.96] & 0.965 & 86 \\
    YOLO26-x @224 & 0.777 & 0.692 & 0.931 & 0.794 & 0.871 [0.72, 0.94] & 0.921 & 162 \\
    YOLO26-x @224 +add & 0.754 & 0.656 & 0.982 & 0.786 & 0.893 [0.78, 0.95] & 0.959 & 166 \\
    \bottomrule
  \end{tabular}}
\end{table}

\begin{table}[!htbp]
  \caption{Object detection (manual tight boxes) on the test split: AP (threshold-free) and $F_{\{1,2\}}$ scores at IoU $0.50$ and averaged over $0.50{:}0.95$. $F_2$ with its 95\% video-level CI, best in \textcolor{blue}{blue}; both training scenarios (+add = trained with the additional stills), protocol in Sec.~\ref{sec:eval_protocol}.}
  \label{tab:box-detection-full}
  \centering
  \small
  \fittable{\begin{tabular}{lcccccccc}
    \toprule
    Model & $AP_{50}$ & $AP_{75}$ & $mAP_{50:95}$ & ${F_1}_{[50]}$ & ${F_1}_{[50:95]}$ & ${F_2}_{[50]}$ [95\% CI] & ${F_2}_{[50:95]}$ & FPS \\
    \midrule
    Co-DETR-R50 +add$^{\ddagger}$ & 0.583 & 0.160 & 0.231 & 0.617 & 0.305 & 0.613 [0.40, 0.75] & 0.302 & 24 \\
    Co-DETR-Swin-L +add$^{\ddagger}$ & 0.683 & 0.150 & 0.268 & 0.601 & 0.303 & 0.692 [0.54, 0.80] & 0.348 & 14 \\
    DINO-R50 +add$^{\ddagger}$ & 0.587 & 0.105 & 0.228 & 0.596 & 0.292 & 0.591 [0.38, 0.73] & 0.289 & 24 \\
    DINO-Swin-L +add$^{\ddagger}$ & 0.609 & 0.071 & 0.211 & 0.655 & 0.293 & 0.679 [0.50, 0.80] & 0.303 & 7 \\
    MaskDINO-R50$^{\dagger}$ & 0.501 & 0.164 & 0.233 & 0.555 & 0.322 & 0.542 [0.25, 0.73] & 0.315 & 31 \\
    MaskDINO-R50 +add$^{\dagger}$ & 0.588 & 0.187 & 0.248 & 0.608 & 0.315 & 0.629 [0.41, 0.78] & 0.325 & 32 \\
    MaskDINO-Swin-L$^{\dagger}$ & 0.588 & 0.174 & 0.257 & 0.638 & 0.352 & 0.612 [0.35, 0.78] & 0.337 & 17 \\
    RF-DETR-n & 0.699 & 0.106 & 0.244 & 0.659 & 0.298 & 0.695 [0.51, 0.81] & 0.314 & 73 \\
    RF-DETR-n +add & 0.730 & 0.191 & 0.291 & 0.659 & 0.330 & 0.716 [0.55, 0.83] & 0.359 & 74 \\
    RF-DETR-s & 0.718 & 0.132 & 0.280 & 0.681 & 0.323 & 0.715 [0.56, 0.83] & 0.340 & 70 \\
    RF-DETR-s +add & 0.715 & 0.139 & 0.285 & 0.683 & 0.340 & \textcolor{blue}{0.722} [0.55, 0.84] & 0.359 & 68 \\
    RF-DETR-m & 0.678 & 0.169 & 0.283 & 0.683 & 0.343 & 0.669 [0.48, 0.81] & 0.337 & 63 \\
    RF-DETR-m +add & 0.706 & 0.182 & 0.296 & 0.679 & 0.335 & 0.710 [0.56, 0.83] & 0.350 & 58 \\
    RF-DETR-l & 0.695 & 0.171 & 0.279 & 0.682 & 0.327 & 0.708 [0.52, 0.84] & 0.339 & 60 \\
    RT-DETR-l & 0.448 & 0.055 & 0.153 & 0.478 & 0.202 & 0.535 [0.35, 0.65] & 0.226 & 74 \\
    RT-DETR-l +add & 0.567 & 0.130 & 0.233 & 0.592 & 0.298 & 0.596 [0.33, 0.75] & 0.301 & 74 \\
    RT-DETR-x & 0.428 & 0.045 & 0.137 & 0.368 & 0.158 & 0.470 [0.25, 0.65] & 0.202 & 61 \\
    RT-DETR-x +add & 0.575 & 0.106 & 0.229 & 0.504 & 0.255 & 0.608 [0.42, 0.74] & 0.308 & 61 \\
    RTMDet-l +add$^{\ddagger}$ & 0.586 & 0.095 & 0.227 & 0.598 & 0.307 & 0.623 [0.41, 0.76] & 0.320 & 103 \\
    Sparse R-CNN-R50 +add$^{\ddagger}$ & 0.568 & 0.070 & 0.205 & 0.559 & 0.262 & 0.599 [0.41, 0.75] & 0.281 & 34 \\
    TOOD-R50 +add$^{\ddagger}$ & 0.528 & 0.101 & 0.216 & 0.494 & 0.254 & 0.543 [0.27, 0.69] & 0.278 & 85 \\
    YOLO26-n & 0.409 & 0.160 & 0.191 & 0.479 & 0.255 & 0.444 [0.15, 0.66] & 0.237 & 281 \\
    YOLO26-n +add & 0.492 & 0.136 & 0.205 & 0.530 & 0.261 & 0.538 [0.31, 0.71] & 0.265 & 277 \\
    YOLO26-s & 0.351 & 0.058 & 0.136 & 0.458 & 0.227 & 0.425 [0.16, 0.63] & 0.210 & 266 \\
    YOLO26-s +add & 0.485 & 0.133 & 0.203 & 0.526 & 0.260 & 0.542 [0.31, 0.71] & 0.268 & 265 \\
    YOLO26-m & 0.374 & 0.101 & 0.151 & 0.466 & 0.244 & 0.504 [0.35, 0.63] & 0.264 & 257 \\
    YOLO26-m +add & 0.497 & 0.159 & 0.226 & 0.512 & 0.273 & 0.531 [0.31, 0.68] & 0.283 & 257 \\
    YOLO26-l & 0.430 & 0.115 & 0.184 & 0.501 & 0.253 & 0.505 [0.27, 0.68] & 0.255 & 196 \\
    YOLO26-l +add & 0.460 & 0.094 & 0.177 & 0.520 & 0.253 & 0.513 [0.29, 0.69] & 0.250 & 186 \\
    YOLO26-x & 0.479 & 0.125 & 0.207 & 0.492 & 0.249 & 0.507 [0.27, 0.67] & 0.257 & 164 \\
    YOLO26-x +add & 0.492 & 0.097 & 0.188 & 0.496 & 0.223 & 0.529 [0.34, 0.66] & 0.238 & 165 \\
    \bottomrule
  \end{tabular}}
  \\[2pt] {\footnotesize $^{\dagger}$Instance-segmentation model scored as a detector: its boxes are the bounding rectangles of the predicted masks, evaluated against the manual-box ground truth; FPS is that of the instance model. $^{\ddagger}$Run trained on the predecessor split s1.2 with the additional stills (one flash-rip video fewer in training; validation identical), evaluated here on the s1.3 test split with the released evaluator.}
\end{table}

\begin{table}[!htbp]
  \caption{Instance segmentation on the test split: AP (threshold-free) and $F_{\{1,2\}}$ scores at IoU $0.50$ and averaged over $0.50{:}0.95$. $F_2$ with its 95\% video-level CI, best in \textcolor{blue}{blue}; both training scenarios (+add = trained with the additional stills), protocol in Sec.~\ref{sec:eval_protocol}.}
  \label{tab:inst-segmentation-full}
  \centering
  \small
  \fittable{\begin{tabular}{lcccccccc}
    \toprule
    Model & $AP_{50}$ & $AP_{75}$ & $mAP_{50:95}$ & ${F_1}_{[50]}$ & ${F_1}_{[50:95]}$ & ${F_2}_{[50]}$ [95\% CI] & ${F_2}_{[50:95]}$ & FPS \\
    \midrule
    Mask2Former-R50 & 0.534 & 0.067 & 0.197 & 0.629 & 0.276 & 0.616 [0.33, 0.80] & 0.270 & 32 \\
    Mask2Former-R50 +add & 0.493 & 0.069 & 0.182 & 0.594 & 0.265 & 0.560 [0.35, 0.69] & 0.250 & 33 \\
    Mask2Former-Swin-L & 0.563 & 0.072 & 0.206 & 0.630 & 0.283 & 0.630 [0.37, 0.79] & 0.283 & 23 \\
    MaskDINO-R50 & 0.510 & 0.104 & 0.210 & 0.557 & 0.266 & 0.544 [0.26, 0.73] & 0.259 & 31 \\
    MaskDINO-R50 +add & 0.565 & 0.133 & 0.224 & 0.593 & 0.275 & 0.601 [0.37, 0.75] & 0.279 & 32 \\
    MaskDINO-Swin-L & 0.585 & 0.113 & 0.223 & 0.639 & 0.299 & 0.619 [0.36, 0.79] & 0.290 & 17 \\
    RF-DETR-Seg-n & 0.651 & 0.128 & 0.258 & 0.671 & 0.324 & 0.656 [0.45, 0.81] & 0.317 & 54 \\
    RF-DETR-Seg-n +add & 0.709 & 0.179 & 0.284 & 0.664 & 0.307 & 0.693 [0.51, 0.82] & 0.321 & 59 \\
    RF-DETR-Seg-s & 0.626 & 0.124 & 0.255 & 0.647 & 0.308 & 0.644 [0.42, 0.79] & 0.306 & 60 \\
    RF-DETR-Seg-s +add & 0.712 & 0.188 & 0.291 & 0.671 & 0.319 & 0.696 [0.51, 0.82] & 0.331 & 56 \\
    RF-DETR-Seg-m & 0.612 & 0.112 & 0.249 & 0.606 & 0.296 & 0.632 [0.42, 0.77] & 0.308 & 50 \\
    RF-DETR-Seg-m +add & 0.704 & 0.219 & 0.299 & 0.694 & 0.331 & 0.681 [0.46, 0.82] & 0.325 & 53 \\
    RF-DETR-Seg-l & 0.630 & 0.151 & 0.267 & 0.602 & 0.305 & 0.629 [0.40, 0.78] & 0.319 & 50 \\
    RF-DETR-Seg-l +add & 0.727 & 0.225 & 0.316 & 0.688 & 0.338 & \textcolor{blue}{0.718} [0.52, 0.84] & 0.353 & 50 \\
    RF-DETR-Seg-x & 0.660 & 0.099 & 0.247 & 0.615 & 0.283 & 0.663 [0.46, 0.79] & 0.305 & -- \\
    RF-DETR-Seg-x +add & 0.684 & 0.120 & 0.267 & 0.662 & 0.310 & 0.691 [0.50, 0.82] & 0.324 & -- \\
    RF-DETR-Seg-2xl & 0.654 & 0.087 & 0.254 & 0.643 & 0.304 & 0.661 [0.45, 0.79] & 0.312 & -- \\
    RF-DETR-Seg-2xl +add & 0.698 & 0.120 & 0.266 & 0.656 & 0.302 & 0.696 [0.51, 0.82] & 0.320 & -- \\
    RTMDet-Ins-x +add$^{\ddagger}$ & 0.666 & 0.040 & 0.224 & 0.622 & 0.256 & 0.662 [0.45, 0.79] & 0.273 & 60 \\
    SparseInst-R50-base & 0.465 & 0.034 & 0.152 & 0.486 & 0.201 & 0.504 [0.28, 0.64] & 0.208 & 49 \\
    SparseInst-R50-base +add & 0.423 & 0.021 & 0.133 & 0.446 & 0.179 & 0.535 [0.34, 0.68] & 0.215 & 49 \\
    SparseInst-R50-GIAM (aug.) & 0.506 & 0.062 & 0.176 & 0.521 & 0.222 & 0.533 [0.31, 0.69] & 0.228 & 48 \\
    SparseInst-R50-GIAM (aug.) +add & 0.553 & 0.106 & 0.211 & 0.518 & 0.223 & 0.574 [0.36, 0.72] & 0.247 & 48 \\
    SparseInst-R101-GIAM & 0.585 & 0.019 & 0.179 & 0.608 & 0.228 & 0.600 [0.35, 0.76] & 0.225 & 44 \\
    SparseInst-R101-GIAM +add & 0.513 & 0.081 & 0.184 & 0.556 & 0.245 & 0.586 [0.34, 0.74] & 0.258 & 44 \\
    SparseInst-R101-GIAM (aug.) & 0.492 & 0.019 & 0.162 & 0.603 & 0.243 & 0.573 [0.29, 0.75] & 0.231 & 42 \\
    SparseInst-R101-GIAM (aug.) +add & 0.528 & 0.056 & 0.185 & 0.552 & 0.231 & 0.571 [0.34, 0.73] & 0.239 & 44 \\
    SparseInst-PVTv2-b1-GIAM & 0.559 & 0.032 & 0.177 & 0.600 & 0.237 & 0.595 [0.37, 0.74] & 0.236 & 44 \\
    SparseInst-PVTv2-b1-GIAM +add & 0.624 & 0.118 & 0.229 & 0.627 & 0.281 & 0.645 [0.43, 0.77] & 0.289 & 42 \\
    SparseInst-PVTv2-b1-GIAM (aug.) & 0.575 & 0.098 & 0.223 & 0.571 & 0.246 & 0.582 [0.34, 0.74] & 0.251 & 41 \\
    SparseInst-PVTv2-b1-GIAM (aug.) +add & 0.622 & 0.125 & 0.240 & 0.625 & 0.273 & 0.624 [0.38, 0.78] & 0.273 & 44 \\
    YOLO26-n & 0.512 & 0.067 & 0.191 & 0.519 & 0.213 & 0.522 [0.26, 0.72] & 0.215 & 234 \\
    YOLO26-n +add & 0.540 & 0.149 & 0.232 & 0.508 & 0.251 & 0.542 [0.28, 0.72] & 0.268 & 239 \\
    YOLO26-s & 0.536 & 0.133 & 0.222 & 0.565 & 0.257 & 0.553 [0.28, 0.73] & 0.251 & 232 \\
    YOLO26-s +add & 0.580 & 0.164 & 0.245 & 0.551 & 0.266 & 0.593 [0.35, 0.75] & 0.286 & 240 \\
    YOLO26-m & 0.521 & 0.148 & 0.234 & 0.511 & 0.254 & 0.538 [0.25, 0.72] & 0.268 & 202 \\
    YOLO26-m +add & 0.602 & 0.151 & 0.257 & 0.586 & 0.282 & 0.616 [0.38, 0.77] & 0.296 & 202 \\
    YOLO26-l & 0.456 & 0.047 & 0.156 & 0.496 & 0.219 & 0.543 [0.35, 0.65] & 0.240 & 166 \\
    YOLO26-l +add & 0.580 & 0.105 & 0.238 & 0.577 & 0.275 & 0.570 [0.32, 0.74] & 0.272 & 162 \\
    YOLO26-x & 0.553 & 0.164 & 0.249 & 0.534 & 0.267 & 0.582 [0.27, 0.75] & 0.291 & 129 \\
    YOLO26-x +add & 0.562 & 0.192 & 0.233 & 0.560 & 0.269 & 0.591 [0.35, 0.74] & 0.284 & 129 \\
    \bottomrule
  \end{tabular}}
  \\[2pt] {\footnotesize $^{\ddagger}$Run trained on the predecessor split s1.2 with the additional stills (one flash-rip video fewer in training; validation identical), evaluated here on the s1.3 test split with the released evaluator.}
\end{table}

\section{Video Views and Temporal Baselines}
\label{sec:video}
\paragraph{Video-instance view.} Because every video is annotated in frame order at $\geq 5$ FPS, the instance labels can be re-expressed as a video instance segmentation (VIS) dataset. The released \texttt{labels/video\_instance\_segmentation/} folder holds YouTube-VIS-format views built deterministically from the same \texttt{labels\_original} masks by \texttt{train/video\_models/ytvis\_view/}: \emph{sparse} (the annotated frames of each video as a low-fps sequence; instance identities linked greedily by mask overlap across consecutive annotated frames and then reviewed visually by the dataset authors on every rip video---52 identity merges in 20 videos over two review rounds, almost all where the camera pans between rips, tilts abruptly or a rip leaves and re-enters the view; 157 tracks over 91 rip videos in train, 48 in val, 51 in test), a \emph{curated} variant with manually corrected identities, and two \emph{densified} training views that propagate masks to the native-fps frames strictly \emph{between} annotated pairs (never into unannotated tails): nearest-copy and time-weighted signed-distance interpolation. Before using interpolated masks for training we validated the interpolation leave-one-out against real annotations: reconstructing an annotated frame from its two annotated neighbours reaches a median IoU of $0.990$ overall and $0.908$ on the frames whose annotation is not a copy of a neighbour ($57.3\%$ of consecutive annotated pairs are pixel-identical, i.e., annotators copied forward and re-drew periodically); the non-copy figure is above the inter-expert mask agreement of Sec.~\ref{sec:protocol_details}. Evaluation always uses the original expert-annotated frames, scored per frame with the unmodified benchmark protocol, so video and image models share metrics and evaluation sets; track-level VIS metrics are provided by the tooling but are identity-sensitive and are not used for ranking.

\paragraph{Baselines.} We adapted two VIS families, MinVIS \cite{minvis} (frame-trained, query-embedding linking at inference) and CAVIS \cite{cavis} (segmenter, online tracker and offline temporal refiner), at R50 and Swin-L, initialized from the same COCO Mask2Former checkpoints as the image baselines (Table~\ref{tab:video_instance}). In the matched Swin-L comparison, CAVIS online improves $AP_{50}$ from $0.563$ to $0.612$ over Mask2Former, and CAVIS with the offline refiner matches the image model's $F_2$ at $0.630$; this is not explained by input resolution (video models run at a 480-pixel short side, image models at 640). Threshold-free $AP_{50}$ gains depend on the configuration, while $F_2$ at the frozen operating point is less consistent, plausibly because track-level confidences transfer validation thresholds less well than per-frame confidences. The parity statement is scoped to the Mask2Former lineage: RF-DETR-Seg (Table~\ref{tab:inst-segmentation}) now sets the benchmark top end (canonical test $F_2$ up to $0.663$, $0.718$ for the large variant trained with the additional stills) above every Mask2Former-lineage model, image or video, and has no video counterpart in this study. In the original-identity runs of Table~\ref{tab:video_instance}, densification helps MinVIS-R50 but has mixed effects for CAVIS-R50. The offline refiner helps on the sparse view ($+0.02$ R50, $+0.03$ Swin-L) but hurts on the native-fps dense view ($-0.10$), where its 15-frame clips span only $\sim0.5$\,s. Constructed identities are not the bottleneck: curated-identity runs land within run-to-run noise. Two CAVIS segmenters trained outside the released pipeline (on the RipBench and on the RipVIS video labels) and re-scored with our unmodified protocol land next to our CAVIS-Swin-L online model ($F_2$ $0.61$, $AP_{50}$ $0.60$), a second-implementation check; a VideoMT model from the same effort could not be run through our evaluator and is not reported. All video results are single-seed, short-schedule runs with wide video-level intervals (Table~\ref{tab:video_instance}; every configuration in Table~\ref{tab:video_instance_full}); the task-level statements above rest on directions that are consistent across models and seeds, not on single gaps.

\paragraph{Views for the remaining spatial tasks and annotation propagation.} Video views for detection, oriented detection and panoptic segmentation are derived from the same instance tracks: each track's per-frame tight box, oriented box and panoptic segment are paired to the mask by maximum IoU and inherit the track identity (all $41{,}547$ instances paired, none ambiguous; median pairing IoU $0.80$--$0.83$ for the tight boxes, i.e., the inter-pass IoU of Section~\ref{sec:bbox_vs_mask}), and they are evaluated under the same per-frame protocol; the view ships with the same builder (\texttt{ytvis\_sparse\_tasks}). The densification results motivate an annotation-propagation route that we describe but did not execute at scale: fitting a per-video model on that video's annotated frames to propose labels for its remaining frames, always followed by expert verification.

\begin{table}[!htbp]
\caption{Video instance segmentation on the RipBench test split, scored per frame with the unmodified benchmark protocol against the paired image baselines. All video models start from the same COCO Mask2Former checkpoints as the image baselines and run at a 480-pixel short side (image models: 640). Views: sparse = annotated frames as low-fps sequences; dense-nearest / dense-interp = native-fps frames with copied / SDF-interpolated masks, used for training only. $F_2$ with its 95\% video-level CI; single seed and short schedules unless noted.}
  \label{tab:video_instance}
  \centering
  \small
  \fittable{\begin{tabular}{llccc}
    \toprule
    Model & View & ${F_2}_{[50]}$ [95\% CI] & $AP_{50}$ & $mAP_{50:95}$ \\
    \midrule
    \multicolumn{5}{l}{\emph{Per-frame image baselines}} \\
    Mask2Former-R50 (image) & -- & 0.616 [0.33, 0.80] & 0.534 & 0.197 \\
    MaskDINO-R50 (image) & -- & 0.544 [0.26, 0.73] & 0.510 & 0.210 \\
    Mask2Former-Swin-L (image) & -- & 0.630 [0.37, 0.79] & 0.563 & 0.206 \\
    MaskDINO-Swin-L (image) & -- & 0.619 [0.36, 0.79] & 0.585 & 0.223 \\
    RF-DETR-Seg-n (image) & -- & 0.656 [0.45, 0.81] & 0.651 & 0.258 \\
    RF-DETR-Seg-s (image) & -- & 0.644 [0.42, 0.79] & 0.626 & 0.255 \\
    RF-DETR-Seg-x (image) & -- & \textcolor{blue}{0.663} [0.46, 0.79] & 0.660 & 0.247 \\

    \midrule
    \multicolumn{5}{l}{\emph{Video models, sparse view}} \\
    MinVIS-R50 & sparse & 0.568 [0.26, 0.76] & 0.566 & 0.225 \\
    MinVIS-R50, 24k schedule & sparse & 0.581 [0.27, 0.78] & 0.570 & 0.240 \\
    \midrule
    \multicolumn{5}{l}{\emph{Video models, sparse (curated) view}} \\
    MinVIS-R50, curated ids & sparse & 0.567 [0.26, 0.76] & 0.547 & 0.218 \\
    \midrule
    \multicolumn{5}{l}{\emph{Video models, sparse view}} \\
    MinVIS-Swin-L & sparse & 0.588 [0.35, 0.78] & 0.595 & 0.226 \\
    CAVIS-R50 online & sparse & 0.576 [0.31, 0.76] & 0.465 & 0.186 \\
    CAVIS-R50 + offline refiner & sparse & 0.597 [0.29, 0.78] & 0.517 & 0.200 \\
    CAVIS-Swin-L online & sparse & 0.603 [0.35, 0.80] & 0.612 & 0.238 \\
    CAVIS-Swin-L + offline refiner & sparse & 0.630 [0.36, 0.81] & 0.610 & 0.229 \\
    Video-Mask2Former-R50 (clip-trained, whole-window inference) & sparse & 0.568 [0.31, 0.72] & 0.580 & 0.241 \\
    DVIS++-R50, segmenter stage only & sparse & 0.517 [0.31, 0.65] & 0.464 & 0.180 \\
    \midrule
    \multicolumn{5}{l}{\emph{Video models, dense-nearest view}} \\
    MinVIS-R50 & dense-nearest & 0.586 [0.28, 0.78] & 0.592 & 0.251 \\
    CAVIS-R50 online & dense-nearest & 0.586 [0.33, 0.79] & 0.579 & 0.244 \\
    \midrule
    \multicolumn{5}{l}{\emph{Video models, dense-interp view}} \\
    MinVIS-R50 & dense-interp & 0.593 [0.29, 0.78] & 0.598 & 0.240 \\
    CAVIS-R50 online & dense-interp & 0.566 [0.29, 0.75] & 0.530 & 0.224 \\
    \midrule
    \multicolumn{5}{l}{\emph{Cross-supervision}} \\
    MinVIS-R50 trained on the RipVIS-label video view & RipVIS-labels & 0.602 [0.30, 0.78] & 0.576 & 0.228 \\
    \midrule
    \multicolumn{5}{l}{\emph{Additional CAVIS-Swin-L segmenters trained outside the released pipeline, re-scored with the same protocol}} \\
    CAVIS-Swin-L, trained on RipBench & external & 0.613 [0.38, 0.78] & 0.601 & 0.247 \\
    CAVIS-Swin-L, trained on RipVIS & external & 0.611 [0.35, 0.79] & 0.598 & 0.243 \\
    \bottomrule
  \end{tabular}}
\end{table}

\begin{table}[!htbp]
  \caption{All video-model configurations evaluated on the RipBench test split (per-frame protocol), including checkpoint and schedule checks, two CAVIS-Swin-L segmenters trained outside the released pipeline, and the retrains on the round-2 identity review; the same 95\% video-level bootstrap CIs. Views as in Table~\ref{tab:video_instance}; the run identifier of every row is in the released provenance table.}
  \label{tab:video_instance_full}
  \centering
  \scriptsize
  \fittable{\begin{tabular}{lccc}
    \toprule
    Configuration & ${F_2}_{[50]}$ [95\% CI] & $AP_{50}$ & $mAP_{50:95}$ \\
    \midrule
    CAVIS-Swin-L, trained on RipBench & 0.613 [0.38, 0.78] & 0.601 & 0.247 \\
    CAVIS-Swin-L, trained on RipVIS & 0.611 [0.35, 0.79] & 0.598 & 0.243 \\
    CAVIS-R50 online, curated dense-nearest & 0.595 [0.31, 0.77] & 0.556 & 0.223 \\
    CAVIS-R50 online, curated & 0.551 [0.28, 0.72] & 0.529 & 0.214 \\
    CAVIS-R50 online, dense-interp & 0.566 [0.29, 0.75] & 0.530 & 0.224 \\
    CAVIS-R50 online, dense-interp, round-2 ids (seed 43) & 0.573 [0.29, 0.76] & 0.537 & 0.207 \\
    CAVIS-R50 + offline refiner, dense-nearest & 0.489 [0.20, 0.75] & 0.451 & 0.198 \\
    CAVIS-R50 online, dense-nearest & 0.586 [0.33, 0.79] & 0.579 & 0.244 \\
    CAVIS-R50 online, dense-nearest, round-2 ids (rerun) & 0.504 [0.20, 0.76] & 0.556 & 0.217 \\
    CAVIS-R50 + offline refiner, sparse & 0.597 [0.29, 0.78] & 0.517 & 0.200 \\
    CAVIS-R50 + offline refiner, sparse, round-2 ids & 0.592 [0.32, 0.77] & 0.486 & 0.174 \\
    CAVIS-R50 online, sparse (ckpt 12.5k) & 0.538 [0.24, 0.73] & 0.479 & 0.200 \\
    CAVIS-R50 online, sparse, round-2 ids & 0.594 [0.32, 0.77] & 0.504 & 0.182 \\
    CAVIS-R50 online, sparse (final) & 0.576 [0.31, 0.76] & 0.465 & 0.186 \\
    CAVIS-Swin-L + offline refiner, sparse & 0.630 [0.36, 0.81] & 0.610 & 0.229 \\
    CAVIS-Swin-L + offline refiner, sparse, round-2 ids & 0.581 [0.30, 0.77] & 0.416 & 0.146 \\
    CAVIS-Swin-L online, sparse & 0.603 [0.35, 0.80] & 0.612 & 0.238 \\
    CAVIS-Swin-L online, sparse, round-2 ids & 0.546 [0.29, 0.75] & 0.559 & 0.223 \\
    DVIS++-R50, segmenter stage, sparse & 0.517 [0.31, 0.65] & 0.464 & 0.180 \\
    Video-Mask2Former-R50, sparse & 0.568 [0.31, 0.72] & 0.580 & 0.241 \\
    MinVIS-R50, curated & 0.567 [0.26, 0.76] & 0.547 & 0.218 \\
    MinVIS-R50, dense-interp (final) & 0.593 [0.29, 0.78] & 0.598 & 0.240 \\
    MinVIS-R50, dense-interp (ckpt 4k) & 0.553 [0.29, 0.77] & 0.552 & 0.240 \\
    MinVIS-R50, dense-nearest (ckpt 4k) & 0.586 [0.28, 0.78] & 0.592 & 0.251 \\
    MinVIS-R50, dense-nearest (ckpt 8k) & 0.585 [0.36, 0.73] & 0.529 & 0.208 \\
    MinVIS-R50, dense-nearest, round-2 ids (rerun) & 0.536 [0.28, 0.72] & 0.530 & 0.202 \\
    MinVIS-R50, trained on RipVIS labels & 0.602 [0.30, 0.78] & 0.576 & 0.228 \\
    MinVIS-R50, sparse, 24k schedule & 0.581 [0.27, 0.78] & 0.570 & 0.240 \\
    MinVIS-R50, sparse (ckpt 8k) & 0.553 [0.26, 0.74] & 0.567 & 0.238 \\
    MinVIS-R50, sparse (final) & 0.568 [0.26, 0.76] & 0.566 & 0.225 \\
    MinVIS-R50, sparse, round-2 ids & 0.591 [0.27, 0.79] & 0.585 & 0.243 \\
    MinVIS-Swin-L, sparse & 0.588 [0.35, 0.78] & 0.595 & 0.226 \\
    MinVIS-Swin-L, sparse, round-2 ids & 0.589 [0.36, 0.79] & 0.605 & 0.210 \\
    \bottomrule
  \end{tabular}}
\end{table}


\section{Dataset Summary}
\label{sec:dataset_summary}

RipBench comprises $304$ videos, evenly split between $152$ rip current and $152$ non-rip videos, totaling $303{,}491$ frames (see Table \ref{tab:dataset_sum_annex}). The dataset is curated to ensure both class balance and substantial diversity along three dimensions that affect model robustness: spatial resolution, frame rate, and camera viewpoint. Of the total frame count, rip current videos contribute approximately $173{,}109$ frames and non-rip videos approximately $130{,}382$ frames. The total duration is approximately $2$ hours and $42$ minutes $32$ seconds, with rip and non-rip segments contributing roughly $1{:}32{:}12$ and $1{:}10{:}20$ respectively. File sizes reflect these differences, with rip videos totaling approximately $18.61$~GB and non-rip videos approximately $9.06$~GB.

To reflect real-world variability, the dataset spans a broad range of frame rates and resolutions. Most videos are recorded at $30$~FPS ($29.97$--$30.01$), with additional videos at $23.98$, $24$, $25$, $50$ and $60$~FPS ($59.94$) (see Table \ref{tab:fps_distribution}). Spatial resolution ranges from low-resolution formats (e.g., $360 \times 640$, $480 \times 360$) to ultra-high resolutions such as $3840 \times 2160$ and $4096 \times 2160$ (see Table \ref{tab:resolution_distribution} for details). This heterogeneity ensures that models trained and evaluated on RipBench encounter varied acquisition conditions rather than overfitting to a narrow capture profile.

The video-level split contains $165$ training, $68$ validation and $71$ test videos ($91$, $30$ and $31$ rip-positive, respectively). Splits were assigned manually by the rip current experts and measure within-pool generalization (Section~\ref{sec:discussion_and_limitations}). Annotations are reflected at the frame level: of the $51{,}011$ sampled frames, $29{,}415$ contain rip-current annotations ($41{,}547$ instances; a further $2{,}466$ standalone stills with $2{,}577$ instances form the opt-in \texttt{additional\_data} training scenario, Appendix~\ref{sec:hyperparameter_tuning}), and the remaining frames, from non-rip videos and from the $344$ frames of rip videos in which the rip leaves the field of view, serve as negative examples. Negative-sample frames are annotated with semantic segmentation labels for water, land, and sky and contain no rip masks, providing contrastive examples for segmentation models. All annotations are aligned to the same frame index structure, enabling multi-task training and consistent per-frame evaluation across model types.

\begin{figure}[!htb]
\centering

\begin{minipage}[t]{0.48\textwidth}
    \captionof{table}{Video dataset summary for rip and non-rip videos.}
    \label{tab:dataset_sum_annex}
    \begin{tabular}{lcc}
    \toprule
    \textbf{Metric} & \textbf{Rip} & \textbf{Non-Rip} \\
    \midrule
    Total Videos    & 152     & 152     \\
    Total Frames    & 173,109 & 130,382 \\
    Duration        & 1:32:12 & 1:10:20 \\
    File Size       & 18.61 GB & 9.06 GB  \\
    \bottomrule
    \end{tabular}
    \vspace{1em}
    
    \captionof{table}{FPS distribution across rip and non-rip videos.}
    \label{tab:fps_distribution}
    \begin{tabular}{lcc}
    \toprule
    \textbf{FPS} & \textbf{Rip} & \textbf{Non-Rip} \\
    \midrule
      29.97 fps                       &          91     &     55\\
      30 fps                          &          30     &     46\\
      25 fps                          &           8     &     19\\
      59.94 fps                       &          14     &     10\\
      23 fps                          &           7     &     12\\
      24 fps                          &           1     &      6\\
      60 fps                          &           0     &      3\\
      50 fps                          &           1     &      1\\
    \bottomrule
    \end{tabular}
\end{minipage}
\hfill
\begin{minipage}[t]{0.48\textwidth}
    \captionof{table}{Resolution distribution across rip and non-rip videos.}
    \label{tab:resolution_distribution}
    \begin{tabular}{lcc}
    \toprule
    \textbf{Resolution} & \textbf{Rip} & \textbf{Non-Rip} \\
    \midrule
      1920x1080                       &          47      &    38\\     
      3840x2160                       &          25      &    58\\
      1280x720                        &          48      &    15\\
      2560x1440                       &           2      &    21\\
      1080x1920                       &           8      &     9\\
      2720x1530                       &          10      &     0\\
      2160x3840                       &           1      &     3\\
      4096x2160                       &           1      &     3\\
      480x360                         &           3      &     0\\
      1440x2560                       &           0      &     2\\
      1280x676                        &           2      &     0\\
      360x640                         &           2      &     0\\
      768x432                         &           0      &     1\\
      2562x1440                       &           0      &     1\\
      2048x1080                       &           0      &     1\\
      720x1280                        &           1      &     0\\
      608x1080                        &           1      &     0\\
      2730x1440                       &           1      &     0\\
    \bottomrule
    \end{tabular}
\end{minipage}

\end{figure}

\begin{figure*}
\centering
\setlength{\tabcolsep}{1pt}
\begin{tabular}{c c c c c}
    
 \vspace{-1mm}
\includegraphics[width=0.245\textwidth,height=0.15\textwidth,clip]{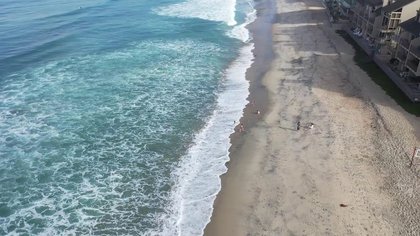} &
\includegraphics[width=0.245\textwidth,height=0.15\textwidth,clip]{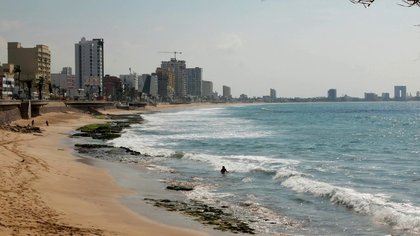} &
\includegraphics[width=0.245\textwidth,height=0.15\textwidth,clip]{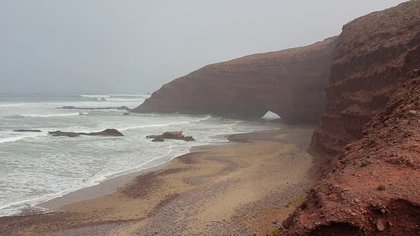} &
\includegraphics[width=0.245\textwidth,height=0.15\textwidth,clip]{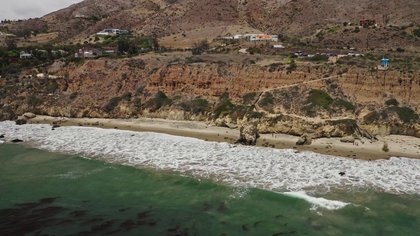}
\tabularnewline

\vspace{-1mm}
\includegraphics[width=0.245\textwidth,height=0.15\textwidth,clip]{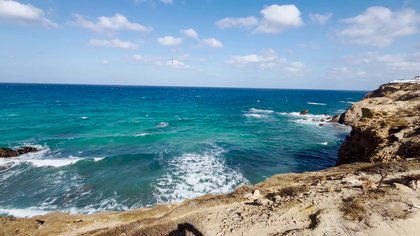} &
\includegraphics[width=0.245\textwidth,height=0.15\textwidth,clip]{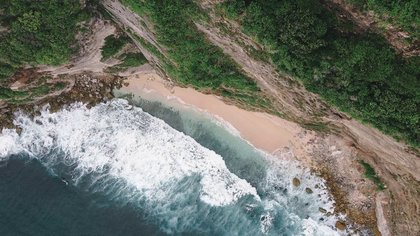} &
\includegraphics[width=0.245\textwidth,height=0.15\textwidth,clip]{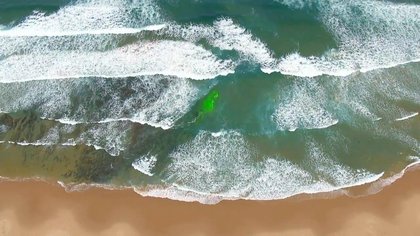} &
\includegraphics[width=0.245\textwidth,height=0.15\textwidth,clip]{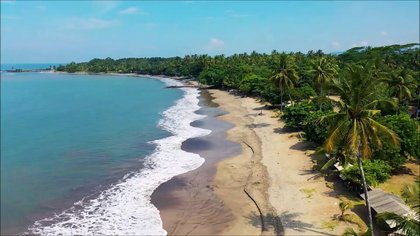}
\tabularnewline

\vspace{-1mm}
\includegraphics[width=0.245\textwidth,height=0.15\textwidth,clip]{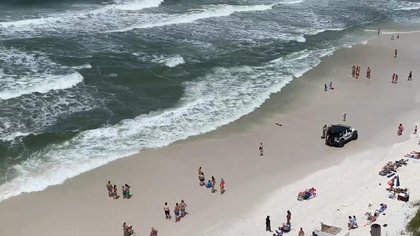} &
\includegraphics[width=0.245\textwidth,height=0.15\textwidth,clip]{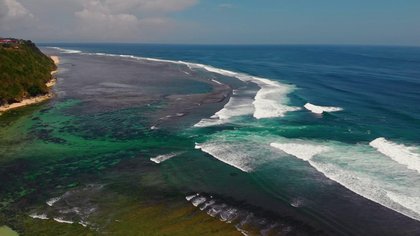} &
\includegraphics[width=0.245\textwidth,height=0.15\textwidth,clip]{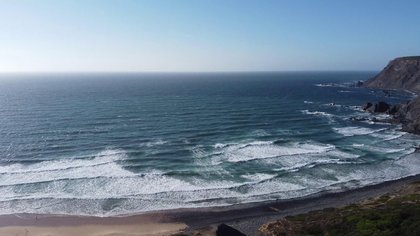} &
\includegraphics[width=0.245\textwidth,height=0.15\textwidth,clip]{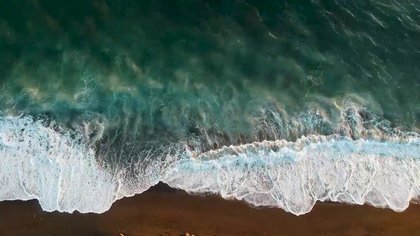}
\tabularnewline

\vspace{-1mm}
\includegraphics[width=0.245\textwidth,height=0.15\textwidth,clip]{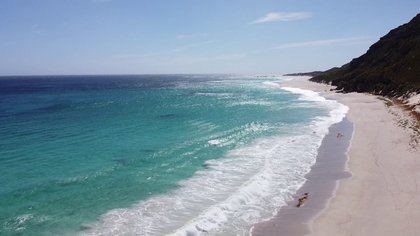} &
\includegraphics[width=0.245\textwidth,height=0.15\textwidth,clip]{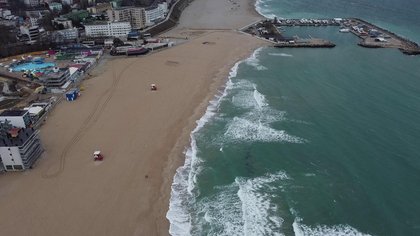} &
\includegraphics[width=0.245\textwidth,height=0.15\textwidth,clip]{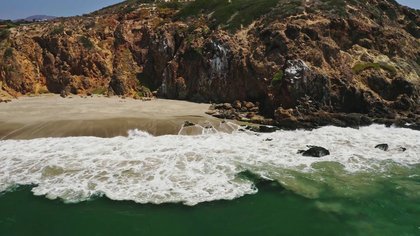} &
\includegraphics[width=0.245\textwidth,height=0.15\textwidth,clip]{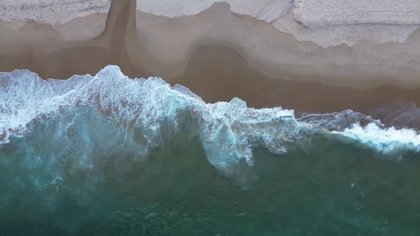}
\tabularnewline

\vspace{-1mm}
\includegraphics[width=0.245\textwidth,height=0.15\textwidth,clip]{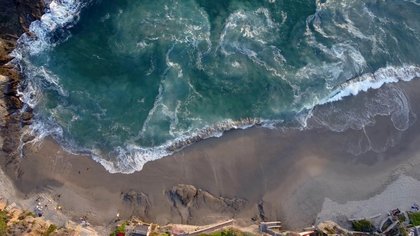} &
\includegraphics[width=0.245\textwidth,height=0.15\textwidth,clip]{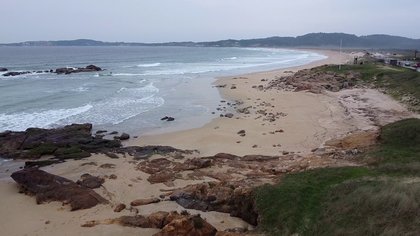} &
\includegraphics[width=0.245\textwidth,height=0.15\textwidth,clip]{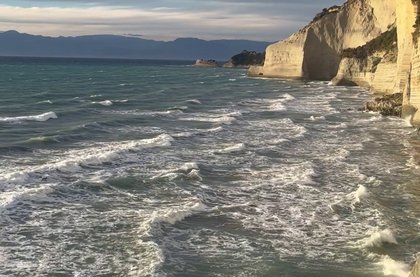} &
\includegraphics[width=0.245\textwidth,height=0.15\textwidth,clip]{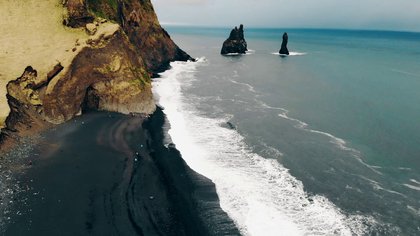}
\tabularnewline

\vspace{-1mm}
\includegraphics[width=0.245\textwidth,height=0.15\textwidth,clip]{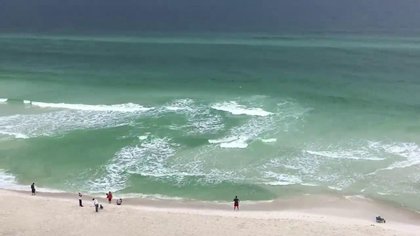} &
\includegraphics[width=0.245\textwidth,height=0.15\textwidth,clip]{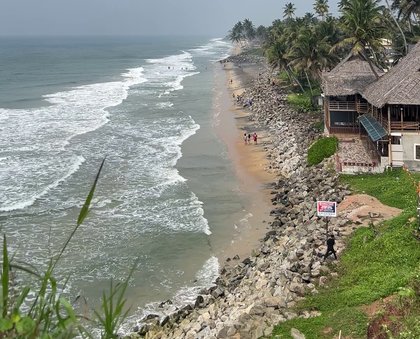} &
\includegraphics[width=0.245\textwidth,height=0.15\textwidth,clip]{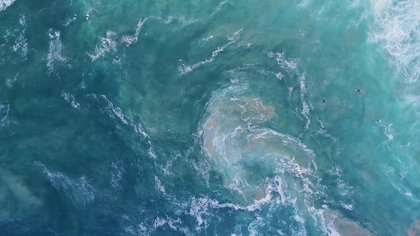} &
\includegraphics[width=0.245\textwidth,height=0.15\textwidth,clip]{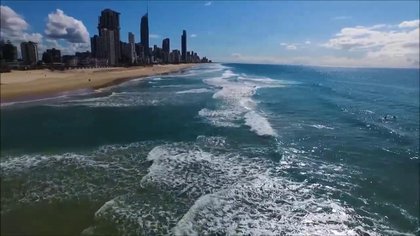}
\tabularnewline

\vspace{-1mm}
\includegraphics[width=0.245\textwidth,height=0.15\textwidth,clip]{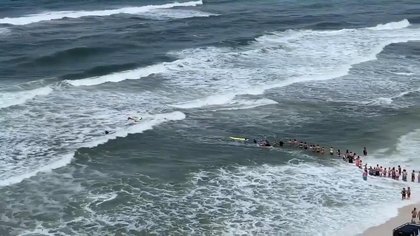} &
\includegraphics[width=0.245\textwidth,height=0.15\textwidth,clip]{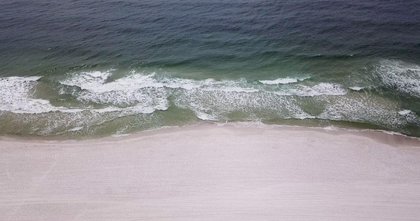} &
\includegraphics[width=0.245\textwidth,height=0.15\textwidth,clip]{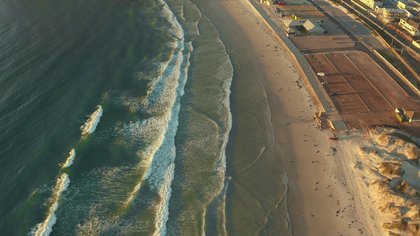} &
\includegraphics[width=0.245\textwidth,height=0.15\textwidth,clip]{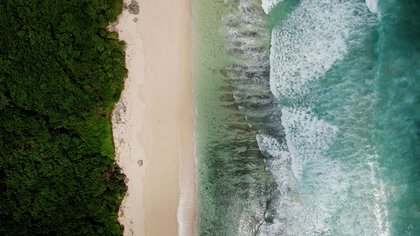}
\tabularnewline

\vspace{-1mm}
\includegraphics[width=0.245\textwidth,height=0.15\textwidth,clip]{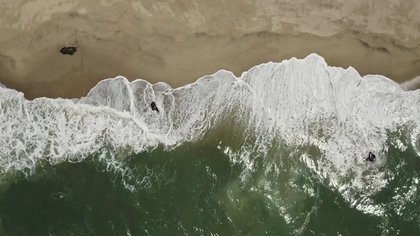} &
\includegraphics[width=0.245\textwidth,height=0.15\textwidth,clip]{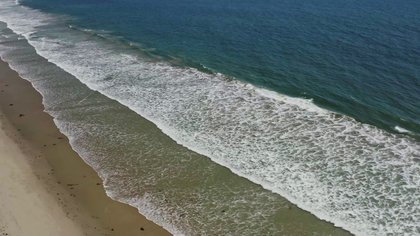} &
\includegraphics[width=0.245\textwidth,height=0.15\textwidth,clip]{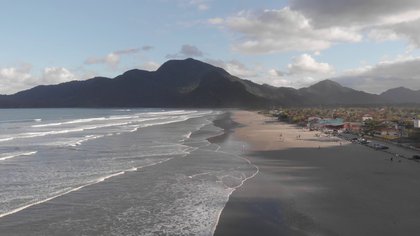} &
\includegraphics[width=0.245\textwidth,height=0.15\textwidth,clip]{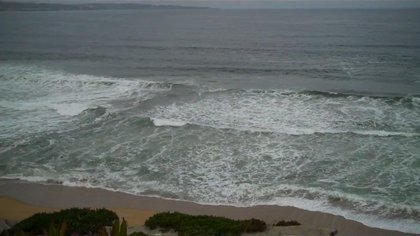}
\tabularnewline

\vspace{-1mm}
\includegraphics[width=0.245\textwidth,height=0.15\textwidth,clip]{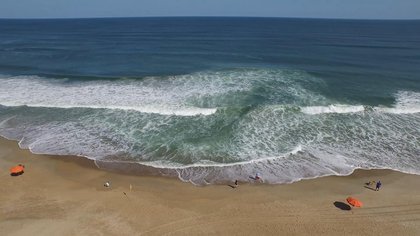} &
\includegraphics[width=0.245\textwidth,height=0.15\textwidth,clip]{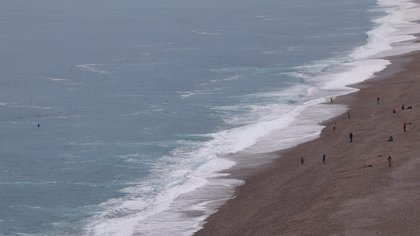} &
\includegraphics[width=0.245\textwidth,height=0.15\textwidth,clip]{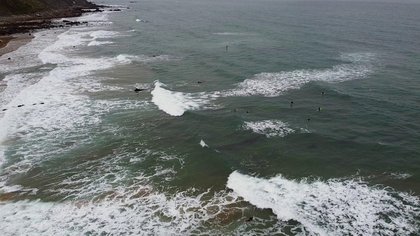} &
\includegraphics[width=0.245\textwidth,height=0.15\textwidth,clip]{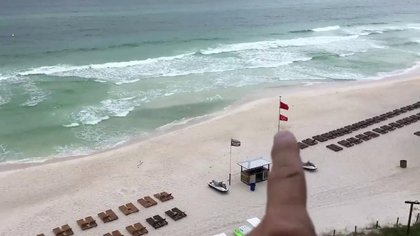}
\tabularnewline

\vspace{-1mm}
\includegraphics[width=0.245\textwidth,height=0.15\textwidth,clip]{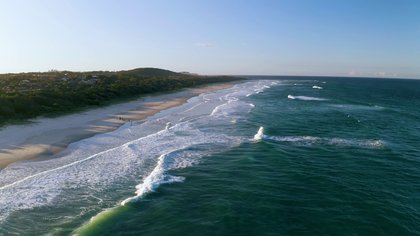} &
\includegraphics[width=0.245\textwidth,height=0.15\textwidth,clip]{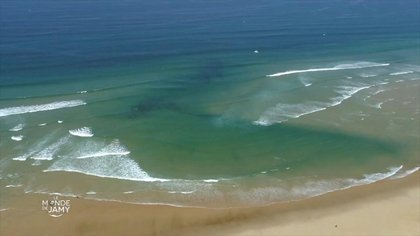} &
\includegraphics[width=0.245\textwidth,height=0.15\textwidth,clip]{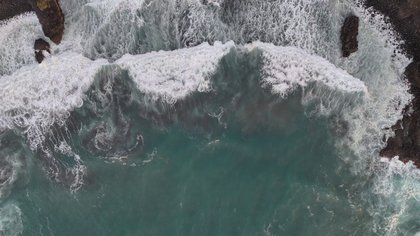} &
\includegraphics[width=0.245\textwidth,height=0.15\textwidth,clip]{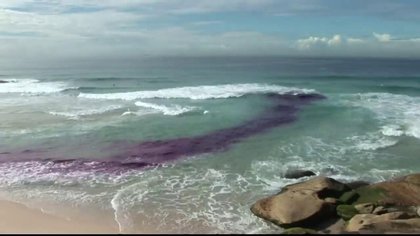}
\tabularnewline

\tabularnewline
     
\end{tabular}
  \caption{Examples from the dataset, showcasing its diverse nature. Here we show frames from 40 randomly selected videos with both rip and no rip frames. \textcolor{red}{\textbf{Can you spot them all?}} Some are easy, while others can be deceiving at first glance. Check Figure \ref{fig:dataset_mixed_annotations} for annotations.}
  \label{fig:dataset_mixed}
\end{figure*}

\begin{figure*}
\centering
\setlength{\tabcolsep}{1pt}

\newcommand{\fixedimg}[1]{%
  \includegraphics[width=0.245\textwidth,height=0.15\textwidth,clip]{#1}%
}

\begin{tabular}{c c c c}

\fixedimg{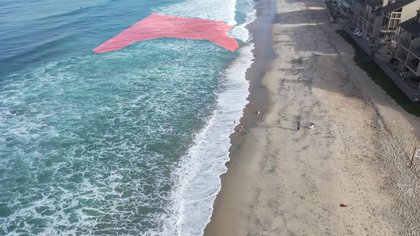} &
\fixedimg{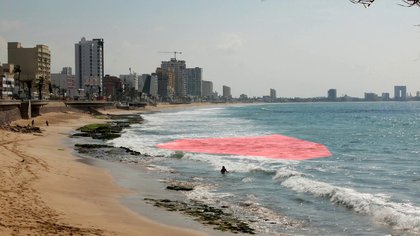} &
\fixedimg{figures/no-rips/RipBench-NR-039_00365.jpg} &
\fixedimg{figures/no-rips/RipBench-NR-077_00438.jpg}
\\[-1mm]

\fixedimg{figures/no-rips/RipBench-NR-030_00180.jpg} &
\fixedimg{figures/no-rips/RipBench-NR-038_00165.jpg} &
\fixedimg{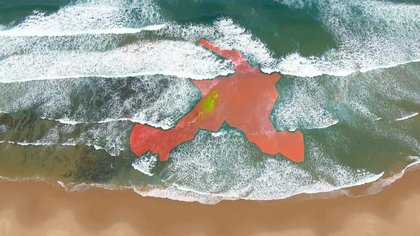} &
\fixedimg{figures/no-rips/RipBench-NR-069_00000.jpg}
\\[-1mm]

\fixedimg{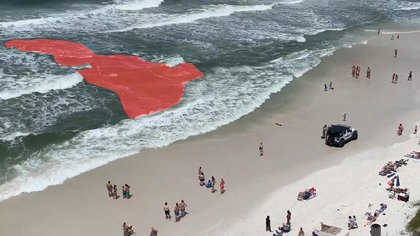} &
\fixedimg{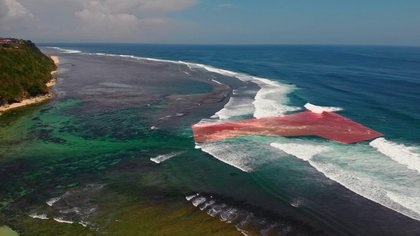} &
\fixedimg{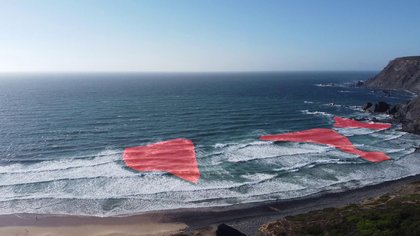} &
\fixedimg{figures/no-rips/RipBench-NR-003_00143.jpg}
\\[-1mm]

\fixedimg{figures/no-rips/RipBench-NR-037_00299.jpg} &
\fixedimg{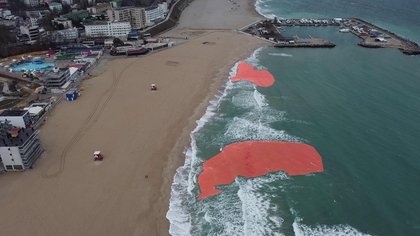} &
\fixedimg{figures/no-rips/RipBench-NR-076_00324.jpg} &
\fixedimg{figures/no-rips/RipBench-NR-128_00125.jpg}
\\[-1mm]

\fixedimg{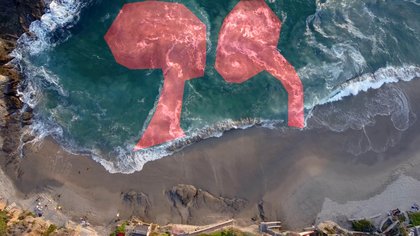} &
\fixedimg{figures/no-rips/RipBench-NR-145_00120.jpg} &
\fixedimg{figures/no-rips/pls.jpg} &
\fixedimg{figures/no-rips/RipBench-NR-056_00000.jpg}
\\[-1mm]

\fixedimg{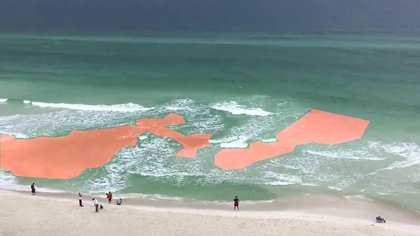} &
\fixedimg{figures/no-rips/RipBench-NR-036_00014.jpg} &
\fixedimg{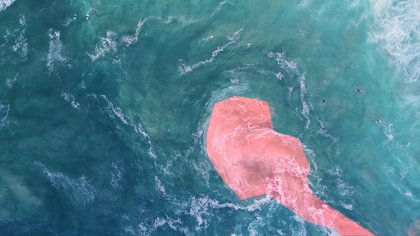} &
\fixedimg{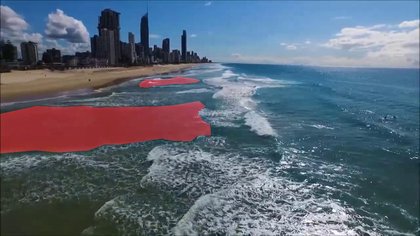}
\\[-1mm]

\fixedimg{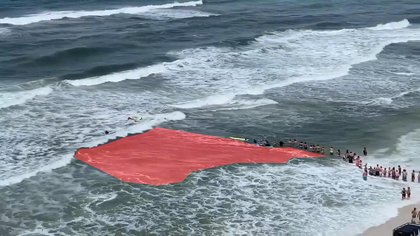} &
\fixedimg{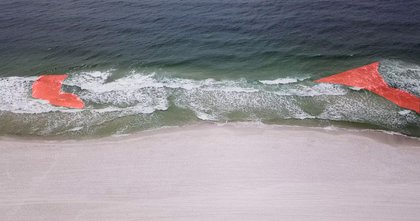} &
\fixedimg{figures/no-rips/RipBench-NR-061_00000.jpg} &
\fixedimg{figures/no-rips/RipBench-NR-060_00285.jpg}
\\[-1mm]

\fixedimg{figures/no-rips/RipBench-NR-074_00216.jpg} &
\fixedimg{figures/no-rips/RipBench-NR-079_00108.jpg} &
\fixedimg{figures/no-rips/RipBench-NR-043_00131.jpg} &
\fixedimg{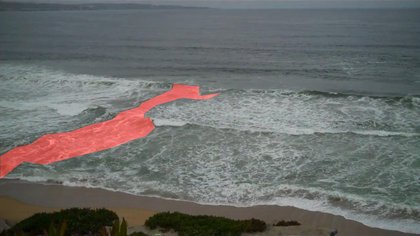}
\\[-1mm]

\fixedimg{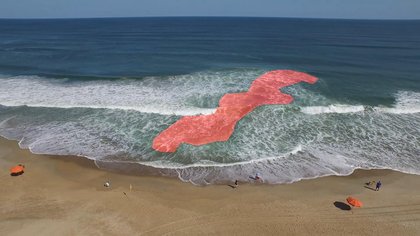} &
\fixedimg{figures/no-rips/RipBench-NR-053_00655.jpg} &
\fixedimg{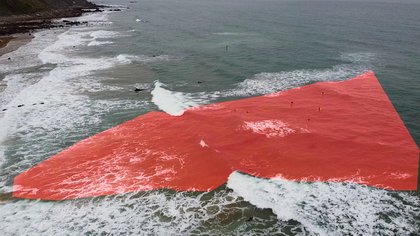} &
\fixedimg{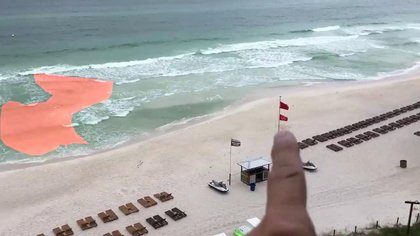}
\\[-1mm]

\fixedimg{figures/no-rips/RipBench-NR-055_00316.jpg} &
\fixedimg{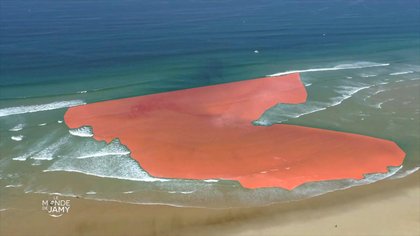} &
\fixedimg{figures/no-rips/RipBench-NR-137_00000.jpg} &
\fixedimg{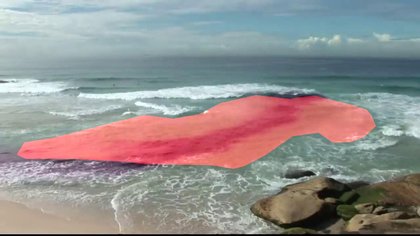}
\\[-1mm]


\end{tabular}

\caption{The same examples as before, with the ground truth masks. How many did you get right?}
\label{fig:dataset_mixed_annotations}
\end{figure*}

\begin{figure*}
\centering
\setlength{\tabcolsep}{1pt}
\begin{tabular}{c c c c c}
    
     \vspace{-1mm} 
     \includegraphics[width=0.245\textwidth]{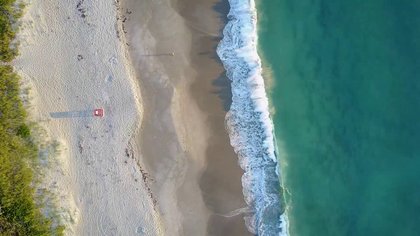} &
     \includegraphics[width=0.245\textwidth]{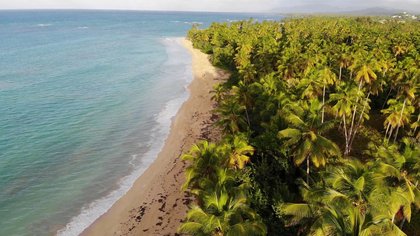} &
     \includegraphics[width=0.245\textwidth]{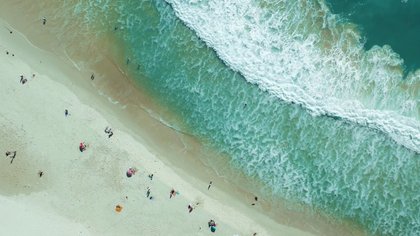} &
     \includegraphics[width=0.245\textwidth]{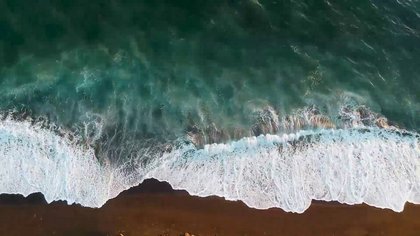} 
     \tabularnewline     
    
     \vspace{-1mm} 
     \includegraphics[width=0.245\textwidth]{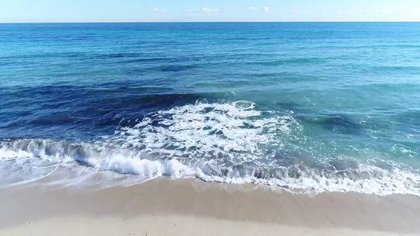} &
     \includegraphics[width=0.245\textwidth]{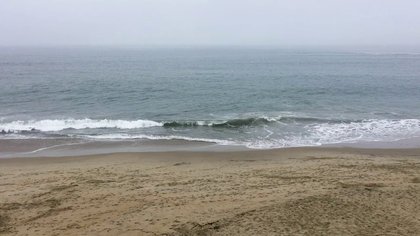} &
     \includegraphics[width=0.245\textwidth]{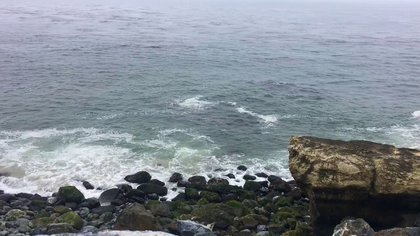} &
     \includegraphics[width=0.245\textwidth]{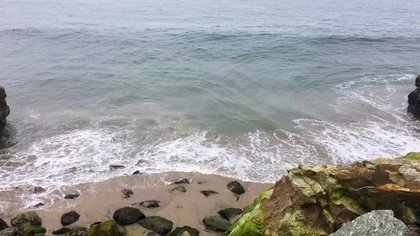} 
     \tabularnewline        
    
     \vspace{-1mm} 
     \includegraphics[width=0.245\textwidth]{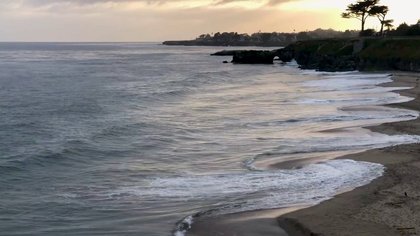} &
     \includegraphics[width=0.245\textwidth]{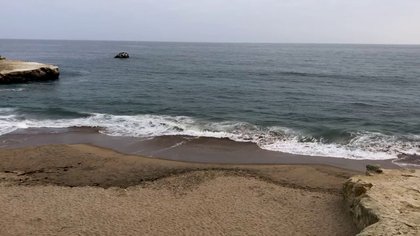} &
     \includegraphics[width=0.245\textwidth]{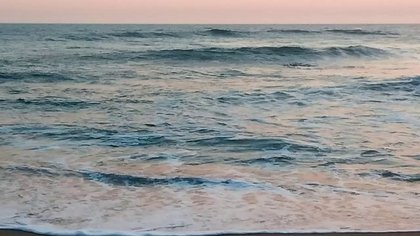} &
     \includegraphics[width=0.245\textwidth]{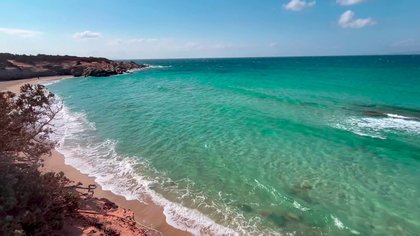} 
     \tabularnewline        
    
     \vspace{-1mm} 
     \includegraphics[width=0.245\textwidth]{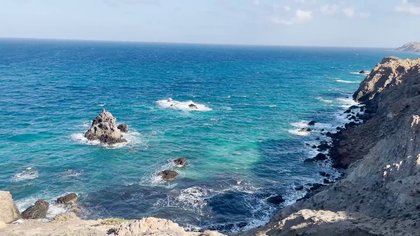} &
     \includegraphics[width=0.245\textwidth]{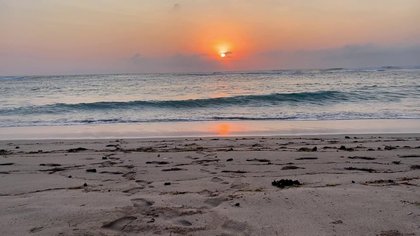} &
     \includegraphics[width=0.245\textwidth]{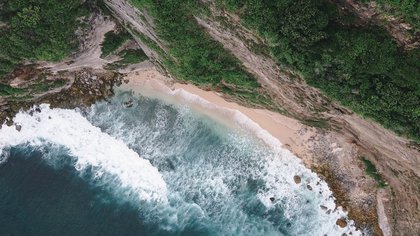} &
     \includegraphics[width=0.245\textwidth]{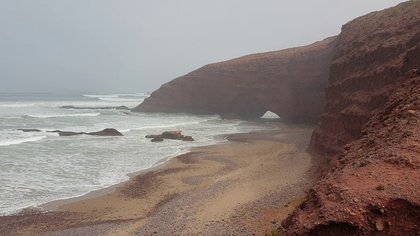} 
     \tabularnewline

     \vspace{-1mm} 
     \includegraphics[width=0.245\textwidth]{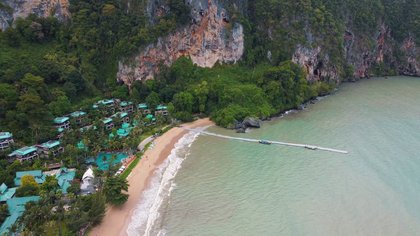} &
     \includegraphics[width=0.245\textwidth]{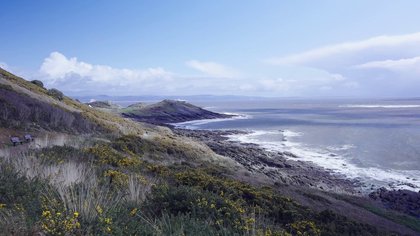} &
     \includegraphics[width=0.245\textwidth]{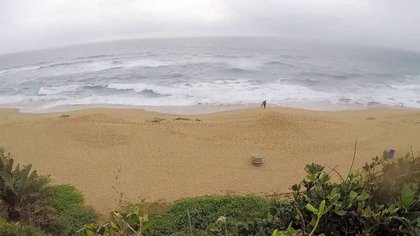} &
     \includegraphics[width=0.245\textwidth]{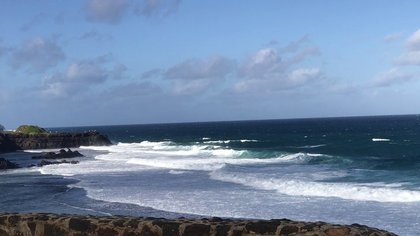} 
     \tabularnewline  
            
     \vspace{-1mm} 
     \includegraphics[width=0.245\textwidth]{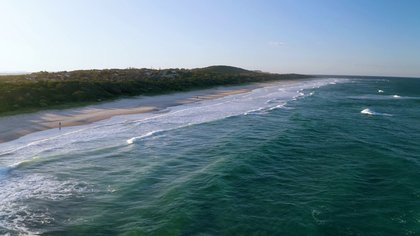} &
     \includegraphics[width=0.245\textwidth]{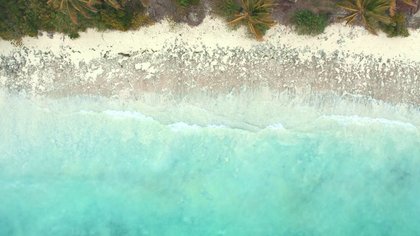} &
     \includegraphics[width=0.245\textwidth]{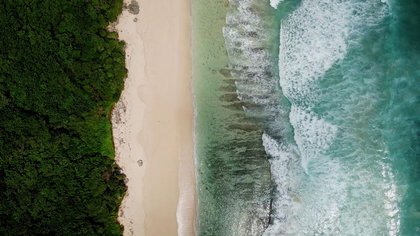} &
     \includegraphics[width=0.245\textwidth]{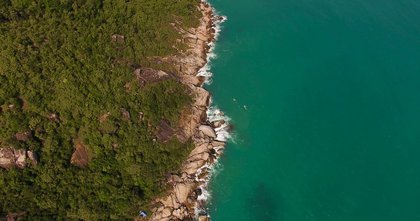} 
     \tabularnewline    
    
     \vspace{-1mm} 
     \includegraphics[width=0.245\textwidth]{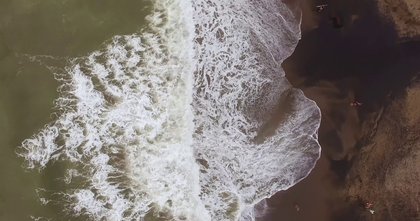} &
     \includegraphics[width=0.245\textwidth]{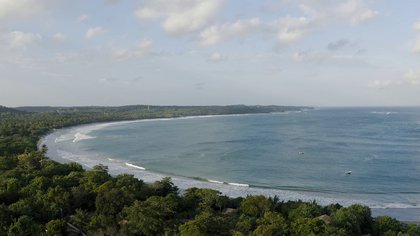} &
     \includegraphics[width=0.245\textwidth]{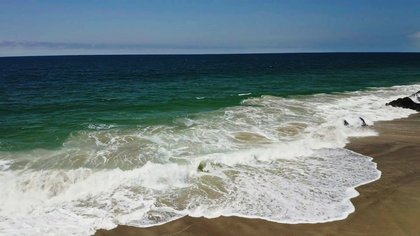} &
     \includegraphics[width=0.245\textwidth]{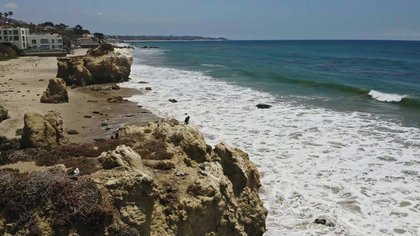} 
     \tabularnewline       
    
     \vspace{-1mm} 
     \includegraphics[width=0.245\textwidth]{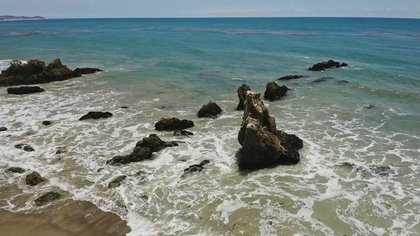} &
     \includegraphics[width=0.245\textwidth]{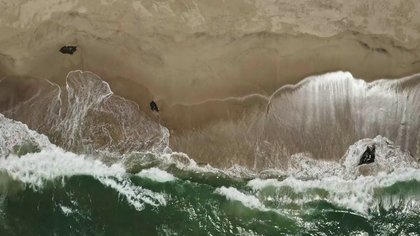} &
     \includegraphics[width=0.245\textwidth]{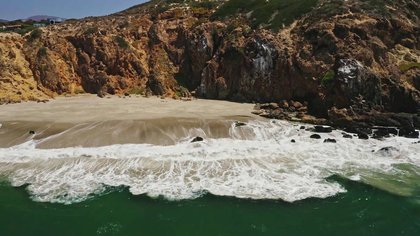} &
     \includegraphics[width=0.245\textwidth]{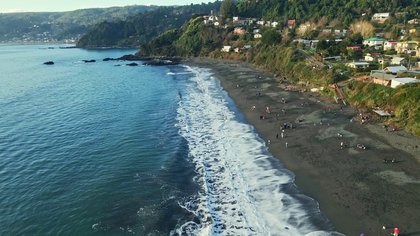} 
     \tabularnewline       
    
     \vspace{-1mm} 
     \includegraphics[width=0.245\textwidth]{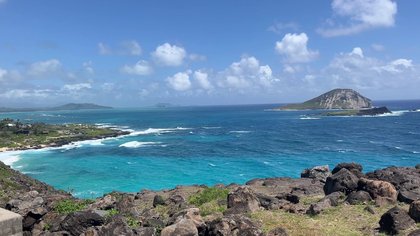} &
     \includegraphics[width=0.245\textwidth]{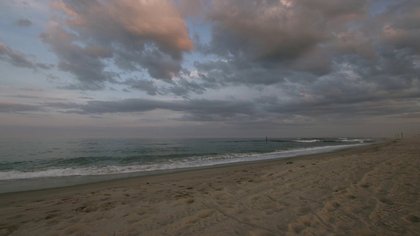} &
     \includegraphics[width=0.245\textwidth]{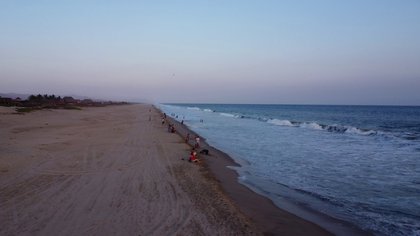} &
     \includegraphics[width=0.245\textwidth]{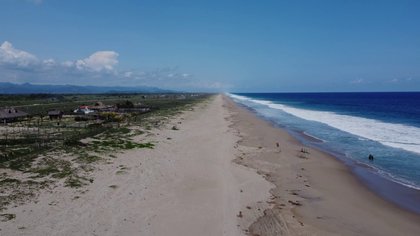} 
     \tabularnewline       
    
     \vspace{-1mm} 
     \includegraphics[width=0.245\textwidth]{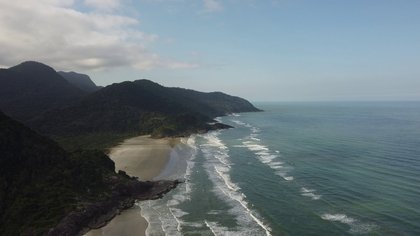} &
     \includegraphics[width=0.245\textwidth]{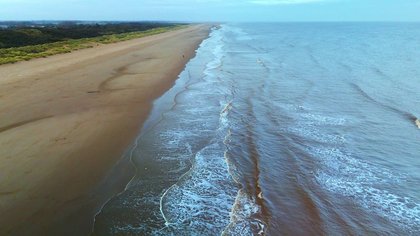} &
     \includegraphics[width=0.245\textwidth]{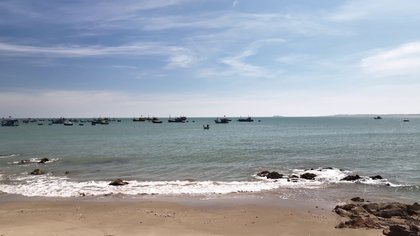} &
     \includegraphics[width=0.245\textwidth]{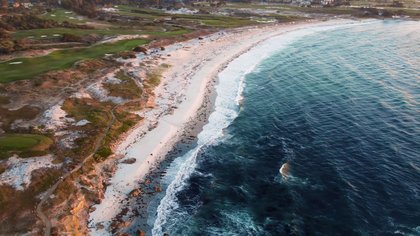} 
     \tabularnewline    
\end{tabular}
  \caption{Examples from our dataset of images without rip currents sampled from 40 videos (out of a total of 152 videos without rips).}
  \label{fig:dataset_no-rips}
\end{figure*}

\begin{figure*}
\centering
\setlength{\tabcolsep}{1pt}
\begin{tabular}{c c c c c}
    
     \vspace{-1mm} 
     \includegraphics[width=0.245\textwidth]{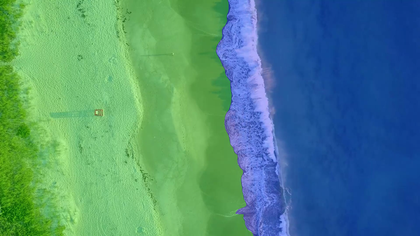} &
     \includegraphics[width=0.245\textwidth]{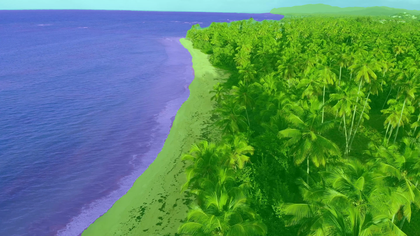} &
     \includegraphics[width=0.245\textwidth]{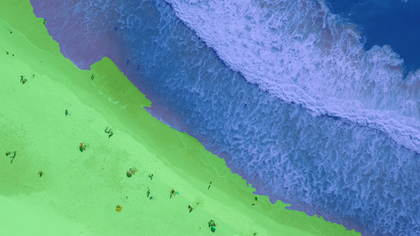} &
     \includegraphics[width=0.245\textwidth]{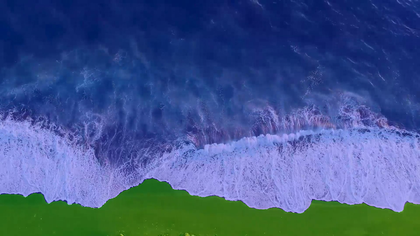} 
     \tabularnewline     
    
     \vspace{-1mm} 
     \includegraphics[width=0.245\textwidth]{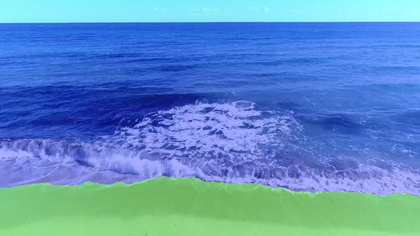} &
     \includegraphics[width=0.245\textwidth]{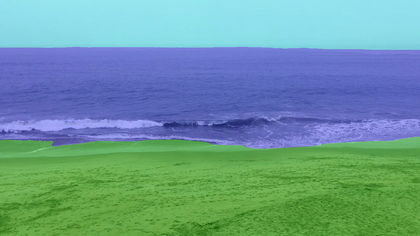} &
     \includegraphics[width=0.245\textwidth]{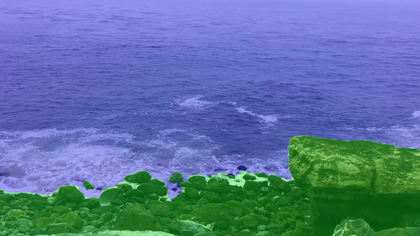} &
     \includegraphics[width=0.245\textwidth]{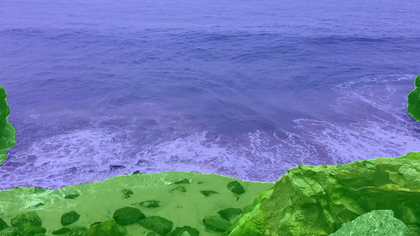} 
     \tabularnewline        
    
     \vspace{-1mm} 
     \includegraphics[width=0.245\textwidth]{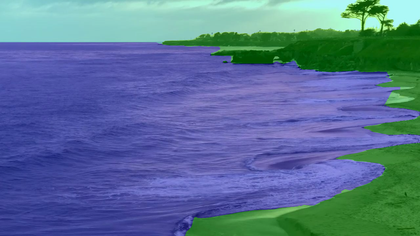} &
     \includegraphics[width=0.245\textwidth]{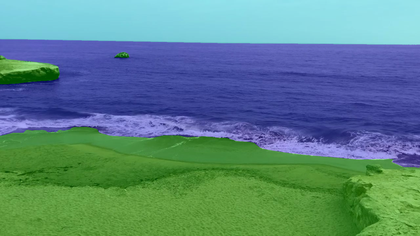} &
     \includegraphics[width=0.245\textwidth]{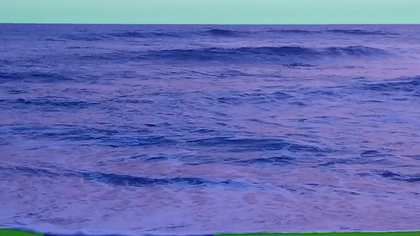} &
     \includegraphics[width=0.245\textwidth]{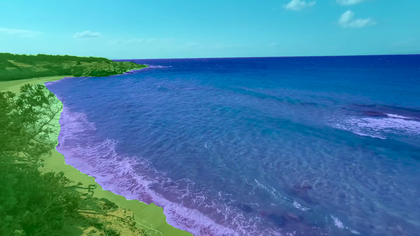} 
     \tabularnewline        
    
     \vspace{-1mm} 
     \includegraphics[width=0.245\textwidth]{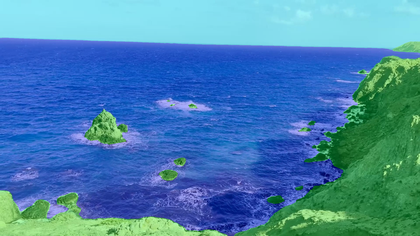} &
     \includegraphics[width=0.245\textwidth]{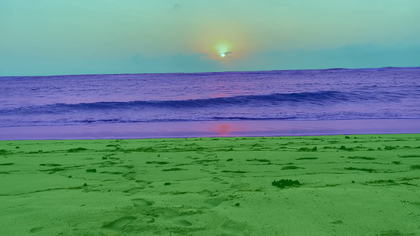} &
     \includegraphics[width=0.245\textwidth]{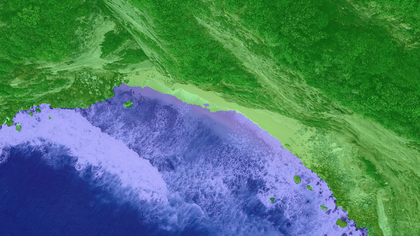} &
     \includegraphics[width=0.245\textwidth]{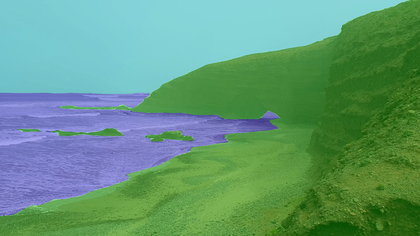} 
     \tabularnewline

     \vspace{-1mm} 
     \includegraphics[width=0.245\textwidth]{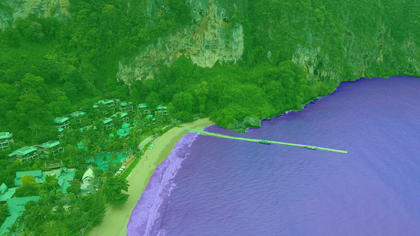} &
     \includegraphics[width=0.245\textwidth]{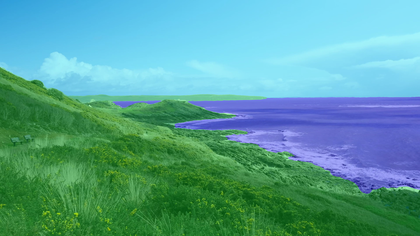} &
     \includegraphics[width=0.245\textwidth]{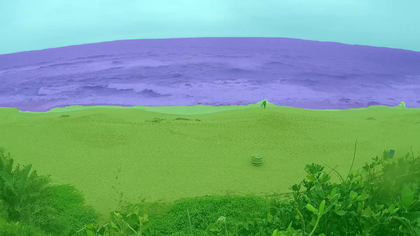} &
     \includegraphics[width=0.245\textwidth]{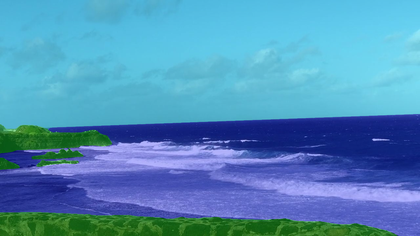} 
     \tabularnewline  
            
     \vspace{-1mm} 
     \includegraphics[width=0.245\textwidth]{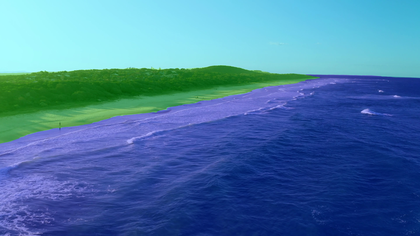} &
     \includegraphics[width=0.245\textwidth]{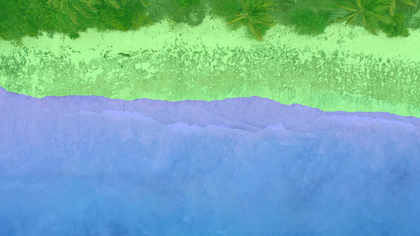} &
     \includegraphics[width=0.245\textwidth]{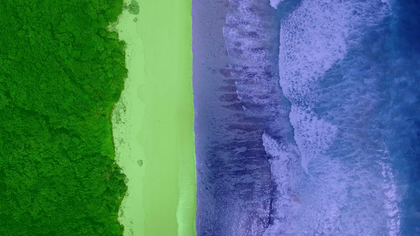} &
     \includegraphics[width=0.245\textwidth]{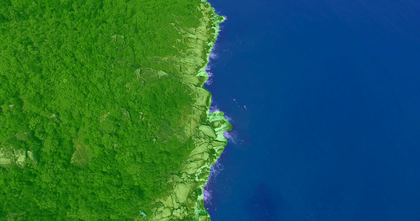} 
     \tabularnewline    
    
     \vspace{-1mm} 
     \includegraphics[width=0.245\textwidth]{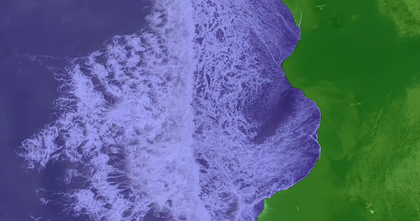} &
     \includegraphics[width=0.245\textwidth]{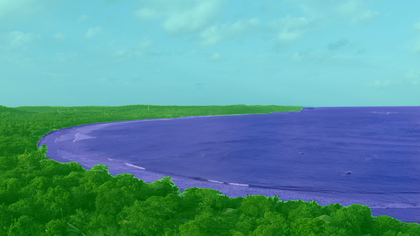} &
     \includegraphics[width=0.245\textwidth]{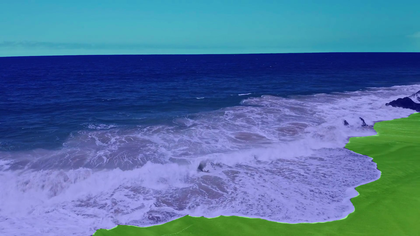} &
     \includegraphics[width=0.245\textwidth]{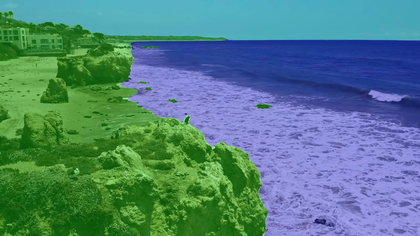} 
     \tabularnewline       
    
     \vspace{-1mm} 
     \includegraphics[width=0.245\textwidth]{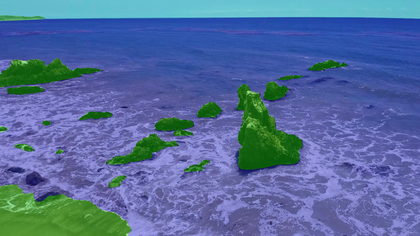} &
     \includegraphics[width=0.245\textwidth]{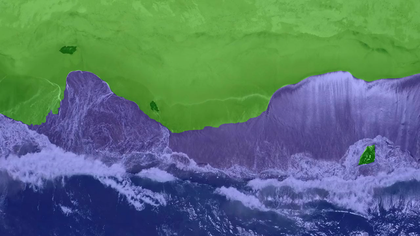} &
     \includegraphics[width=0.245\textwidth]{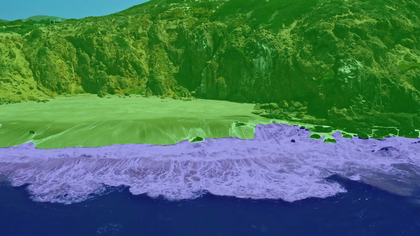} &
     \includegraphics[width=0.245\textwidth]{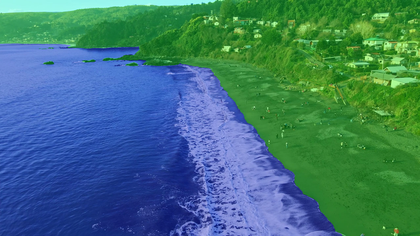} 
     \tabularnewline       
    
     \vspace{-1mm} 
     \includegraphics[width=0.245\textwidth]{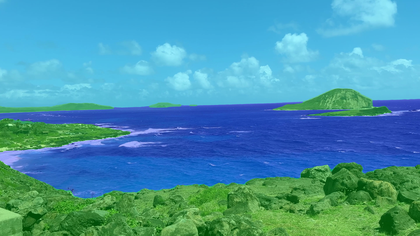} &
     \includegraphics[width=0.245\textwidth]{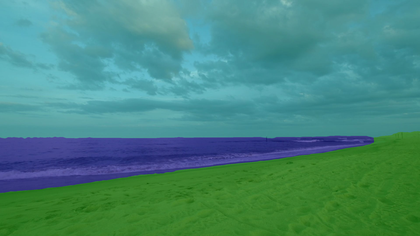} &
     \includegraphics[width=0.245\textwidth]{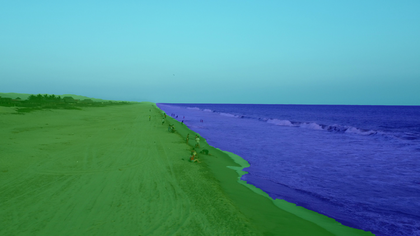} &
     \includegraphics[width=0.245\textwidth]{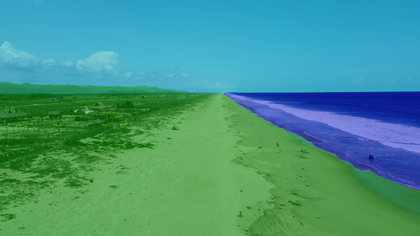} 
     \tabularnewline       
    
     \vspace{-1mm} 
     \includegraphics[width=0.245\textwidth]{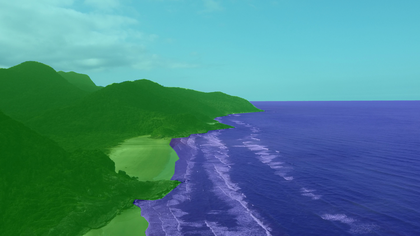} &
     \includegraphics[width=0.245\textwidth]{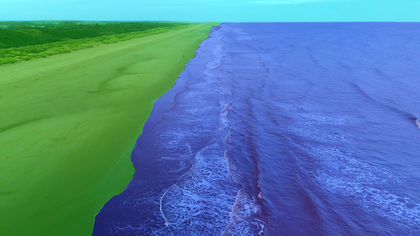} &
     \includegraphics[width=0.245\textwidth]{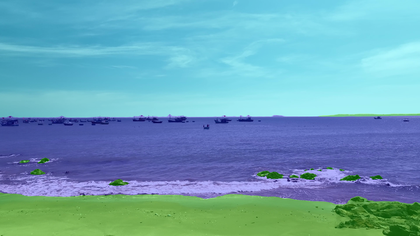} &
     \includegraphics[width=0.245\textwidth]{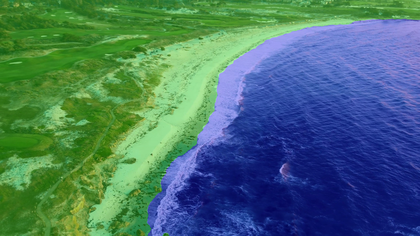} 
     \tabularnewline    
\end{tabular}
  \caption{Examples from our dataset of images without rip currents sampled from 40 videos (out of a total of 152 videos without rips), annotated for \textbf{semantic segmentation} with classes land (\textcolor{green}{green}), water (\textcolor{blue}{blue}) and sky (\textcolor{cyan}{cyan}).}
  \label{fig:dataset_no-rips_semantic}

\end{figure*}

Figures~\ref{fig:dataset_mixed}--\ref{fig:dataset_no-rips_semantic} present representative $4\times10$ grids: frames with and without rips, with and without annotations, and the rip videos with each annotation type (axis-aligned and oriented boxes, instance, semantic and panoptic masks, and all three combined); each caption states what its grid shows.

\providecommand{\variantimg}[1]{%
  \includegraphics[width=0.245\textwidth,height=0.15\textwidth,clip]{#1}%
}

\begin{figure*}
\centering
\setlength{\tabcolsep}{1pt}
\begin{tabular}{c c c c}
\variantimg{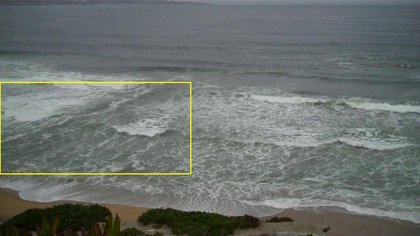} &
\variantimg{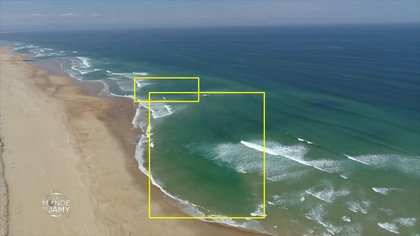} &
\variantimg{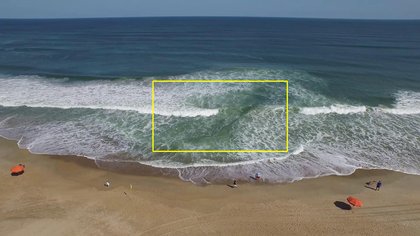} &
\variantimg{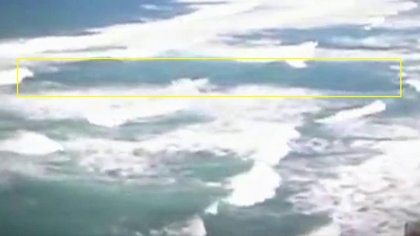}
\\[-1mm]

\variantimg{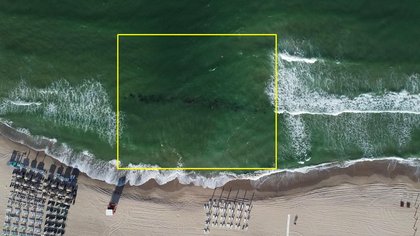} &
\variantimg{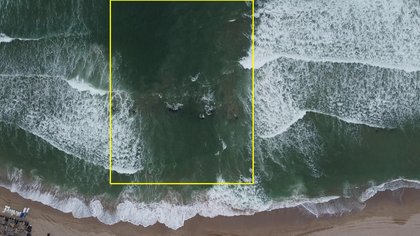} &
\variantimg{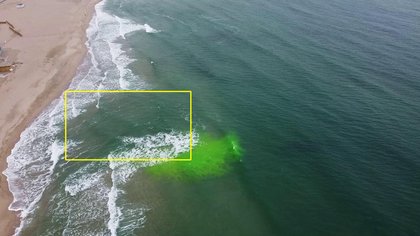} &
\variantimg{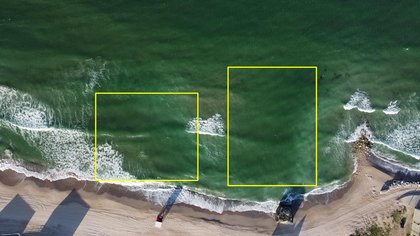}
\\[-1mm]

\variantimg{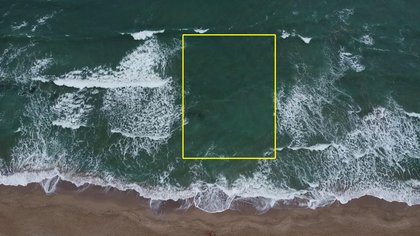} &
\variantimg{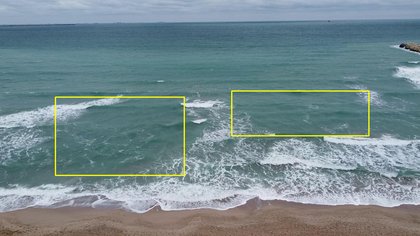} &
\variantimg{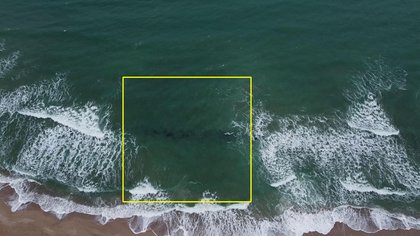} &
\variantimg{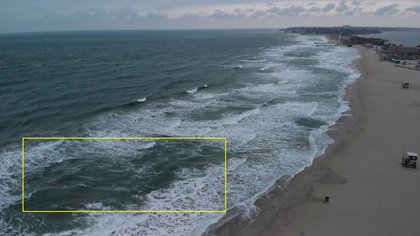}
\\[-1mm]

\variantimg{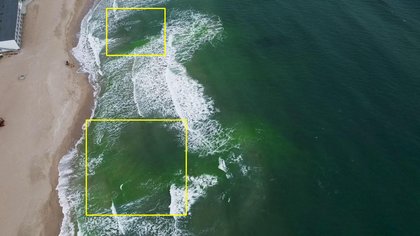} &
\variantimg{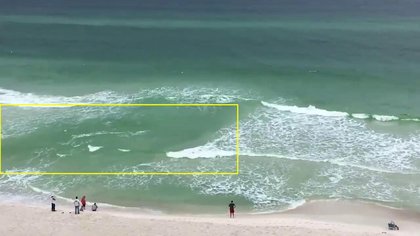} &
\variantimg{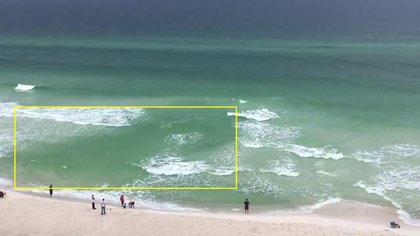} &
\variantimg{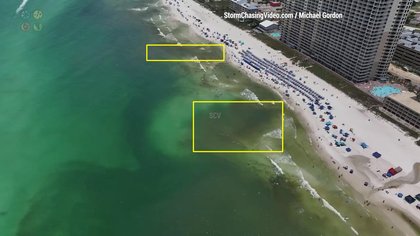}
\\[-1mm]

\variantimg{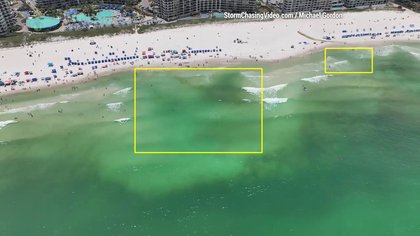} &
\variantimg{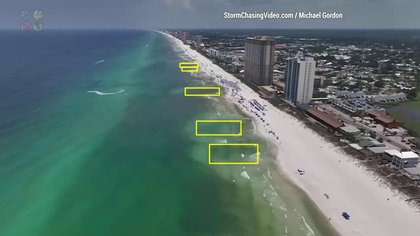} &
\variantimg{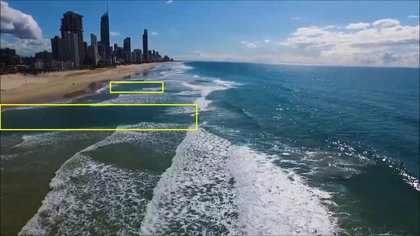} &
\variantimg{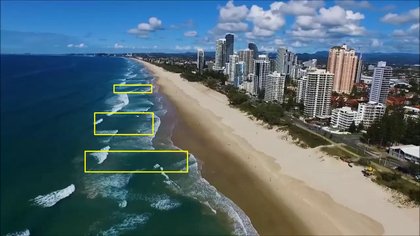}
\\[-1mm]

\variantimg{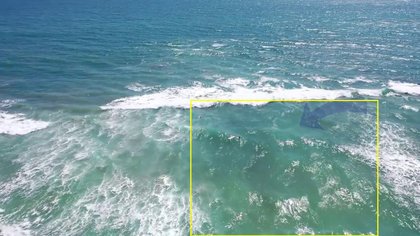} &
\variantimg{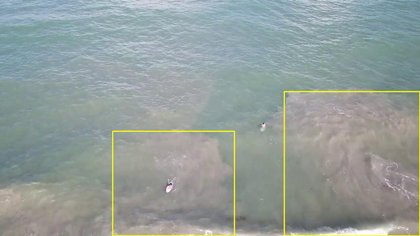} &
\variantimg{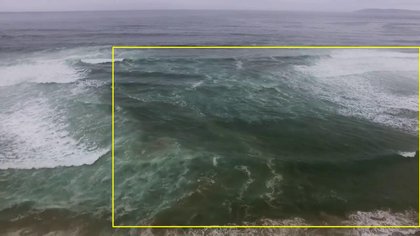} &
\variantimg{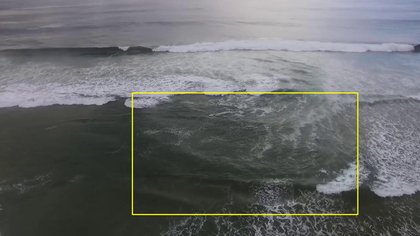}
\\[-1mm]

\variantimg{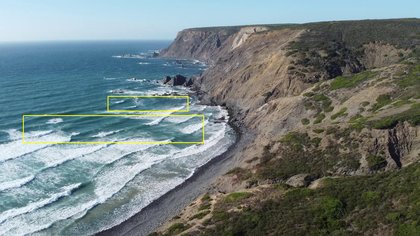} &
\variantimg{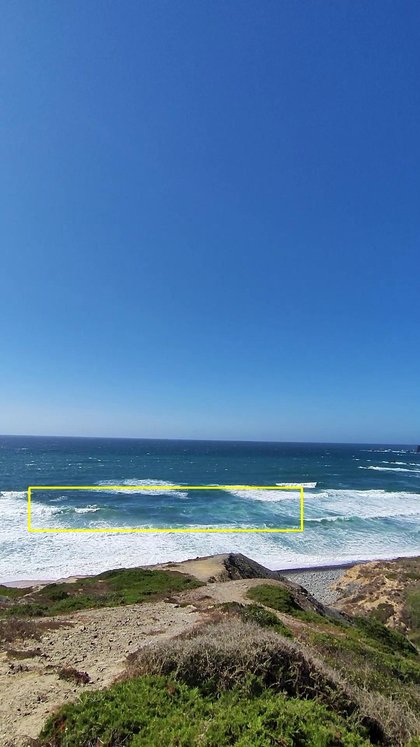} &
\variantimg{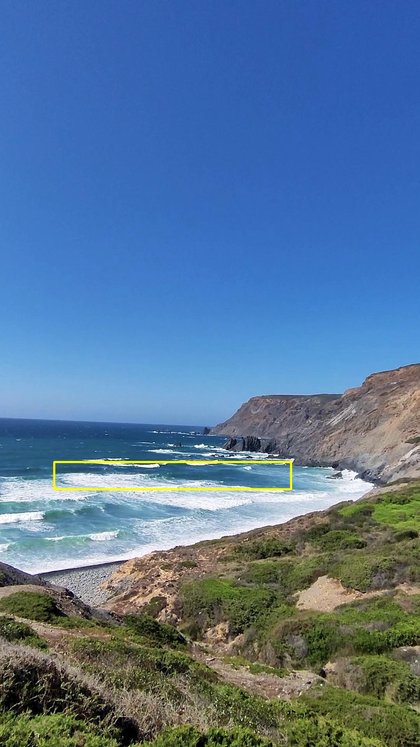} &
\variantimg{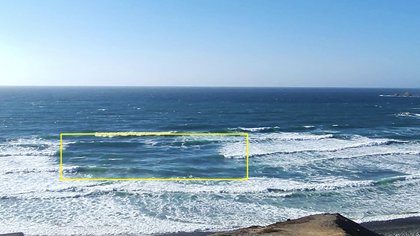}
\\[-1mm]

\variantimg{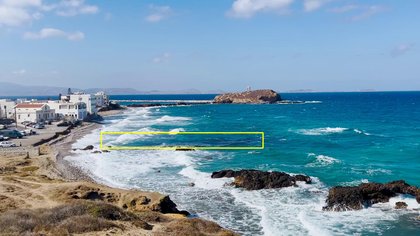} &
\variantimg{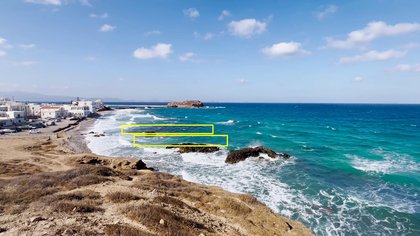} &
\variantimg{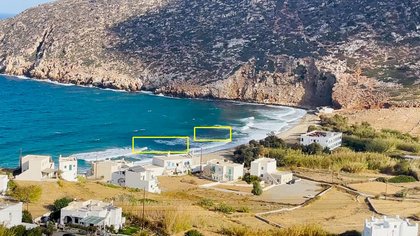} &
\variantimg{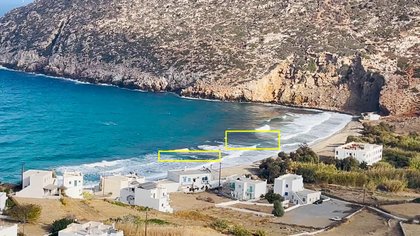}
\\[-1mm]

\variantimg{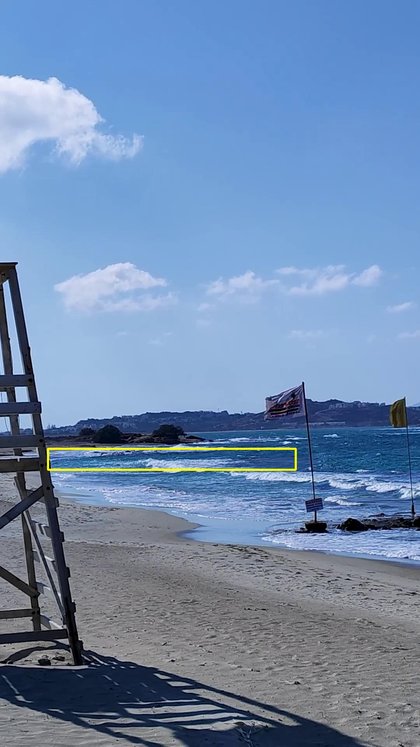} &
\variantimg{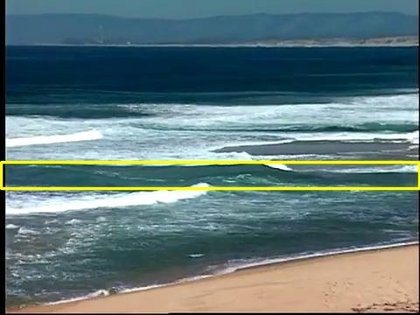} &
\variantimg{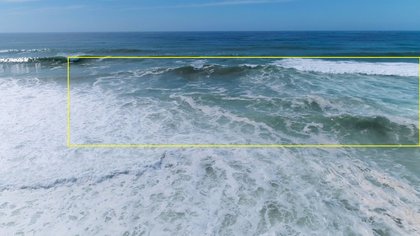} &
\variantimg{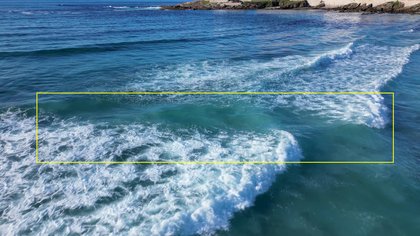}
\\[-1mm]

\variantimg{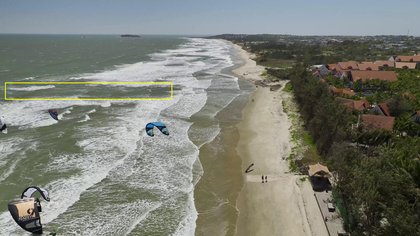} &
\variantimg{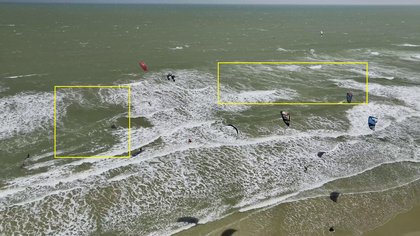} &
\variantimg{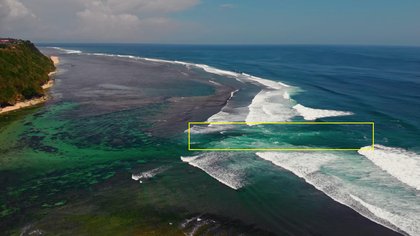} &
\variantimg{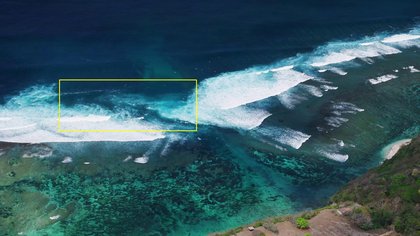}
\end{tabular}
\caption{Examples from our dataset of images with rip currents sampled from 40 videos (out of
a total of 152 videos with rips), with \textcolor{yellow}{yellow} axis-aligned bounding box annotations.}
\label{fig:ripbench-bbox-examples}
\end{figure*}

\begin{figure*}
\centering
\setlength{\tabcolsep}{1pt}
\begin{tabular}{c c c c}
\variantimg{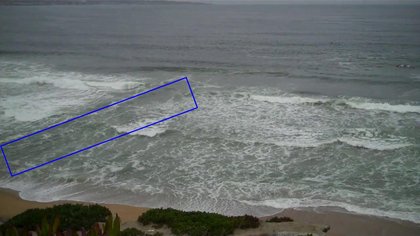} &
\variantimg{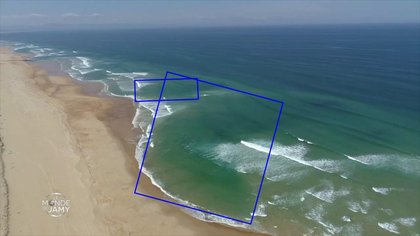} &
\variantimg{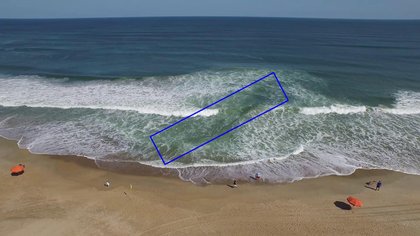} &
\variantimg{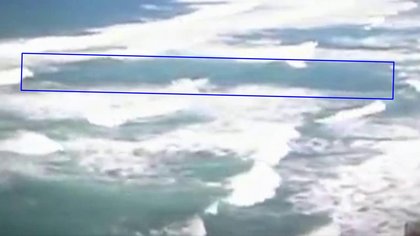}
\\[-1mm]

\variantimg{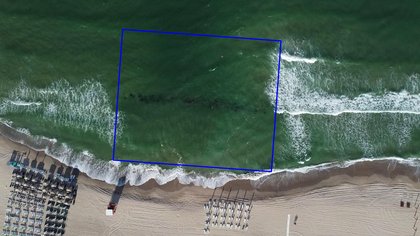} &
\variantimg{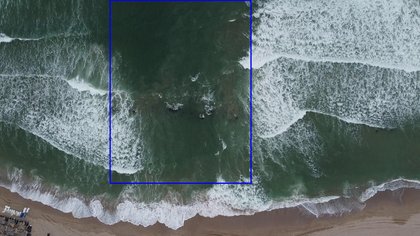} &
\variantimg{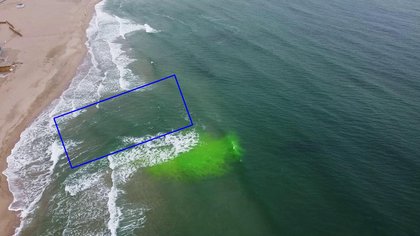} &
\variantimg{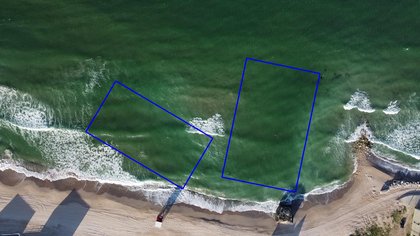}
\\[-1mm]

\variantimg{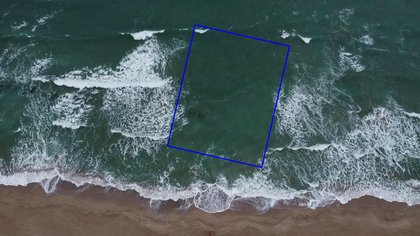} &
\variantimg{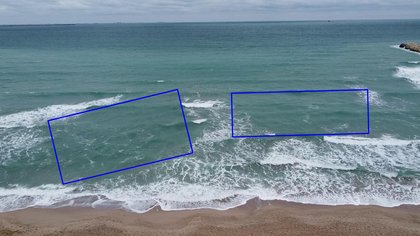} &
\variantimg{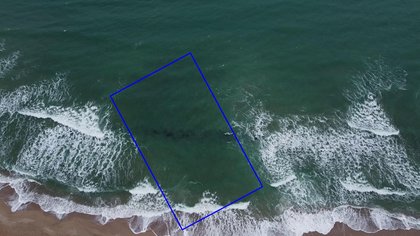} &
\variantimg{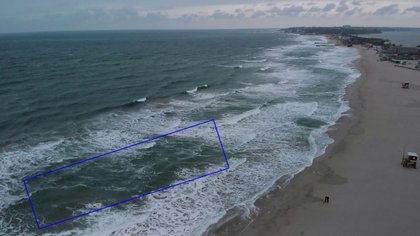}
\\[-1mm]

\variantimg{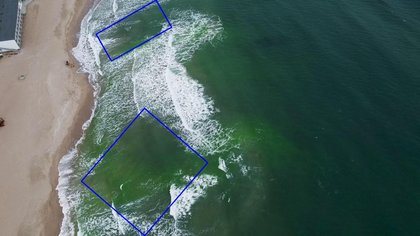} &
\variantimg{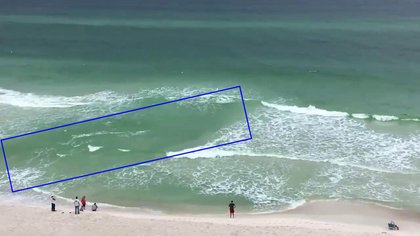} &
\variantimg{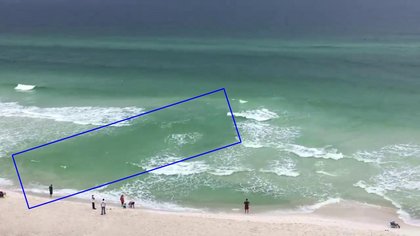} &
\variantimg{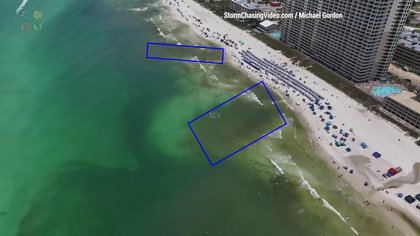}
\\[-1mm]

\variantimg{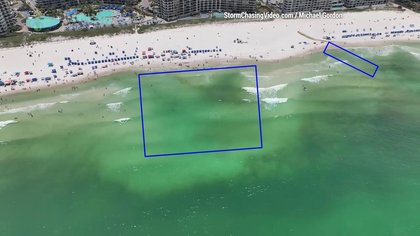} &
\variantimg{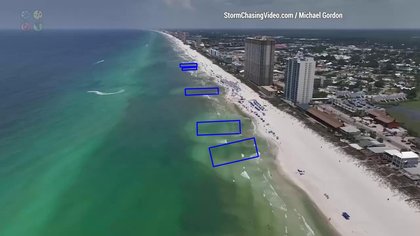} &
\variantimg{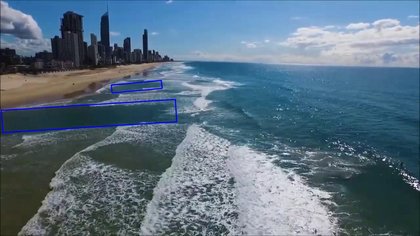} &
\variantimg{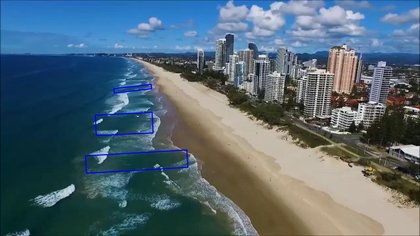}
\\[-1mm]

\variantimg{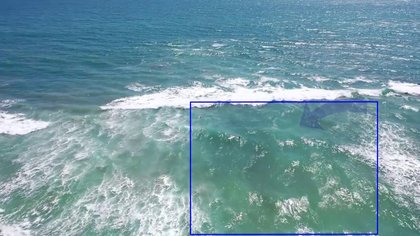} &
\variantimg{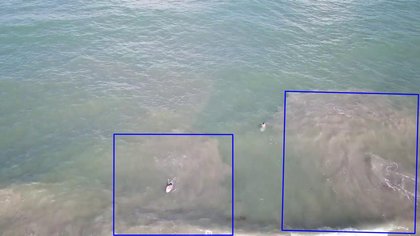} &
\variantimg{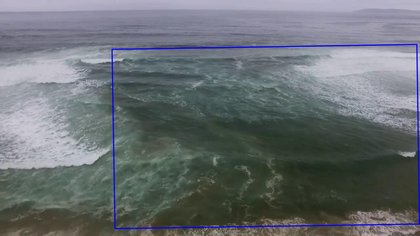} &
\variantimg{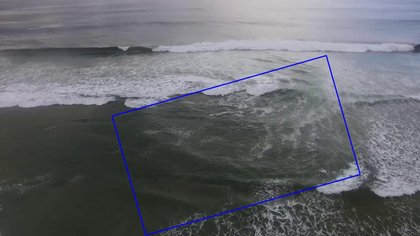}
\\[-1mm]

\variantimg{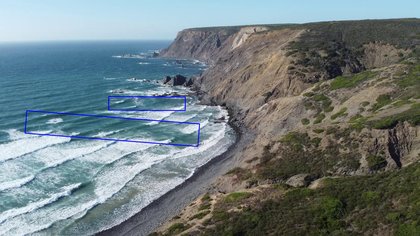} &
\variantimg{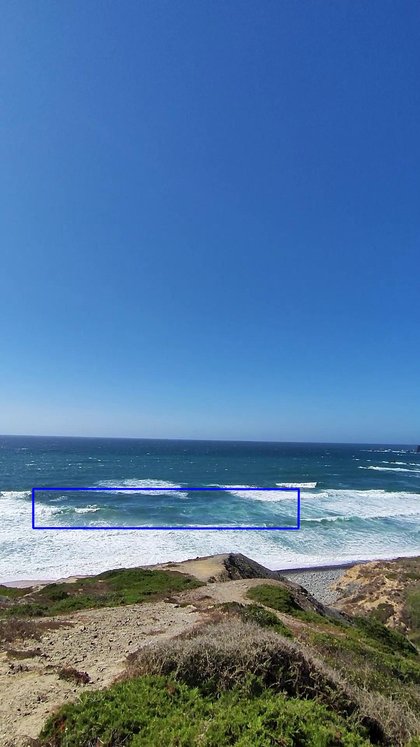} &
\variantimg{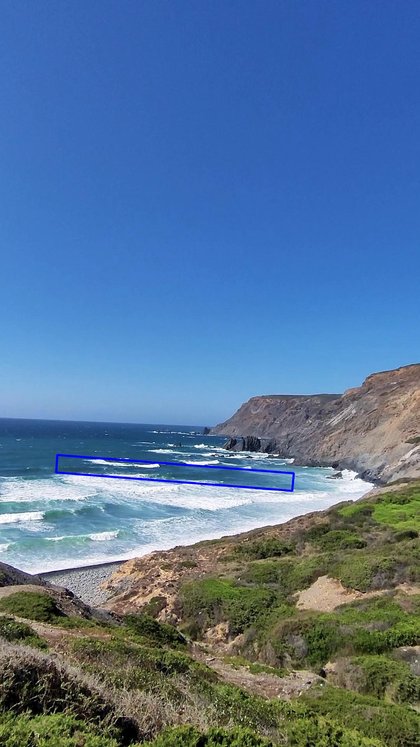} &
\variantimg{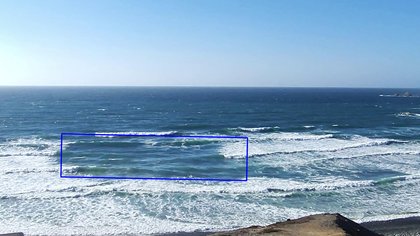}
\\[-1mm]

\variantimg{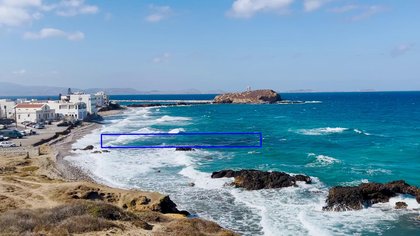} &
\variantimg{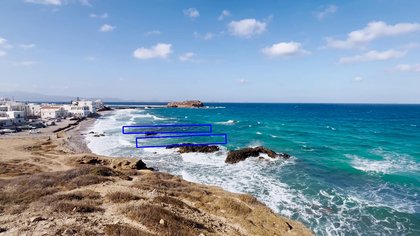} &
\variantimg{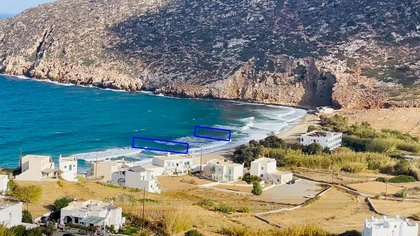} &
\variantimg{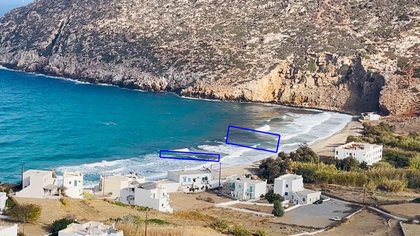}
\\[-1mm]

\variantimg{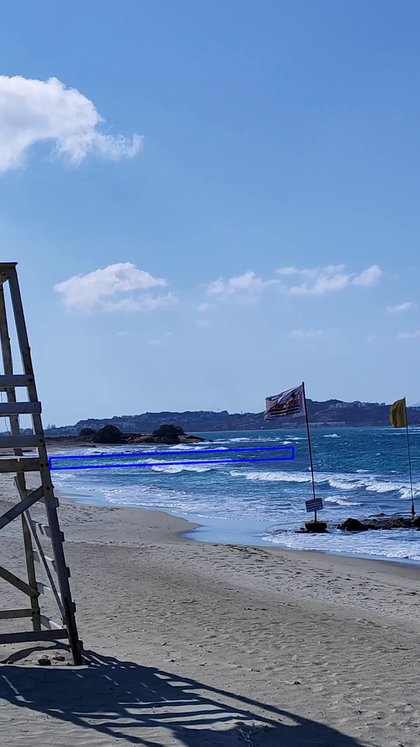} &
\variantimg{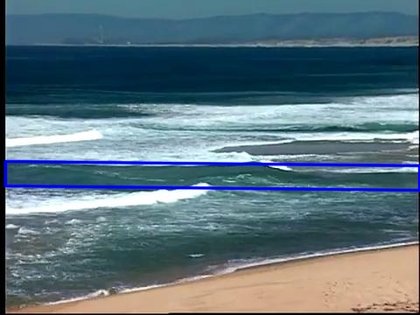} &
\variantimg{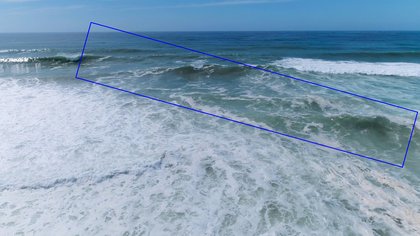} &
\variantimg{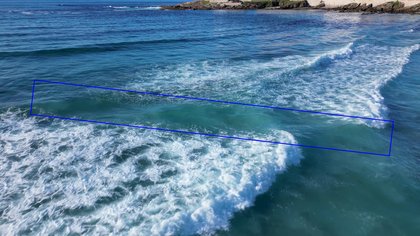}
\\[-1mm]

\variantimg{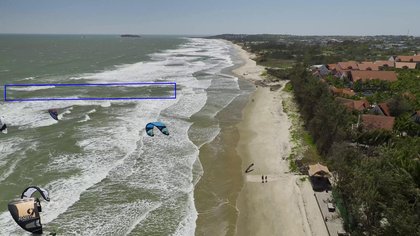} &
\variantimg{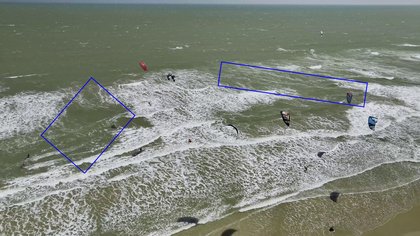} &
\variantimg{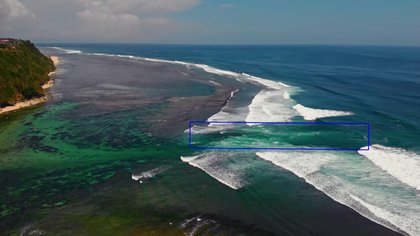} &
\variantimg{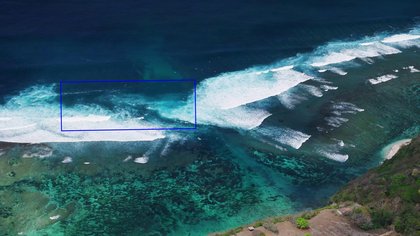}
\end{tabular}
\caption{Examples from our dataset of images with rip currents sampled from 40 videos (out of
a total of 152 videos with rips), with \textcolor{blue}{blue} oriented bounding box annotations.}
\label{fig:ripbench-obb-examples}
\end{figure*}

\begin{figure*}
\centering
\setlength{\tabcolsep}{1pt}
\begin{tabular}{c c c c}
\variantimg{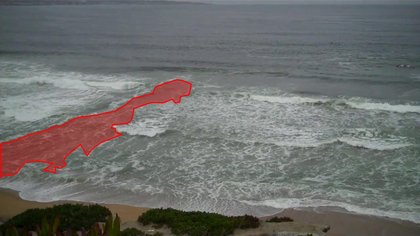} &
\variantimg{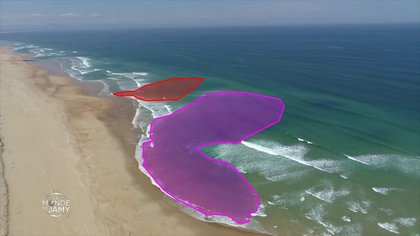} &
\variantimg{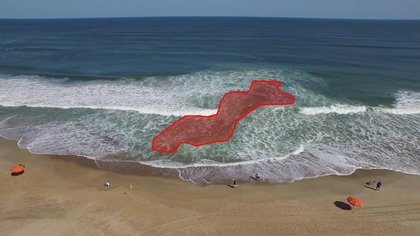} &
\variantimg{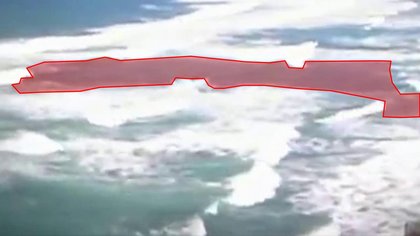}
\\[-1mm]

\variantimg{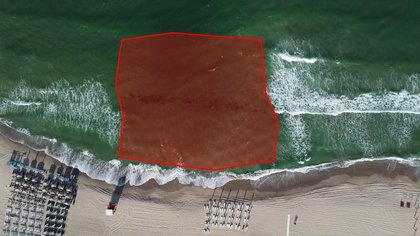} &
\variantimg{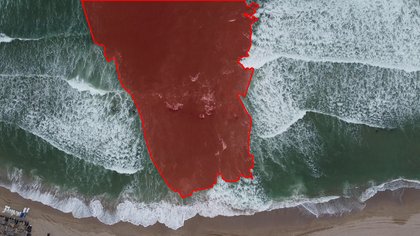} &
\variantimg{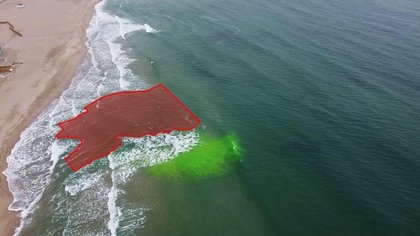} &
\variantimg{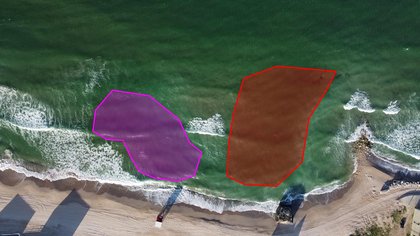}
\\[-1mm]

\variantimg{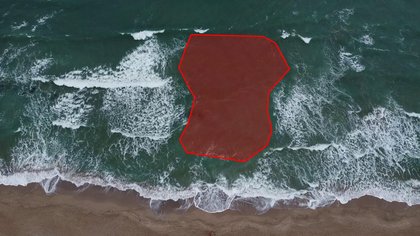} &
\variantimg{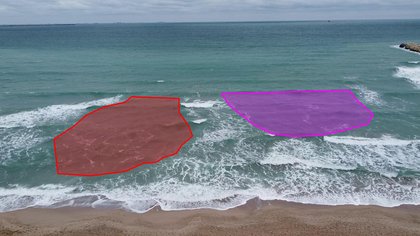} &
\variantimg{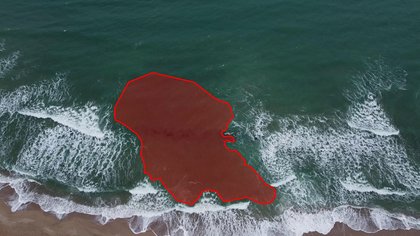} &
\variantimg{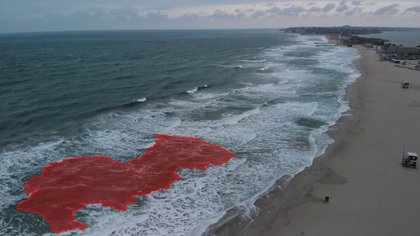}
\\[-1mm]

\variantimg{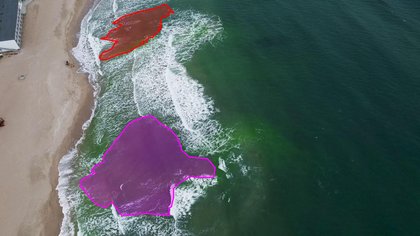} &
\variantimg{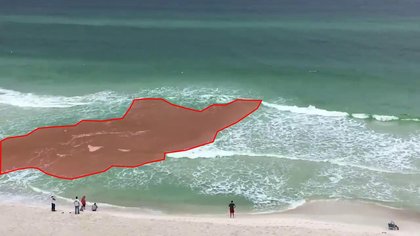} &
\variantimg{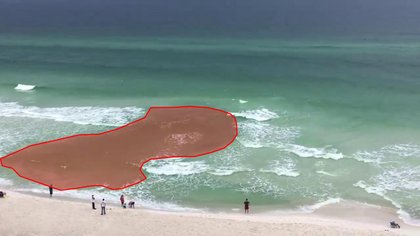} &
\variantimg{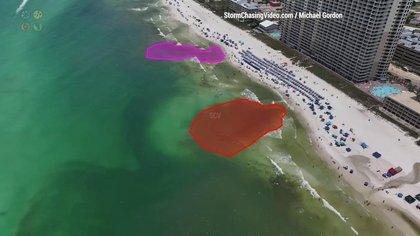}
\\[-1mm]

\variantimg{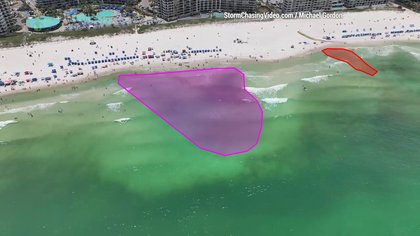} &
\variantimg{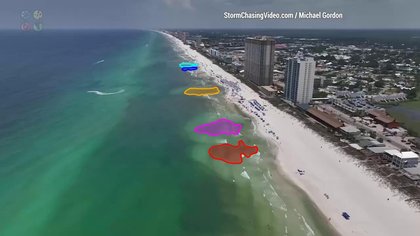} &
\variantimg{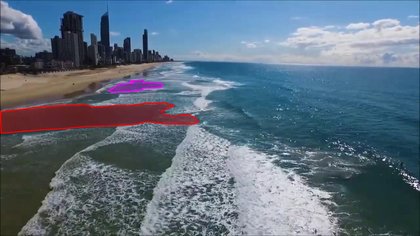} &
\variantimg{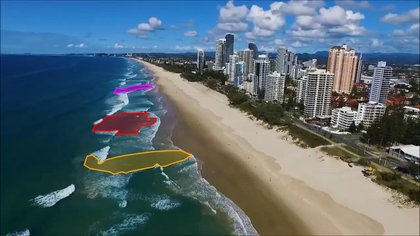}
\\[-1mm]

\variantimg{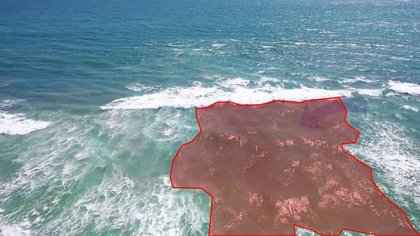} &
\variantimg{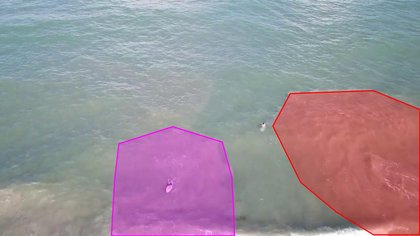} &
\variantimg{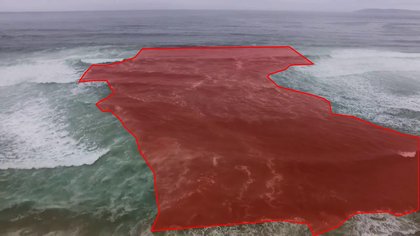} &
\variantimg{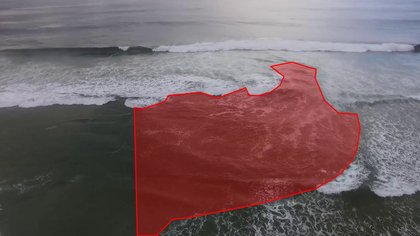}
\\[-1mm]

\variantimg{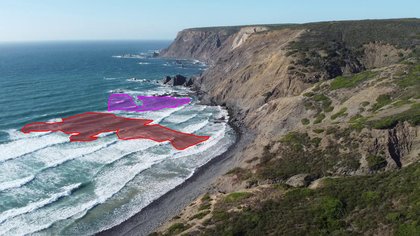} &
\variantimg{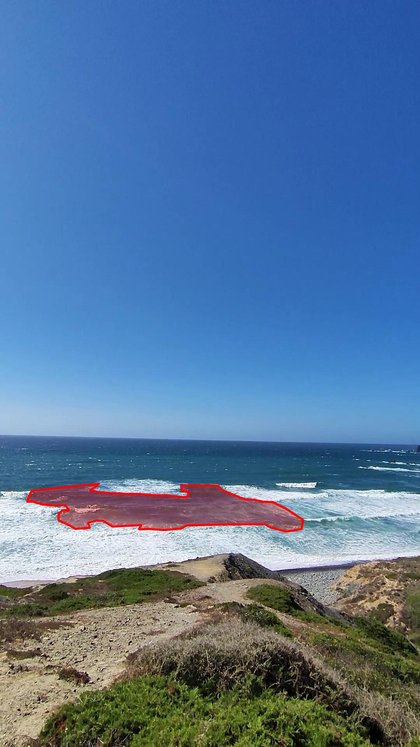} &
\variantimg{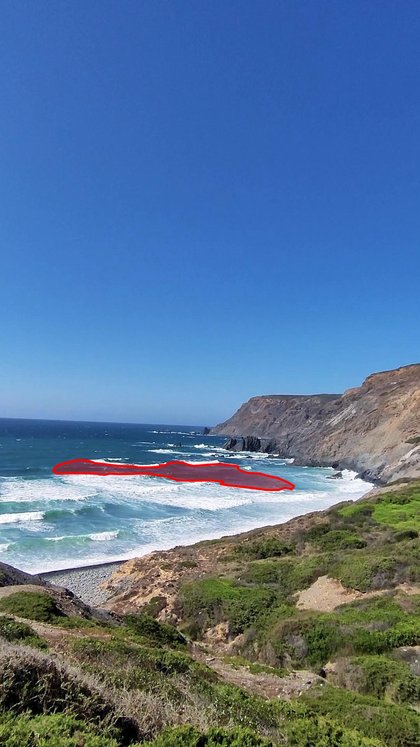} &
\variantimg{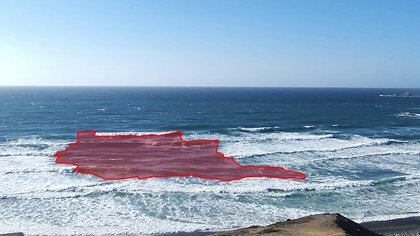}
\\[-1mm]

\variantimg{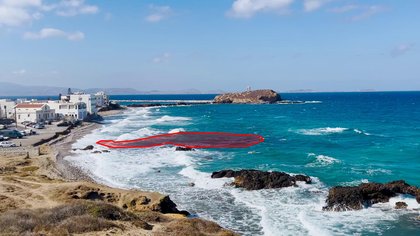} &
\variantimg{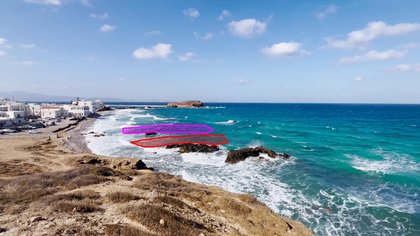} &
\variantimg{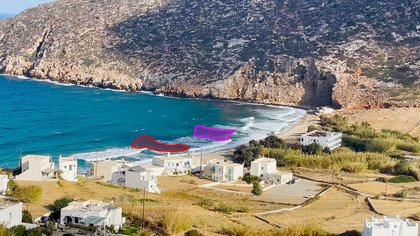} &
\variantimg{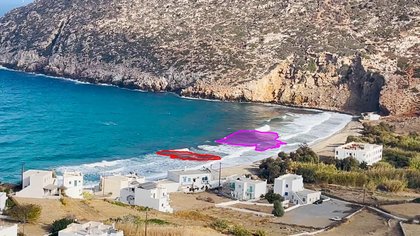}
\\[-1mm]

\variantimg{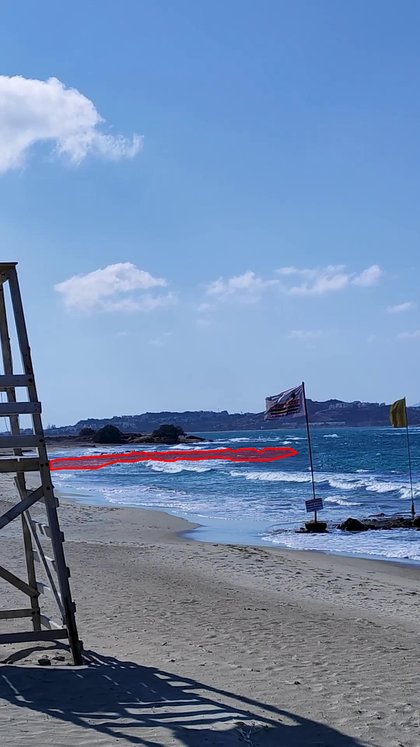} &
\variantimg{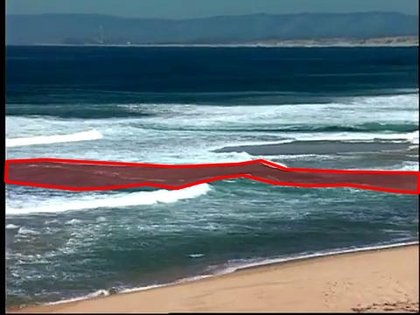} &
\variantimg{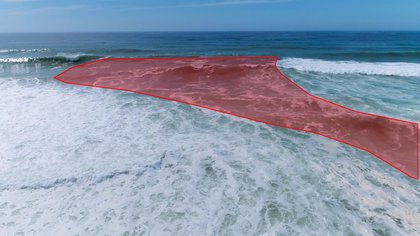} &
\variantimg{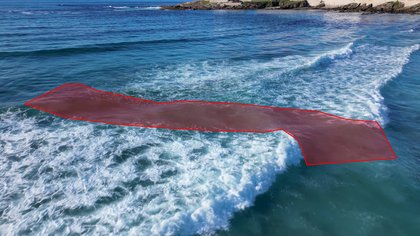}
\\[-1mm]

\variantimg{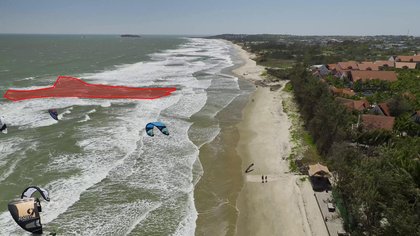} &
\variantimg{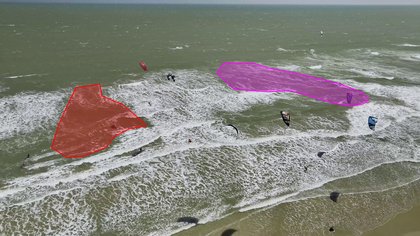} &
\variantimg{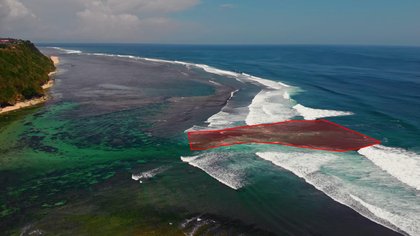} &
\variantimg{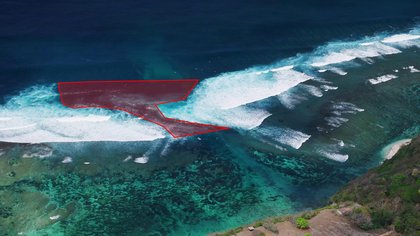}
\end{tabular}
\caption{Examples from our dataset of images with rip currents sampled from 40 videos (out of
a total of 152 videos with rips), with \textcolor{red}{red} and \textcolor{cyan}{cyan} instance segmentation annotations.}
\label{fig:ripbench-instance-seg-examples}
\end{figure*}

\begin{figure*}
\centering
\setlength{\tabcolsep}{1pt}
\begin{tabular}{c c c c}
\variantimg{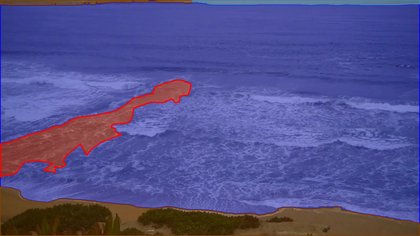} &
\variantimg{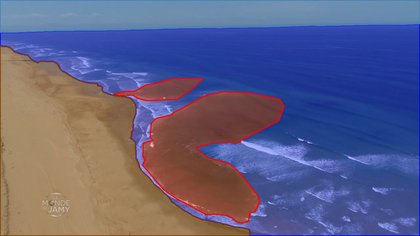} &
\variantimg{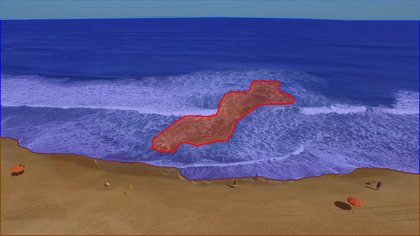} &
\variantimg{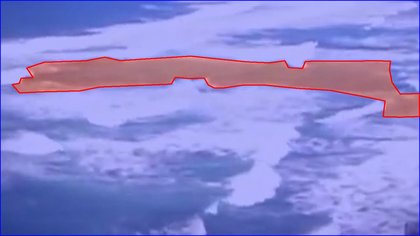}
\\[-1mm]

\variantimg{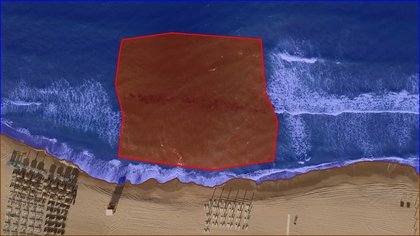} &
\variantimg{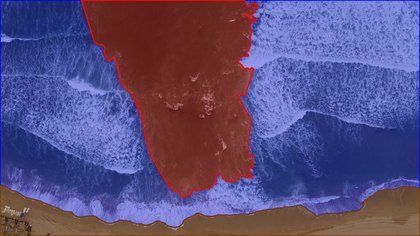} &
\variantimg{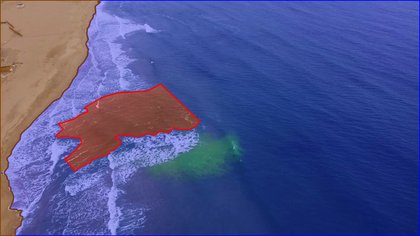} &
\variantimg{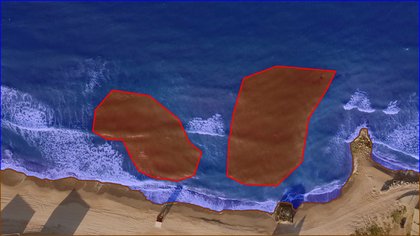}
\\[-1mm]

\variantimg{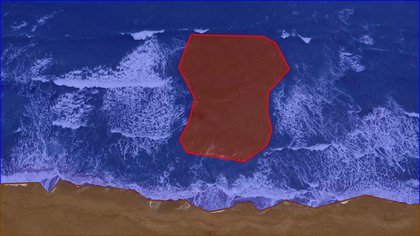} &
\variantimg{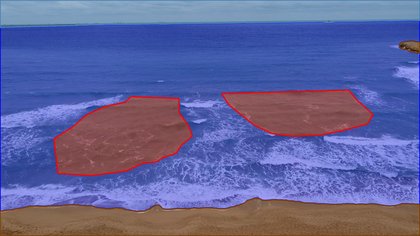} &
\variantimg{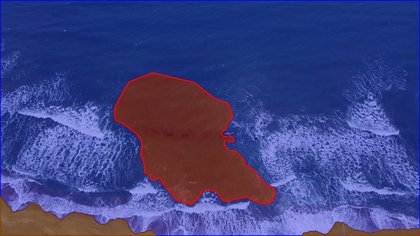} &
\variantimg{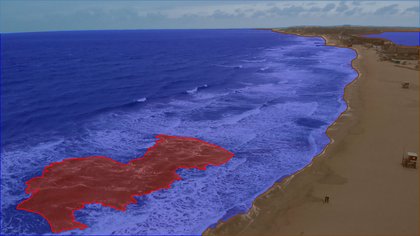}
\\[-1mm]

\variantimg{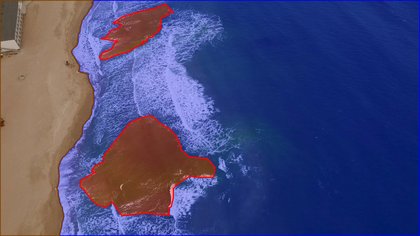} &
\variantimg{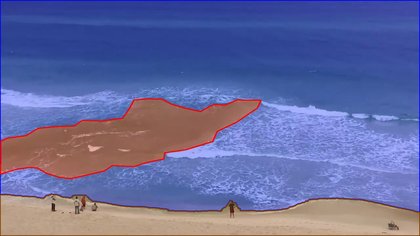} &
\variantimg{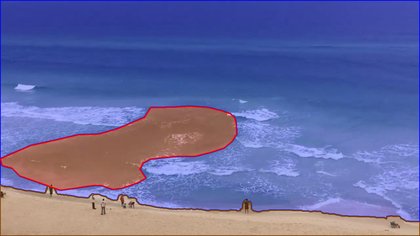} &
\variantimg{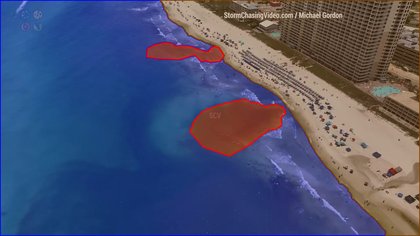}
\\[-1mm]

\variantimg{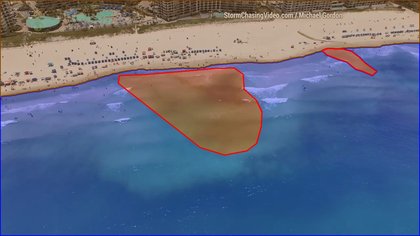} &
\variantimg{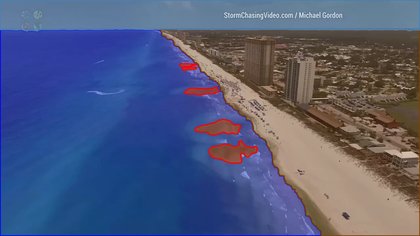} &
\variantimg{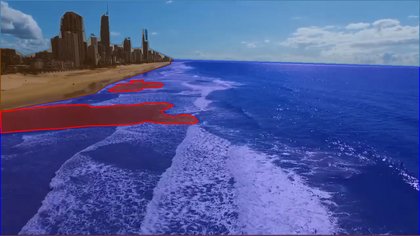} &
\variantimg{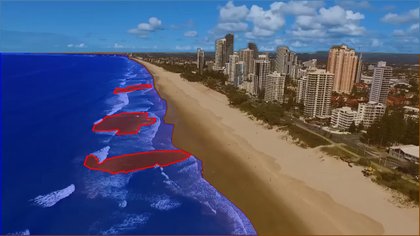}
\\[-1mm]

\variantimg{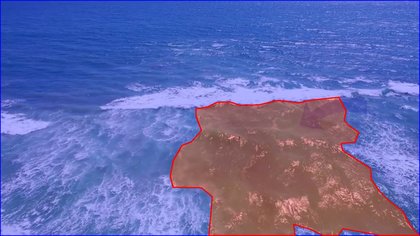} &
\variantimg{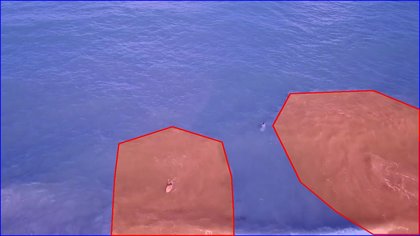} &
\variantimg{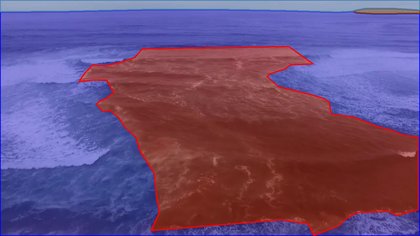} &
\variantimg{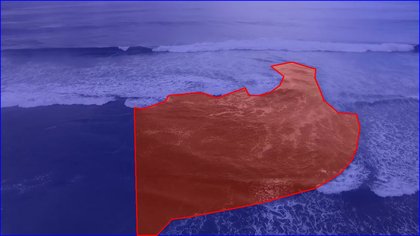}
\\[-1mm]

\variantimg{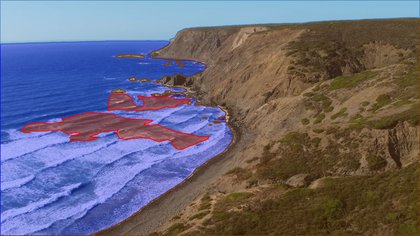} &
\variantimg{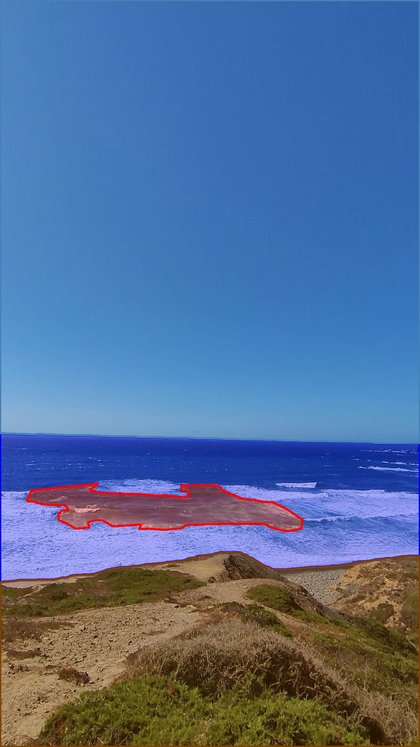} &
\variantimg{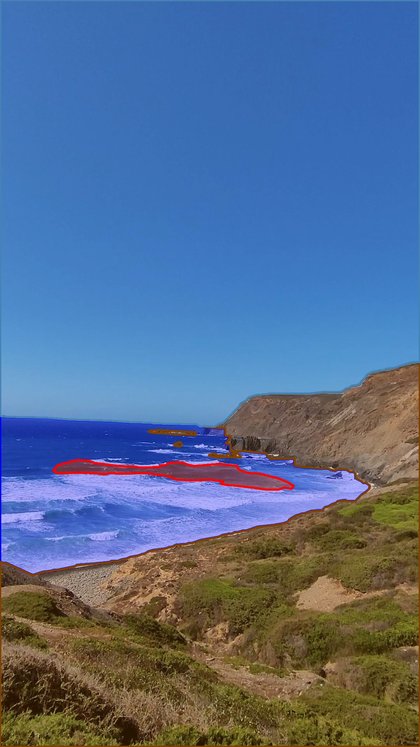} &
\variantimg{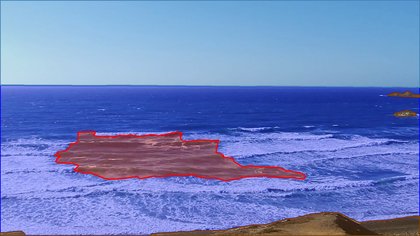}
\\[-1mm]

\variantimg{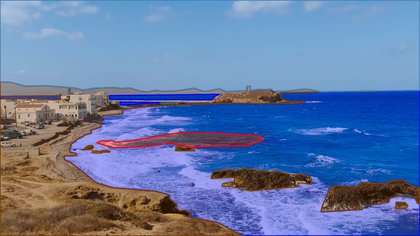} &
\variantimg{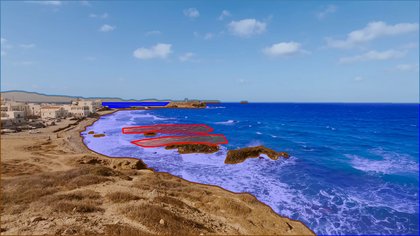} &
\variantimg{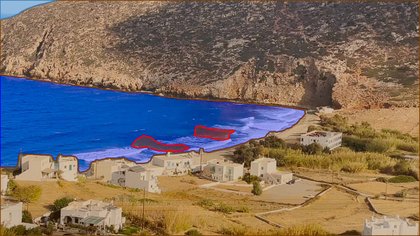} &
\variantimg{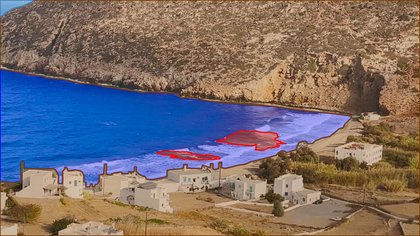}
\\[-1mm]

\variantimg{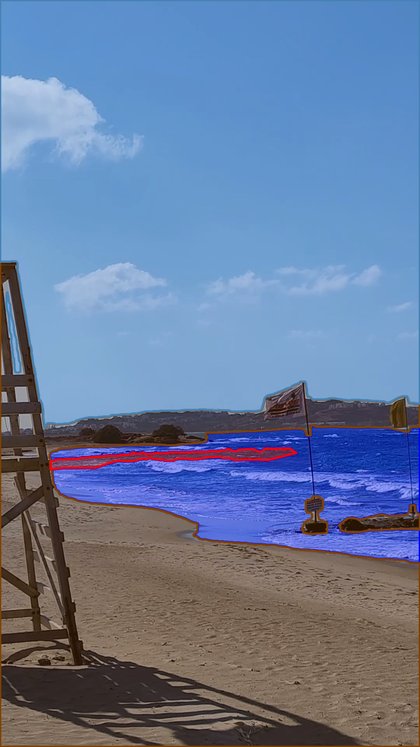} &
\variantimg{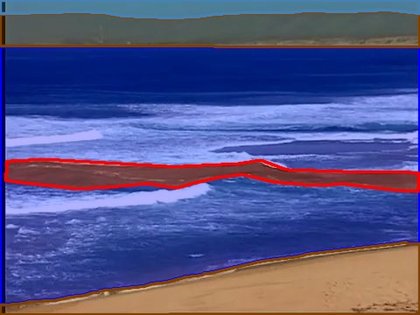} &
\variantimg{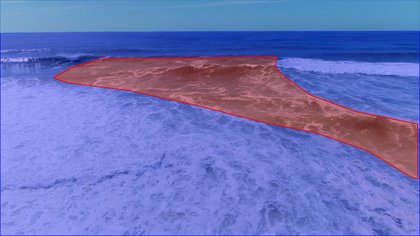} &
\variantimg{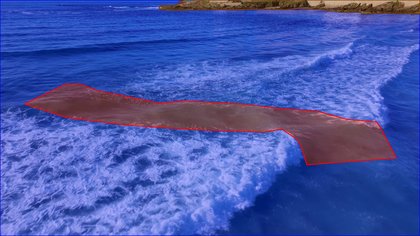}
\\[-1mm]

\variantimg{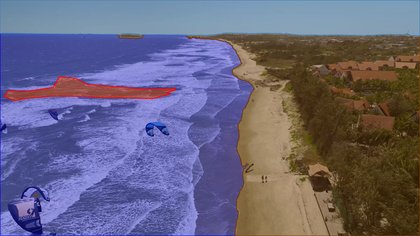} &
\variantimg{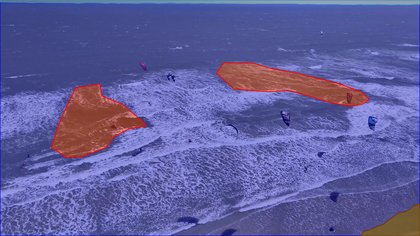} &
\variantimg{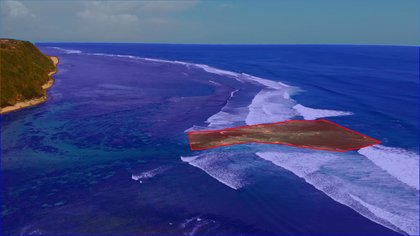} &
\variantimg{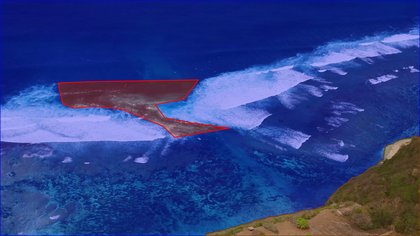}
\end{tabular}
\caption{Examples from our dataset of images with rip currents sampled from 40 videos (out of
a total of 152 videos with rips), with semantic segmentation annotations: \textcolor{Aquamarine}{aquamarine} for sky, \textcolor{Tan}{tan} for land, \textcolor{RoyalBlue}{royal blue} for water and \textcolor{red}{red} masks for rip currents.}
\label{fig:ripbench-semantic-seg-examples}
\end{figure*}

\begin{figure*}
\centering
\setlength{\tabcolsep}{1pt}
\begin{tabular}{c c c c}
\variantimg{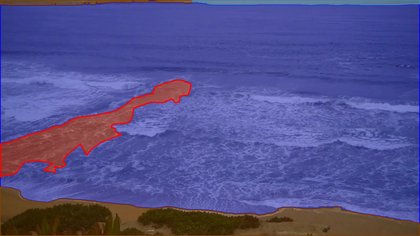} &
\variantimg{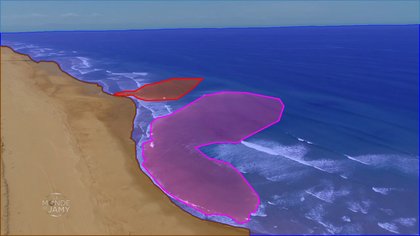} &
\variantimg{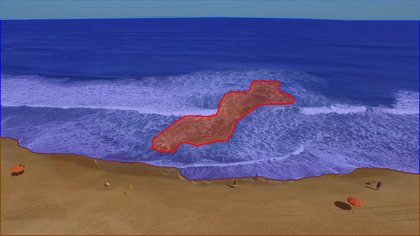} &
\variantimg{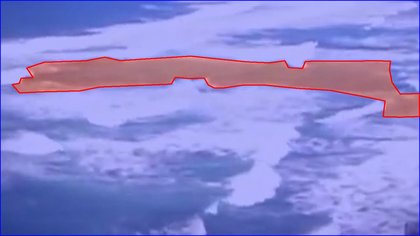}
\\[-1mm]

\variantimg{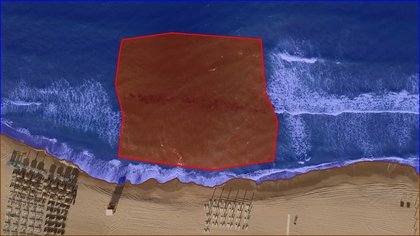} &
\variantimg{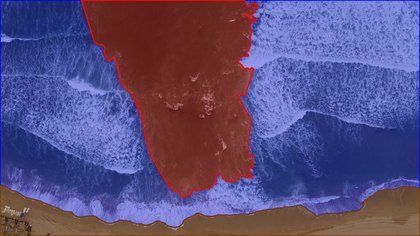} &
\variantimg{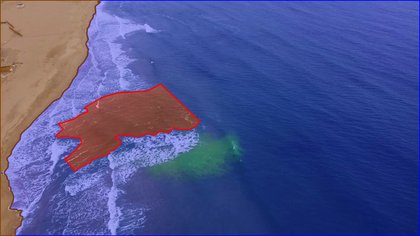} &
\variantimg{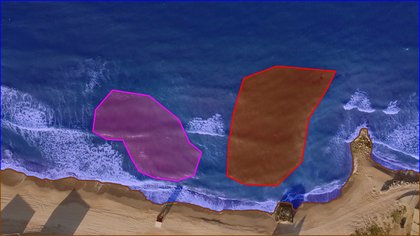}
\\[-1mm]

\variantimg{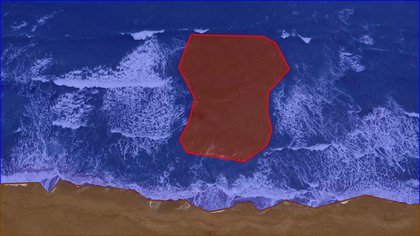} &
\variantimg{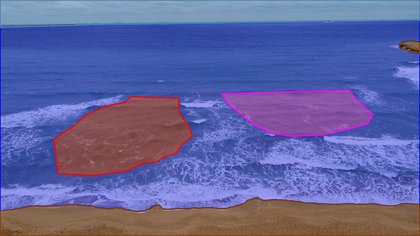} &
\variantimg{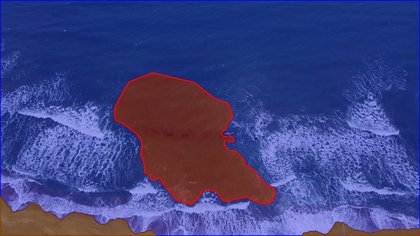} &
\variantimg{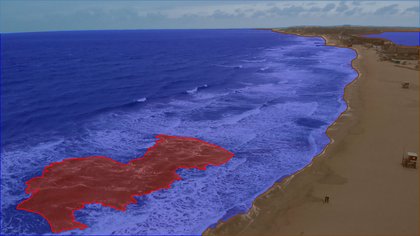}
\\[-1mm]

\variantimg{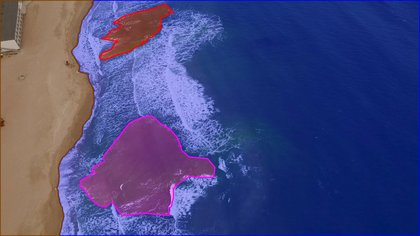} &
\variantimg{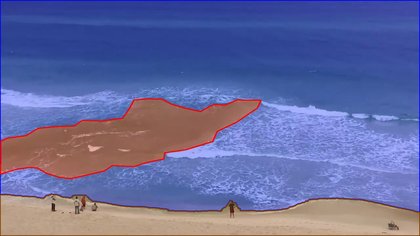} &
\variantimg{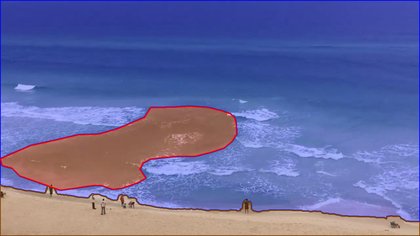} &
\variantimg{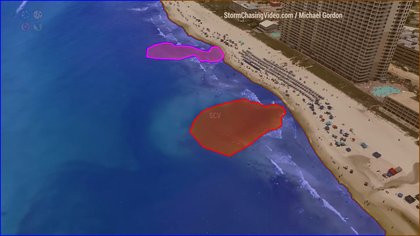}
\\[-1mm]

\variantimg{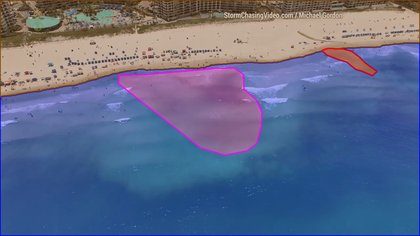} &
\variantimg{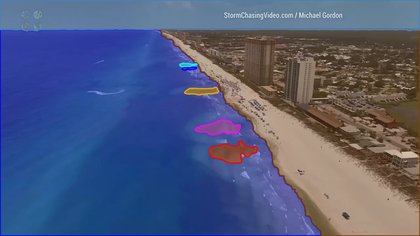} &
\variantimg{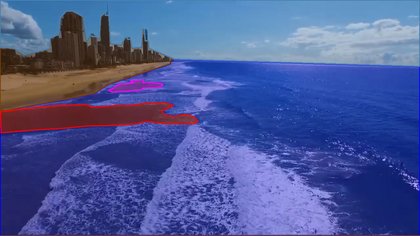} &
\variantimg{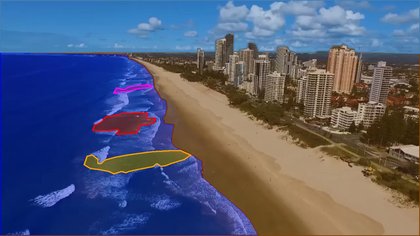}
\\[-1mm]

\variantimg{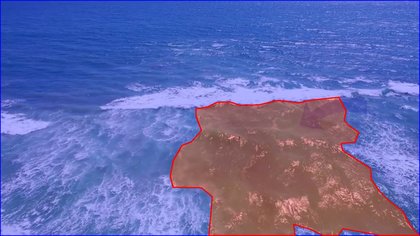} &
\variantimg{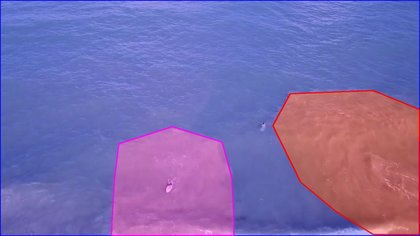} &
\variantimg{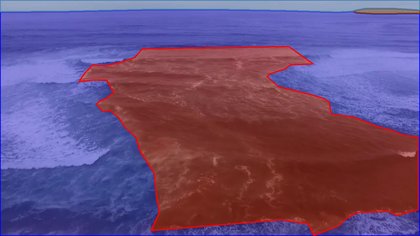} &
\variantimg{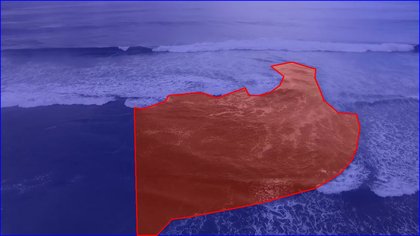}
\\[-1mm]

\variantimg{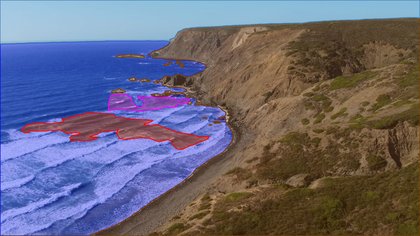} &
\variantimg{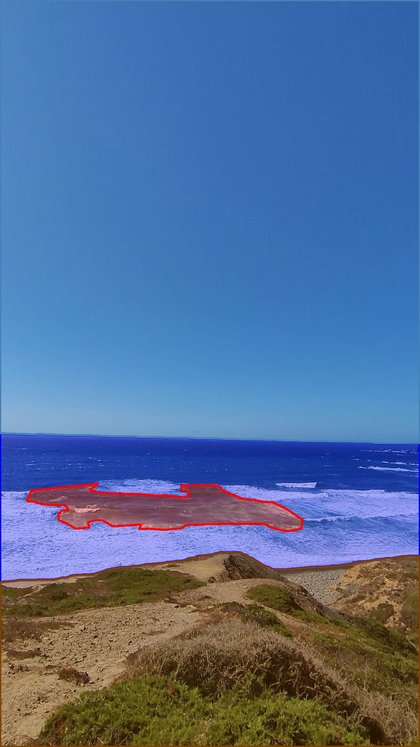} &
\variantimg{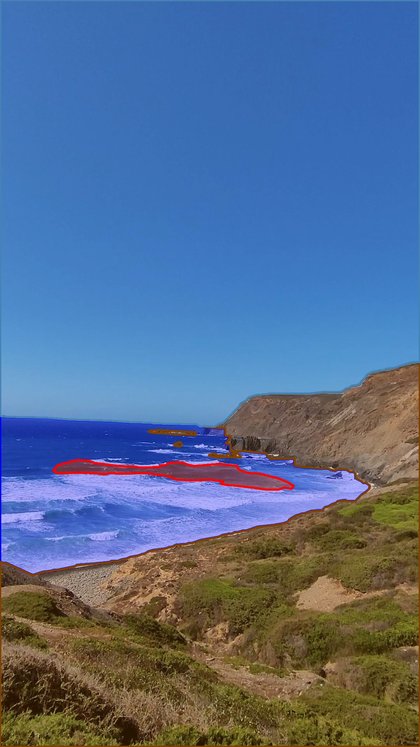} &
\variantimg{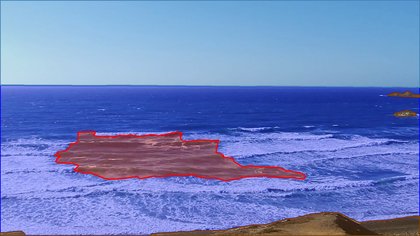}
\\[-1mm]

\variantimg{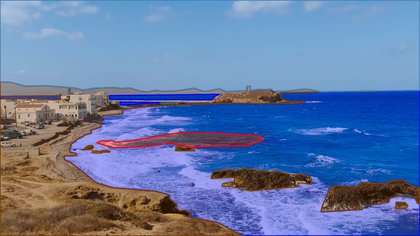} &
\variantimg{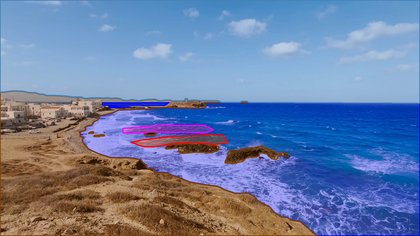} &
\variantimg{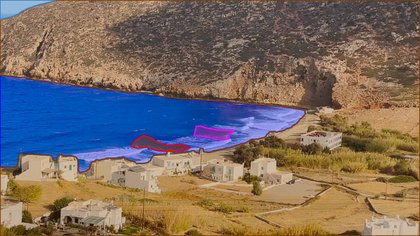} &
\variantimg{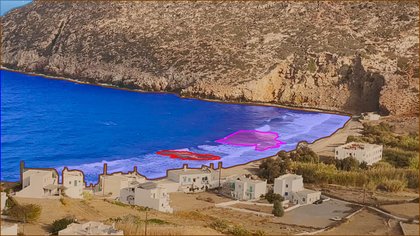}
\\[-1mm]

\variantimg{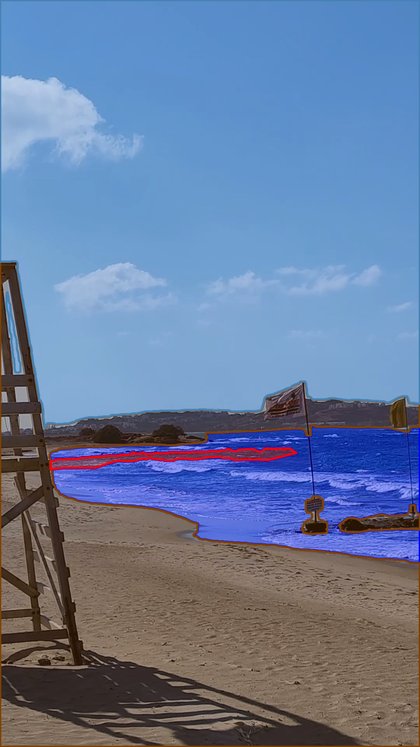} &
\variantimg{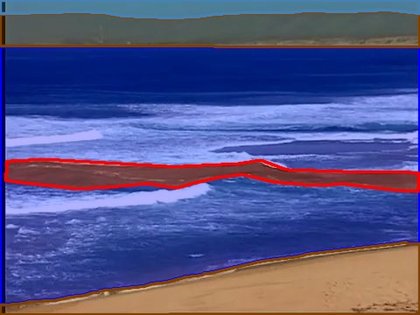} &
\variantimg{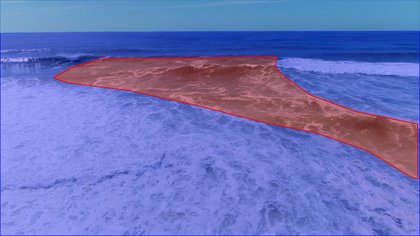} &
\variantimg{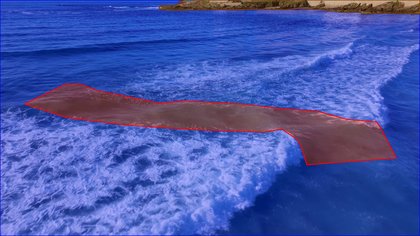}
\\[-1mm]

\variantimg{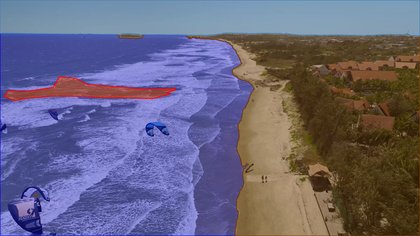} &
\variantimg{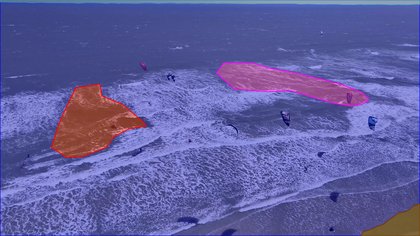} &
\variantimg{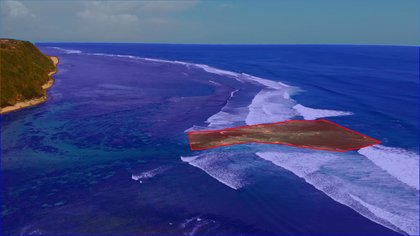} &
\variantimg{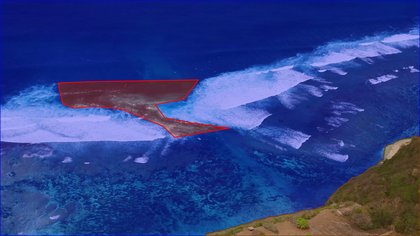}
\end{tabular}
\caption{Examples from our dataset of images with rip currents sampled from 40 videos (out of
a total of 152 videos with rips), with panoptic segmentation annotations: \textcolor{Aquamarine}{aquamarine} for sky, \textcolor{Tan}{tan} for land, \textcolor{RoyalBlue}{royal blue} for water and \textcolor{red}{red} and \textcolor{cyan}{cyan} masks for rip currents.}
\label{fig:ripbench-panoptic-seg-examples}
\end{figure*}

\begin{figure*}
\centering
\setlength{\tabcolsep}{1pt}
\begin{tabular}{c c c c}
\variantimg{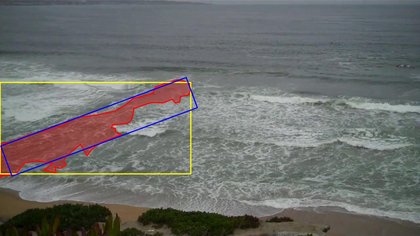} &
\variantimg{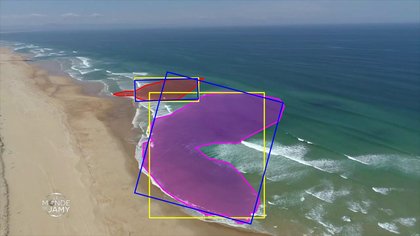} &
\variantimg{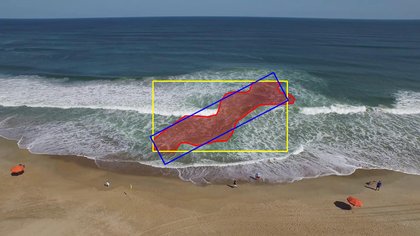} &
\variantimg{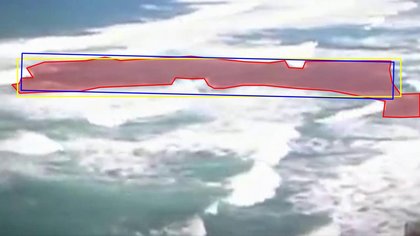}
\\[-1mm]

\variantimg{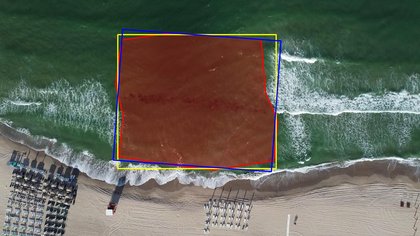} &
\variantimg{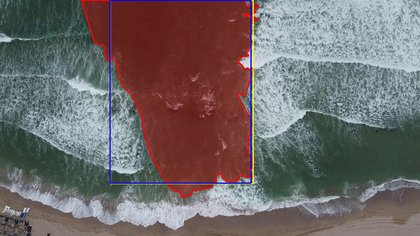} &
\variantimg{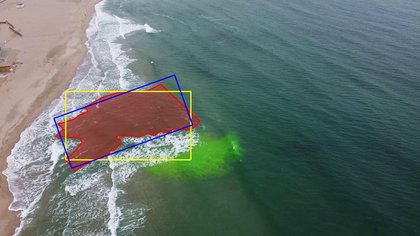} &
\variantimg{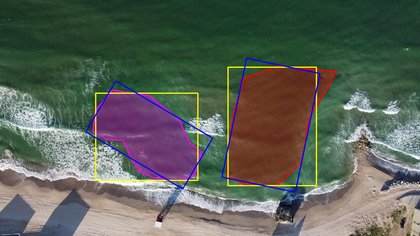}
\\[-1mm]

\variantimg{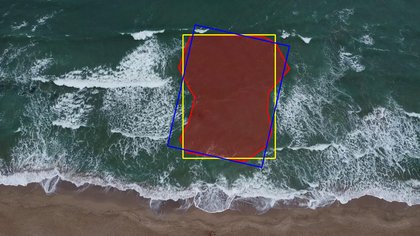} &
\variantimg{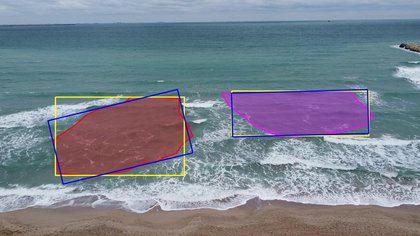} &
\variantimg{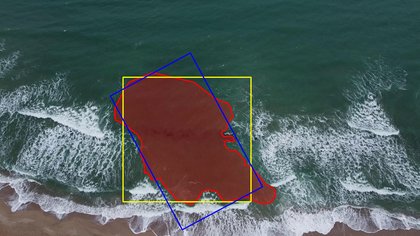} &
\variantimg{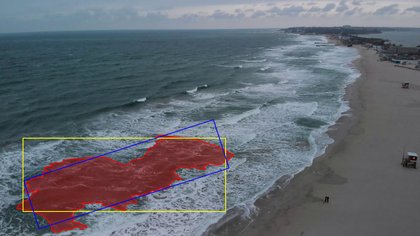}
\\[-1mm]

\variantimg{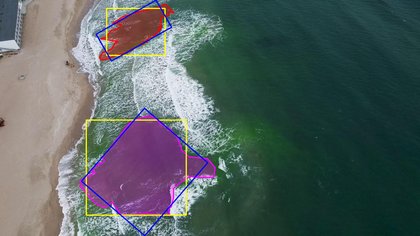} &
\variantimg{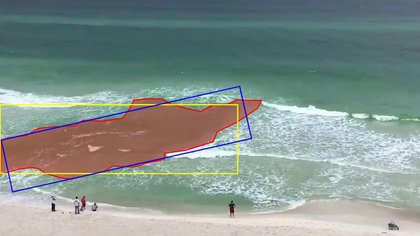} &
\variantimg{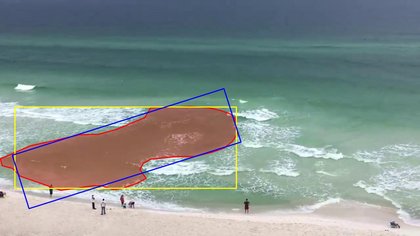} &
\variantimg{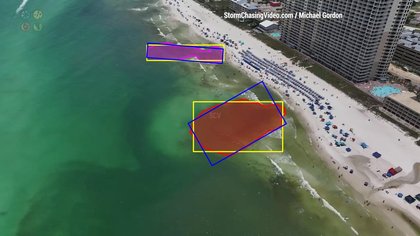}
\\[-1mm]

\variantimg{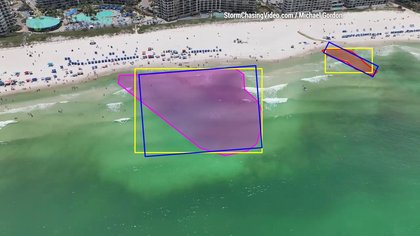} &
\variantimg{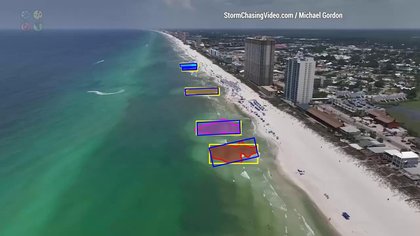} &
\variantimg{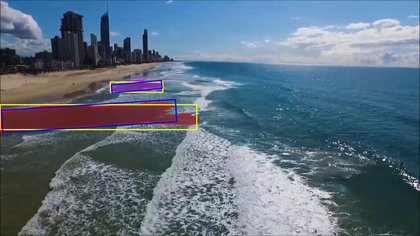} &
\variantimg{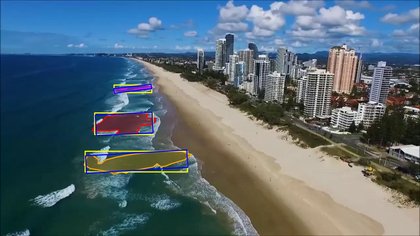}
\\[-1mm]

\variantimg{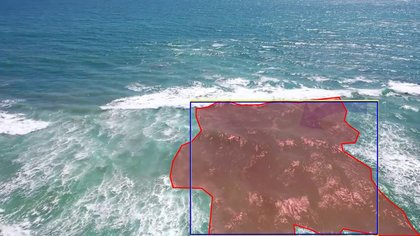} &
\variantimg{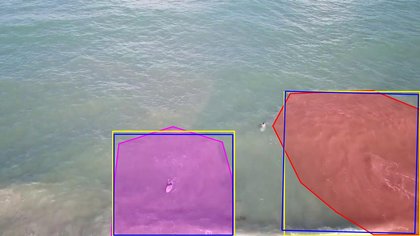} &
\variantimg{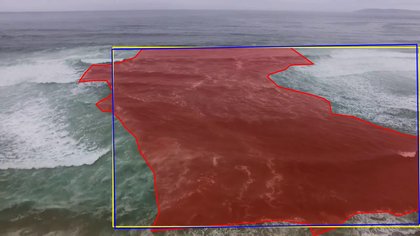} &
\variantimg{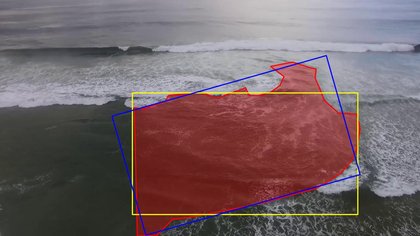}
\\[-1mm]

\variantimg{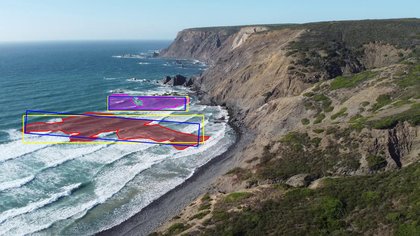} &
\variantimg{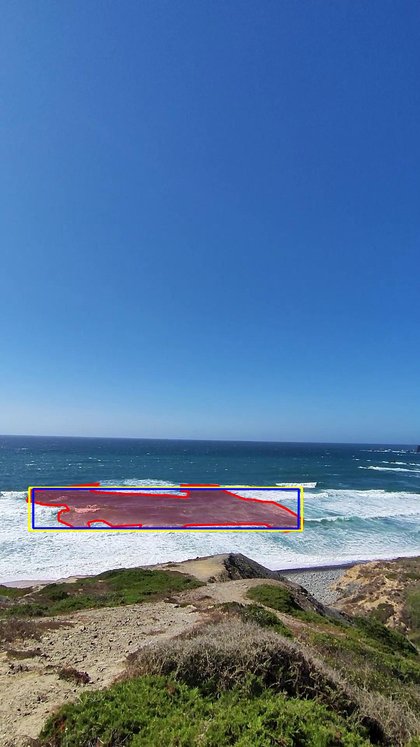} &
\variantimg{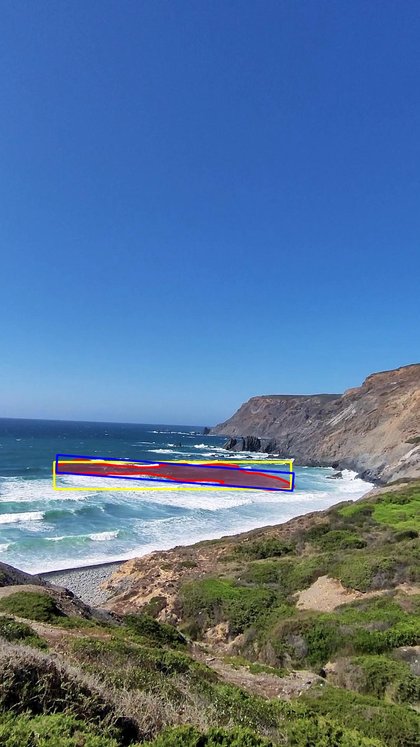} &
\variantimg{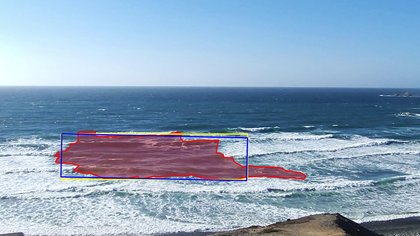}
\\[-1mm]

\variantimg{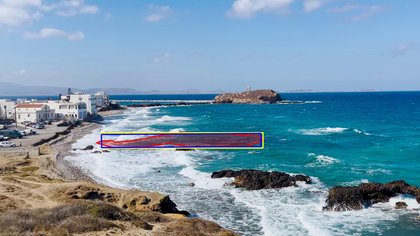} &
\variantimg{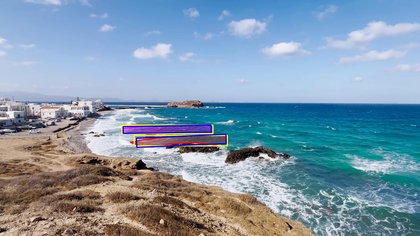} &
\variantimg{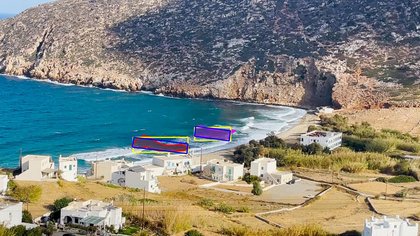} &
\variantimg{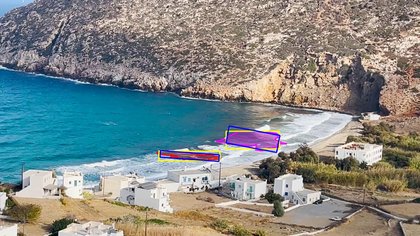}
\\[-1mm]

\variantimg{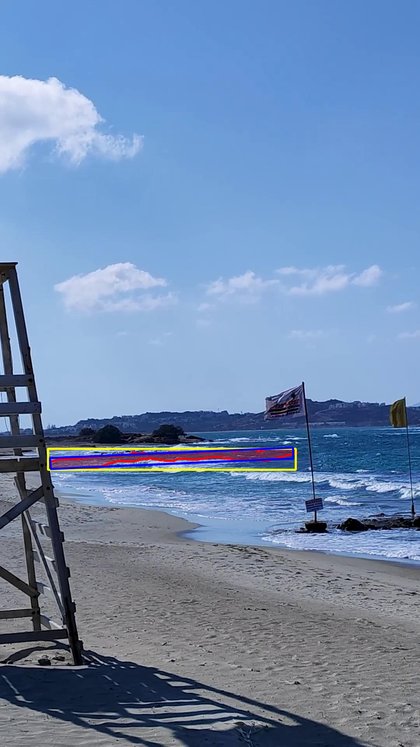} &
\variantimg{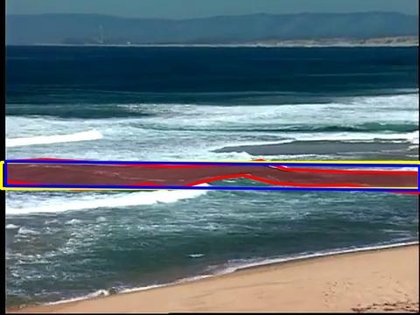} &
\variantimg{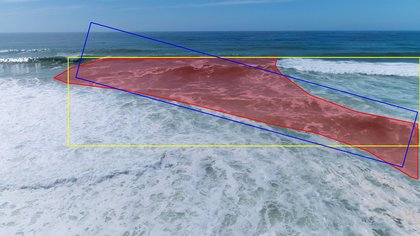} &
\variantimg{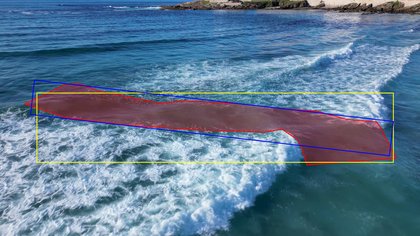}
\\[-1mm]

\variantimg{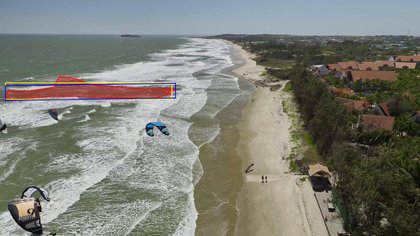} &
\variantimg{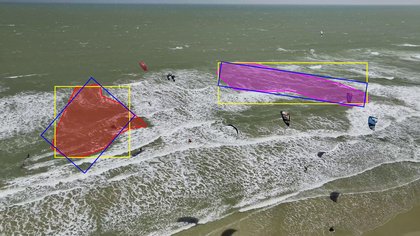} &
\variantimg{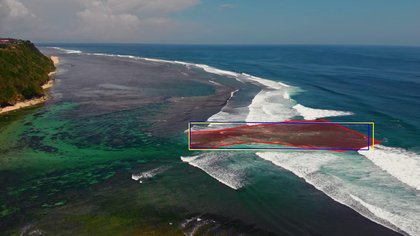} &
\variantimg{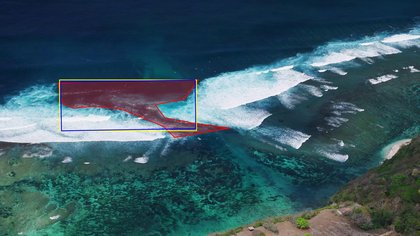}
\end{tabular}
\caption{Examples from our dataset of images with rip currents sampled from 40 videos (out of
a total of 152 videos with rips), with rip currents annotations combined. \textcolor{yellow}{Yellow} for axis-aligned bounding boxes, \textcolor{blue}{blue} for oriented bounding boxes and \textcolor{red}{red} and \textcolor{cyan}{cyan} masks for rip currents.}
\label{fig:ripbench-all-combined-examples}
\end{figure*}

\section{People in the Footage}
\label{sec:people_audit}

RipBench shows public beaches, so people appear in the footage. We did not blur or remove them: the frames are the raw evidence a rip current model has to handle, and blurring would alter the water regions that are annotated in crowded scenes. To document the exposure instead, we ran a person detector (YOLO26-n, COCO weights, inference at $1{,}920$~px, confidence $0.15$, deliberately permissive) over all $51{,}011$ sampled frames of the $304$ videos. A person is detected in $216$ of the $304$ videos and in $19{,}331$ frames ($37.9\%$); in the large majority of these frames the people are distant figures a few pixels tall. Only $70$ videos contain a detected person at least $120$~px tall in at least one frame ($1{,}475$ frames), and $20$ videos contain one at least $300$~px tall ($303$ frames), the range at which a face could plausibly be recognisable. No names or identities of the people shown are recorded anywhere in the dataset, and the release carries no audio. The per-video and per-frame audit (\texttt{provenance/people\_audit\_videos.tsv}, \texttt{provenance/people\_audit\_frames.tsv}, including the detector boxes) ships with the dataset, so that users who need a stricter policy can filter or blur the affected frames themselves.

\newpage

\clearpage

\section*{NeurIPS Paper Checklist}

\begin{enumerate}

\item {\bf Claims}
    \item[] Question: Do the main claims made in the abstract and introduction accurately reflect the paper's contributions and scope?
    \item[] Answer: \answerYes{}
    \item[] Justification: The abstract and introduction summarize the contributions of RipBench, namely a unified multi-task benchmark with consistent annotations across classification, detection, and segmentation, baseline evaluations, and an analysis of task-dependent performance. Claims about saturation in classification and the relative difficulty of localization and segmentation are supported by the experimental results.

    \item[] Guidelines:
    \begin{itemize}
        \item The answer \answerNA{} means that the abstract and introduction do not include the claims made in the paper.
        \item The abstract and/or introduction should clearly state the claims made, including the contributions made in the paper and important assumptions and limitations. A \answerNo{} or \answerNA{} answer to this question will not be perceived well by the reviewers. 
        \item The claims made should match theoretical and experimental results, and reflect how much the results can be expected to generalize to other settings. 
        \item It is fine to include aspirational goals as motivation as long as it is clear that these goals are not attained by the paper. 
    \end{itemize}

\item {\bf Limitations}
    \item[] Question: Does the paper discuss the limitations of the work performed by the authors?
        \item[] Answer: \answerYes{}
    \item[] Justification: Limitations are discussed in Section~\ref{sec:discussion_and_limitations}, covering safety-critical deployment caveats, dataset coverage gaps, annotation ambiguity for amorphous boundaries, the confound between task difficulty and annotation targets in cross-task comparisons, and the gap between benchmark performance and operational deployment readiness.

    \item[] Guidelines:
    \begin{itemize}
        \item The answer \answerNA{} means that the paper has no limitation while the answer \answerNo{} means that the paper has limitations, but those are not discussed in the paper. 
        \item The authors are encouraged to create a separate ``Limitations'' section in their paper.
        \item The paper should point out any strong assumptions and how robust the results are to violations of these assumptions (e.g., independence assumptions, noiseless settings, model well-specification, asymptotic approximations only holding locally). The authors should reflect on how these assumptions might be violated in practice and what the implications would be.
        \item The authors should reflect on the scope of the claims made, e.g., if the approach was only tested on a few datasets or with a few runs. In general, empirical results often depend on implicit assumptions, which should be articulated.
        \item The authors should reflect on the factors that influence the performance of the approach. For example, a facial recognition algorithm may perform poorly when image resolution is low or images are taken in low lighting. Or a speech-to-text system might not be used reliably to provide closed captions for online lectures because it fails to handle technical jargon.
        \item The authors should discuss the computational efficiency of the proposed algorithms and how they scale with dataset size.
        \item If applicable, the authors should discuss possible limitations of their approach to address problems of privacy and fairness.
        \item While the authors might fear that complete honesty about limitations might be used by reviewers as grounds for rejection, a worse outcome might be that reviewers discover limitations that aren't acknowledged in the paper. The authors should use their best judgment and recognize that individual actions in favor of transparency play an important role in developing norms that preserve the integrity of the community. Reviewers will be specifically instructed to not penalize honesty concerning limitations.
    \end{itemize}

\item {\bf Theory assumptions and proofs}
    \item[] Question: For each theoretical result, does the paper provide the full set of assumptions and a complete (and correct) proof?
    \item[] Answer: \answerNA{}
    \item[] Justification: The paper introduces a benchmark dataset and presents empirical results; it does not include theoretical results or formal proofs.
    \item[] Guidelines:
    \begin{itemize}
        \item The answer \answerNA{} means that the paper does not include theoretical results. 
        \item All the theorems, formulas, and proofs in the paper should be numbered and cross-referenced.
        \item All assumptions should be clearly stated or referenced in the statement of any theorems.
        \item The proofs can either appear in the main paper or the supplemental material, but if they appear in the supplemental material, the authors are encouraged to provide a short proof sketch to provide intuition. 
        \item Inversely, any informal proof provided in the core of the paper should be complemented by formal proofs provided in appendix or supplemental material.
        \item Theorems and Lemmas that the proof relies upon should be properly referenced. 
    \end{itemize}

    \item {\bf Experimental result reproducibility}
    \item[] Question: Does the paper fully disclose all the information needed to reproduce the main experimental results of the paper to the extent that it affects the main claims and/or conclusions of the paper (regardless of whether the code and data are provided or not)?
    \item[] Answer: \answerYes{}
    \item[] Justification: Experimental setup is described in Section~\ref{sec:experimental_setup}, with full per-task training configurations, hyperparameters, optimizers, augmentation strategies, and batch sizes provided in Supplementary Section~\ref{sec:hyperparameter_tuning}. Standardized train/validation/test splits, metric definitions, and evaluation protocol are documented. The dataset, annotations, and baseline code are released through the project link.

    \item[] Guidelines:
    \begin{itemize}
        \item The answer \answerNA{} means that the paper does not include experiments.
        \item If the paper includes experiments, a \answerNo{} answer to this question will not be perceived well by the reviewers: Making the paper reproducible is important, regardless of whether the code and data are provided or not.
        \item If the contribution is a dataset and\slash or model, the authors should describe the steps taken to make their results reproducible or verifiable. 
        \item Depending on the contribution, reproducibility can be accomplished in various ways. For example, if the contribution is a novel architecture, describing the architecture fully might suffice, or if the contribution is a specific model and empirical evaluation, it may be necessary to either make it possible for others to replicate the model with the same dataset, or provide access to the model. In general. releasing code and data is often one good way to accomplish this, but reproducibility can also be provided via detailed instructions for how to replicate the results, access to a hosted model (e.g., in the case of a large language model), releasing of a model checkpoint, or other means that are appropriate to the research performed.
        \item While NeurIPS does not require releasing code, the conference does require all submissions to provide some reasonable avenue for reproducibility, which may depend on the nature of the contribution. For example
        \begin{enumerate}
            \item If the contribution is primarily a new algorithm, the paper should make it clear how to reproduce that algorithm.
            \item If the contribution is primarily a new model architecture, the paper should describe the architecture clearly and fully.
            \item If the contribution is a new model (e.g., a large language model), then there should either be a way to access this model for reproducing the results or a way to reproduce the model (e.g., with an open-source dataset or instructions for how to construct the dataset).
            \item We recognize that reproducibility may be tricky in some cases, in which case authors are welcome to describe the particular way they provide for reproducibility. In the case of closed-source models, it may be that access to the model is limited in some way (e.g., to registered users), but it should be possible for other researchers to have some path to reproducing or verifying the results.
        \end{enumerate}
    \end{itemize}

\item {\bf Open access to data and code}
    \item[] Question: Does the paper provide open access to the data and code, with sufficient instructions to faithfully reproduce the main experimental results, as described in supplemental material?
    \item[] Answer: \answerYes{}
    \item[] Justification: The dataset, annotations, baseline code, and standardized splits are made available via the links in the croissant file and in the OpenReview form. The dataset is released under CC-BY-NC 4.0 due to licensing constraints on a subset of the source videos. 

    \item[] Guidelines:
    \begin{itemize}
        \item The answer \answerNA{} means that paper does not include experiments requiring code.
        \item Please see the NeurIPS code and data submission guidelines (\url{https://neurips.cc/public/guides/CodeSubmissionPolicy}) for more details.
        \item While we encourage the release of code and data, we understand that this might not be possible, so \answerNo{} is an acceptable answer. Papers cannot be rejected simply for not including code, unless this is central to the contribution (e.g., for a new open-source benchmark).
        \item The instructions should contain the exact command and environment needed to run to reproduce the results. See the NeurIPS code and data submission guidelines (\url{https://neurips.cc/public/guides/CodeSubmissionPolicy}) for more details.
        \item The authors should provide instructions on data access and preparation, including how to access the raw data, preprocessed data, intermediate data, and generated data, etc.
        \item The authors should provide scripts to reproduce all experimental results for the new proposed method and baselines. If only a subset of experiments are reproducible, they should state which ones are omitted from the script and why.
        \item At submission time, to preserve anonymity, the authors should release anonymized versions (if applicable).
        \item Providing as much information as possible in supplemental material (appended to the paper) is recommended, but including URLs to data and code is permitted.
    \end{itemize}

\item {\bf Experimental setting/details}
    \item[] Question: Does the paper specify all the training and test details (e.g., data splits, hyperparameters, how they were chosen, type of optimizer) necessary to understand the results?
    \item[] Answer: \answerYes{}
    \item[] Justification: The video-level train/validation/test split is fixed across all task formulations. Per-task hyperparameters, optimizers, learning rate schedules, batch sizes, and augmentation strategies are detailed in Supplementary Section~\ref{sec:hyperparameter_tuning}.

    \item[] Guidelines:
    \begin{itemize}
        \item The answer \answerNA{} means that the paper does not include experiments.
        \item The experimental setting should be presented in the core of the paper to a level of detail that is necessary to appreciate the results and make sense of them.
        \item The full details can be provided either with the code, in appendix, or as supplemental material.
    \end{itemize}

\item {\bf Experiment statistical significance}
    \item[] Question: Does the paper report error bars suitably and correctly defined or other appropriate information about the statistical significance of the experiments?
    \item[] Answer: \answerYes{}
    \item[] Justification: Every headline table reports $95\%$ confidence intervals obtained by a non-parametric bootstrap over the $71$ test videos ($B = 1{,}000$ resamples with replacement, fixed seed), i.e., the intervals capture sensitivity to test-set composition at the video level; frames are not resampled because frames within a video are strongly correlated and the videos are the effective independent units. Paired comparisons (two models or two tasks scored on the same resampled videos) are used for all claims that compare configurations, and we state explicitly which differences are not resolved at this sample size. Training randomness is quantified separately by multi-seed reruns of the headline configuration per task under identical conditions (Appendix~\ref{sec:multiseed}); we report the seed spread rather than assuming Normality. The bootstrap script ships with the evaluation tooling.

    \item[] Guidelines:
    \begin{itemize}
        \item The answer \answerNA{} means that the paper does not include experiments.
        \item The authors should answer \answerYes{} if the results are accompanied by error bars, confidence intervals, or statistical significance tests, at least for the experiments that support the main claims of the paper.
        \item The factors of variability that the error bars are capturing should be clearly stated (for example, train/test split, initialization, random drawing of some parameter, or overall run with given experimental conditions).
        \item The method for calculating the error bars should be explained (closed form formula, call to a library function, bootstrap, etc.)
        \item The assumptions made should be given (e.g., Normally distributed errors).
        \item It should be clear whether the error bar is the standard deviation or the standard error of the mean.
        \item It is OK to report 1-sigma error bars, but one should state it. The authors should preferably report a 2-sigma error bar than state that they have a 96\% CI, if the hypothesis of Normality of errors is not verified.
        \item For asymmetric distributions, the authors should be careful not to show in tables or figures symmetric error bars that would yield results that are out of range (e.g., negative error rates).
        \item If error bars are reported in tables or plots, the authors should explain in the text how they were calculated and reference the corresponding figures or tables in the text.
    \end{itemize}

\item {\bf Experiments compute resources}
    \item[] Question: For each experiment, does the paper provide sufficient information on the computer resources (type of compute workers, memory, time of execution) needed to reproduce the experiments?
    \item[] Answer: \answerYes{}
    \item[] Justification: Compute resources are described in Supplementary Section~\ref{sec:hyperparameter_tuning}. Training of the reported models was performed on single NVIDIA L40 GPUs of a SLURM cluster (24-hour segments, resumed), inference speed on an RTX 4090; the submitted version's retired rows used NVIDIA L40S, and RTX 4090 GPUs depending on the model and batch size requirements.

    \item[] Guidelines:
    \begin{itemize}
        \item The answer \answerNA{} means that the paper does not include experiments.
        \item The paper should indicate the type of compute workers CPU or GPU, internal cluster, or cloud provider, including relevant memory and storage.
        \item The paper should provide the amount of compute required for each of the individual experimental runs as well as estimate the total compute. 
        \item The paper should disclose whether the full research project required more compute than the experiments reported in the paper (e.g., preliminary or failed experiments that didn't make it into the paper). 
    \end{itemize}
    
\item {\bf Code of ethics}
    \item[] Question: Does the research conducted in the paper conform, in every respect, with the NeurIPS Code of Ethics \url{https://neurips.cc/public/EthicsGuidelines}?
    \item[] Answer: \answerYes{}
    \item[] Justification: The research conforms with the NeurIPS Code of Ethics. The dataset comprises publicly available footage and newly collected videos of natural coastal phenomena; no personally identifying information is targeted, annotated or recorded, and no human subjects research is conducted. People do appear in the public-beach footage and are not blurred; the exposure is audited and documented in Appendix~\ref{sec:people_audit}, and the audit ships with the dataset.

    \item[] Guidelines:
    \begin{itemize}
        \item The answer \answerNA{} means that the authors have not reviewed the NeurIPS Code of Ethics.
        \item If the authors answer \answerNo, they should explain the special circumstances that require a deviation from the Code of Ethics.
        \item The authors should make sure to preserve anonymity (e.g., if there is a special consideration due to laws or regulations in their jurisdiction).
    \end{itemize}

\item {\bf Broader impacts}
    \item[] Question: Does the paper discuss both potential positive societal impacts and negative societal impacts of the work performed?
    \item[] Answer: \answerYes{}
    \item[] Justification: Positive societal impacts are discussed in Section~\ref{sec:discussion_and_limitations}: improved rip current detection has direct potential to reduce coastal drownings, which globally cause hundreds of fatalities annually (Supplementary Section~\ref{sec:global_studies}). Potential negative impacts and risks of misuse are also addressed: deployment without human oversight is explicitly discouraged, since model failures in ambiguous or rare scenarios could contribute to false reassurance and increased rather than decreased risk. We frame RipBench as a foundation for expert-assisted monitoring rather than autonomous safety systems.

    \item[] Guidelines:
    \begin{itemize}
        \item The answer \answerNA{} means that there is no societal impact of the work performed.
        \item If the authors answer \answerNA{} or \answerNo, they should explain why their work has no societal impact or why the paper does not address societal impact.
        \item Examples of negative societal impacts include potential malicious or unintended uses (e.g., disinformation, generating fake profiles, surveillance), fairness considerations (e.g., deployment of technologies that could make decisions that unfairly impact specific groups), privacy considerations, and security considerations.
        \item The conference expects that many papers will be foundational research and not tied to particular applications, let alone deployments. However, if there is a direct path to any negative applications, the authors should point it out. For example, it is legitimate to point out that an improvement in the quality of generative models could be used to generate Deepfakes for disinformation. On the other hand, it is not needed to point out that a generic algorithm for optimizing neural networks could enable people to train models that generate Deepfakes faster.
        \item The authors should consider possible harms that could arise when the technology is being used as intended and functioning correctly, harms that could arise when the technology is being used as intended but gives incorrect results, and harms following from (intentional or unintentional) misuse of the technology.
        \item If there are negative societal impacts, the authors could also discuss possible mitigation strategies (e.g., gated release of models, providing defenses in addition to attacks, mechanisms for monitoring misuse, mechanisms to monitor how a system learns from feedback over time, improving the efficiency and accessibility of ML).
    \end{itemize}
    
\item {\bf Safeguards}
    \item[] Question: Does the paper describe safeguards that have been put in place for responsible release of data or models that have a high risk for misuse (e.g., pre-trained language models, image generators, or scraped datasets)?
    \item[] Answer: \answerYes{}
    \item[] Justification: RipBench is public-beach footage, so people appear in it; we did not blur them, because blurring alters the water regions that are annotated in crowded scenes. The release therefore ships safeguards rather than claiming the issue is absent: a permissive person-detection audit over all $51{,}011$ sampled frames, released per video and per frame (Appendix~\ref{sec:people_audit}), so that users can select or exclude frames by exposure; no names, identities or audio anywhere in the dataset; and a CC BY-NC 4.0 licence that excludes commercial use. The released models perform rip current detection on visual input and pose no apparent dual-use risk beyond the deployment caveats already discussed in Section~\ref{sec:discussion_and_limitations}.

    \item[] Guidelines:
    \begin{itemize}
        \item The answer \answerNA{} means that the paper poses no such risks.
        \item Released models that have a high risk for misuse or dual-use should be released with necessary safeguards to allow for controlled use of the model, for example by requiring that users adhere to usage guidelines or restrictions to access the model or implementing safety filters. 
        \item Datasets that have been scraped from the Internet could pose safety risks. The authors should describe how they avoided releasing unsafe images.
        \item We recognize that providing effective safeguards is challenging, and many papers do not require this, but we encourage authors to take this into account and make a best faith effort.
    \end{itemize}

\item {\bf Licenses for existing assets}
    \item[] Question: Are the creators or original owners of assets (e.g., code, data, models), used in the paper, properly credited and are the license and terms of use explicitly mentioned and properly respected?
    \item[] Answer: \answerYes{}
    \item[] Justification: All prior datasets reused or extended in RipBench are cited (Section~\ref{sec:related_work}, Table~\ref{tab:rip_dataset_comparison}). The combined dataset is released under CC-BY-NC 4.0, with attribution to source videos and original annotation contributors where licensing permits. Baseline architectures are credited to their original authors, and pretrained weights are obtained from publicly distributed sources under their respective licenses.

    \item[] Guidelines:
    \begin{itemize}
        \item The answer \answerNA{} means that the paper does not use existing assets.
        \item The authors should cite the original paper that produced the code package or dataset.
        \item The authors should state which version of the asset is used and, if possible, include a URL.
        \item The name of the license (e.g., CC-BY 4.0) should be included for each asset.
        \item For scraped data from a particular source (e.g., website), the copyright and terms of service of that source should be provided.
        \item If assets are released, the license, copyright information, and terms of use in the package should be provided. For popular datasets, \url{paperswithcode.com/datasets} has curated licenses for some datasets. Their licensing guide can help determine the license of a dataset.
        \item For existing datasets that are re-packaged, both the original license and the license of the derived asset (if it has changed) should be provided.
        \item If this information is not available online, the authors are encouraged to reach out to the asset's creators.
    \end{itemize}

\item {\bf New assets}
    \item[] Question: Are new assets introduced in the paper well documented and is the documentation provided alongside the assets?
    \item[] Answer: \answerYes{}
    \item[] Justification: RipBench is a new benchmark introduced in this paper. The dataset, annotations, splits, baseline code, and metric implementations are released, accompanied by documentation describing the annotation protocol, per-task evaluation procedure (Supplementary Section~\ref{sec:metrics}), and training configurations (Supplementary Section~\ref{sec:hyperparameter_tuning}).

    \item[] Guidelines:
    \begin{itemize}
        \item The answer \answerNA{} means that the paper does not release new assets.
        \item Researchers should communicate the details of the dataset\slash code\slash model as part of their submissions via structured templates. This includes details about training, license, limitations, etc. 
        \item The paper should discuss whether and how consent was obtained from people whose asset is used.
        \item At submission time, remember to anonymize your assets (if applicable). You can either create an anonymized URL or include an anonymized zip file.
    \end{itemize}

\item {\bf Crowdsourcing and research with human subjects}
    \item[] Question: For crowdsourcing experiments and research with human subjects, does the paper include the full text of instructions given to participants and screenshots, if applicable, as well as details about compensation (if any)? 
    \item[] Answer: \answerNA{}
    \item[] Justification: The paper does not involve crowdsourcing or research with human subjects. Annotations were performed by two co-authors with domain expertise in coastal geomorphology and rip current dynamics, not by external crowdworkers.

    \item[] Guidelines:
    \begin{itemize}
        \item The answer \answerNA{} means that the paper does not involve crowdsourcing nor research with human subjects.
        \item Including this information in the supplemental material is fine, but if the main contribution of the paper involves human subjects, then as much detail as possible should be included in the main paper. 
        \item According to the NeurIPS Code of Ethics, workers involved in data collection, curation, or other labor should be paid at least the minimum wage in the country of the data collector. 
    \end{itemize}

\item {\bf Institutional review board (IRB) approvals or equivalent for research with human subjects}
    \item[] Question: Does the paper describe potential risks incurred by study participants, whether such risks were disclosed to the subjects, and whether Institutional Review Board (IRB) approvals (or an equivalent approval/review based on the requirements of your country or institution) were obtained?
    \item[] Answer: \answerNA{}
    \item[] Justification: The paper does not involve research with human subjects. Annotations were performed by co-authors, and the dataset comprises footage of natural coastal phenomena rather than data collected from study participants.

    \item[] Guidelines:
    \begin{itemize}
        \item The answer \answerNA{} means that the paper does not involve crowdsourcing nor research with human subjects.
        \item Depending on the country in which research is conducted, IRB approval (or equivalent) may be required for any human subjects research. If you obtained IRB approval, you should clearly state this in the paper. 
        \item We recognize that the procedures for this may vary significantly between institutions and locations, and we expect authors to adhere to the NeurIPS Code of Ethics and the guidelines for their institution. 
        \item For initial submissions, do not include any information that would break anonymity (if applicable), such as the institution conducting the review.
    \end{itemize}

\item {\bf Declaration of LLM usage}
    \item[] Question: Does the paper describe the usage of LLMs if it is an important, original, or non-standard component of the core methods in this research? Note that if the LLM is used only for writing, editing, or formatting purposes and does \emph{not} impact the core methodology, scientific rigor, or originality of the research, declaration is not required.
    \item[] Answer: \answerNA{}
    \item[] Justification: LLMs were not used as an important, original, or non-standard component of the core methodology. Any LLM use was limited to writing, editing, and formatting.
    \item[] Guidelines:
    \begin{itemize}
        \item The answer \answerNA{} means that the core method development in this research does not involve LLMs as any important, original, or non-standard components.
        \item Please refer to our LLM policy in the NeurIPS handbook for what should or should not be described.
    \end{itemize}

\end{enumerate}

\end{document}